\documentclass{article}
\usepackage[nonatbib,preprint]{neurips_2026}

\usepackage[utf8]{inputenc} 
\usepackage[T1]{fontenc}    
\usepackage{hyperref}       
\usepackage{url}            
\usepackage{booktabs}       
\usepackage{amsfonts}       
\usepackage{nicefrac}       
\usepackage{microtype}      
\usepackage[dvipsnames]{xcolor} 
\usepackage{rotating} 
\usepackage{tikz}
\usepackage{tcolorbox}
\tcbuselibrary{raster}
\usepackage{enumitem}
\usepackage{tkz-euclide}
\usepackage{bbding}
\PassOptionsToPackage{numbers}{natbib}
\usepackage{natbib}
\usepackage{amssymb}
\usepackage{hyperref}
\usepackage{graphicx}
\usepackage{booktabs}
\usepackage{xspace}
\usepackage{multirow}
\usepackage{amsthm}
\usepackage{amsmath}
\usepackage[capitalize]{cleveref}
\crefname{definition}{Def.}{Defs.}
\Crefname{definition}{Def.}{Defs.}
\crefname{appendix}{App.}{Apps.}
\Crefname{appendix}{App.}{Apps.}
\crefname{section}{Sec.}{Sections}
\Crefname{section}{Sec.}{Sections}
\crefname{theorem}{Thm.}{Thms.}
\Crefname{theorem}{Thm.}{Thms.}
\crefname{figure}{Fig.}{Figs.}
\Crefname{figure}{Fig.}{Figs.}
\crefname{table}{Tab.}{Tabs.}
\Crefname{table}{Tab.}{Tabs.}

\usepackage{minitoc}

\usepackage{thmtools}
\newtheorem{theorem}{Theorem}
\renewcommand{\thetheorem}{%
  \ifnum\value{theorem}=-1
    00%
  \else
    \number\numexpr\value{theorem}\relax
  \fi
}
\newtheorem{lemma}{Lemma}

\newtheorem{definition}{\textbf{Definition}}

\newtheorem{setting}{}

\def\cguarantee{ForestGreen}
\def\cfailure{Red}
\def\csuccess{BurntOrange}
\def\vb{{\boldsymbol{b}}}
\def\vx{{\boldsymbol{x}}}

\def\veta{{\boldsymbol{\eta}}}
\def\vu{{\boldsymbol{u}}}
\def\vv{{\boldsymbol{v}}}

\def\vz{{\boldsymbol{z}}}
\def\vw{{\boldsymbol{w}}}
\def\vnu{{\boldsymbol{\nu}}}
\def\vzeta{{\boldsymbol{\zeta}}}
\def\vomega{{\boldsymbol{\omega}}}
\def\vmu{{\boldsymbol{\mu}}}
\def\veta{{\boldsymbol{\eta}}}
\def\vzero{{\boldsymbol{0}}}

\def\vtau{{\boldsymbol{\tau}}}

\def\supp{{\ensuremath{\text{supp}}\xspace}}
\def\wa{\ensuremath{\frac{1}{\alpha(\alpha-1)}q_{\vzero}^{1-\alpha}}\xspace}
\def\wao{\ensuremath{\frac{1}{\alpha(\alpha-1)}q_{0}^{1-\alpha}}\xspace}
\def\wkl{\ensuremath{-\log q_{\vzero}}\xspace}
\def\parset{\ensuremath{\Omega\xspace}}
\def\notparset{\ensuremath{\overline{\Omega}\xspace}}

\def\csupp{\ensuremath{p} is supported around \ensuremath{\vmu} on a set of positive measure\xspace}
\def\csuppI{\ensuremath{I} is supported around \ensuremath{\vmu} on a set of positive measure\xspace}
\def\yes{\textcolor{ForestGreen}{\CheckmarkBold}}
\def\no{\textcolor{Red}{\XSolidBrush}}
\def\limc{\ensuremath{\lim_{u \rightarrow 0^+}-g'(u) \geq 0}\xspace}
\def\limca{\ensuremath{\lim_{u \rightarrow 0^+}g'(u) \geq 0}\xspace}

\def\lr{{\ensuremath{10^{-3}}}}

\def\nruns{$5$\xspace}

\hypersetup{
colorlinks=true,linkcolor=pink!50!brown,citecolor=teal,urlcolor=orange
}

\title{Even More Guarantees for Variational Inference\\in the Presence of Symmetries}

\author{
    Lena Zellinger\\
    School of Informatics\\
    University of Edinburgh\\
    l.zellinger@sms.ed.ac.uk
  \And
    Antonio Vergari\\
    School of Informatics\\
    University of Edinburgh\\
    avergari@ed.ac.uk
}

\begin{document}

\maketitle

\begin{abstract}
When approximating an intractable density via variational inference \emph{(VI)} the variational family is typically chosen as a simple parametric family that very likely does not contain the target.
This raises the question: \emph{Under which conditions can we recover characteristics of the target despite misspecification?}
In this work, we extend previous theoretical results on robust VI with location-scale families under target symmetries in two substantial ways: (1) We open them up to a wider range of divergences by providing sufficient conditions for exact recovery of the target mean and correlation matrix when using the forward Kullback-Leibler divergence and $\alpha$-divergences. (2) By doing so, we find that we can drop the restrictive assumption of a log-concave target made in previous work, allowing us to give guarantees for a wider range of targets, including multi-modal ones. 
In our experiments, we show how our guarantees can serve as guidelines for the choice of the variational family and $\alpha$-value and we illustrate on a diverse set of examples how and why optimization can fail in the absence of our sufficient conditions.
\end{abstract}

\section{Introduction}
Many tasks in machine learning (ML) and statistics revolve around inferring intractable statistical quantities of a given target distribution $p$, such as expected predictions \citep{khosravi2019expect,rainforth2020target,chen2023algorithms, wasserman2000bayesian} and moments \citep{tierney1986accurate, bowyer2023using}. 
Variational inference \citep[VI,][]{ranganath2014black, blei2017variational} recasts the problem of performing these intractable inference tasks over $p$ as optimization: Given a variational family $\mathcal{Q}$, we find a surrogate $q^{\star} \in \mathcal{Q}$ by minimizing a divergence of choice $D$, and then use $q^{\star}$ to perform tractable inference. Typical choices for $\mathcal{Q}$ include simple parametric distributions \citep{tomczak2020efficient, modi2023variational} and finite mixture models \citep{morningstar2021automatic, arenz2022unified, hotti2024efficient, zellingerapproximate}. 

\begin{figure}[t]
\begin{minipage}{0.73\textwidth}
\large
\resizebox{\textwidth}{!}{
\begin{tabular}{rccccc}
\toprule
& \multicolumn{2}{c}{FKL} & 
\multicolumn{1}{c}{$\alpha=2.0$} & \multicolumn{1}{c}{$\alpha=0.5$}\\
\cmidrule(lr){2-3}
$q$ & \multicolumn{1}{c}{Gaussian} & \multicolumn{1}{c}{Student-t} & \multicolumn{1}{c}{Student-t} & \multicolumn{1}{c}{Student-t}\\
$\Delta\vmu$ & 
$\textcolor{ForestGreen}{\boldsymbol{2.1 \cdot 10^{-4}}}^{\boldsymbol{*}}$ ~(\ref{theorem:theorem_fkl_1}) & $\textcolor{orange}{\boldsymbol{2.9 \cdot 10^{-4}}}$ & $\textcolor{ForestGreen}{\boldsymbol{1.7 \cdot 10^{-4}}}$ ~(\ref{theorem:theorem_alpha_1}) & $\textcolor{orange}{\boldsymbol{1.5 \cdot 10^{-3}}}$ \\

$\Delta$Corr & 
$\textcolor{ForestGreen}{\boldsymbol{4.7 \cdot 10^{-5}}}^{\boldsymbol{*}}$ & $\textcolor{orange}{\boldsymbol{9.4 \cdot 10^{-3}}}$ & $\textcolor{red}{\boldsymbol{1.8 \cdot 10^{-2}}}$ & $\textcolor{red}{\boldsymbol{3.2 \cdot 10^{-2}}}$ \\
\includegraphics[scale=0.22]{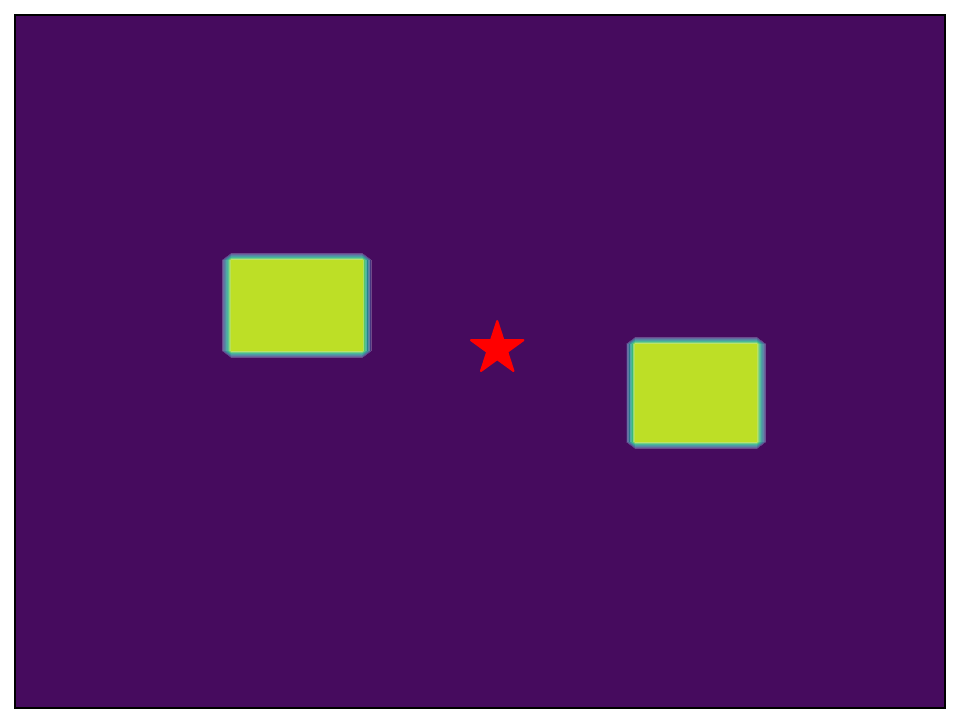} &
\includegraphics[scale=0.22]{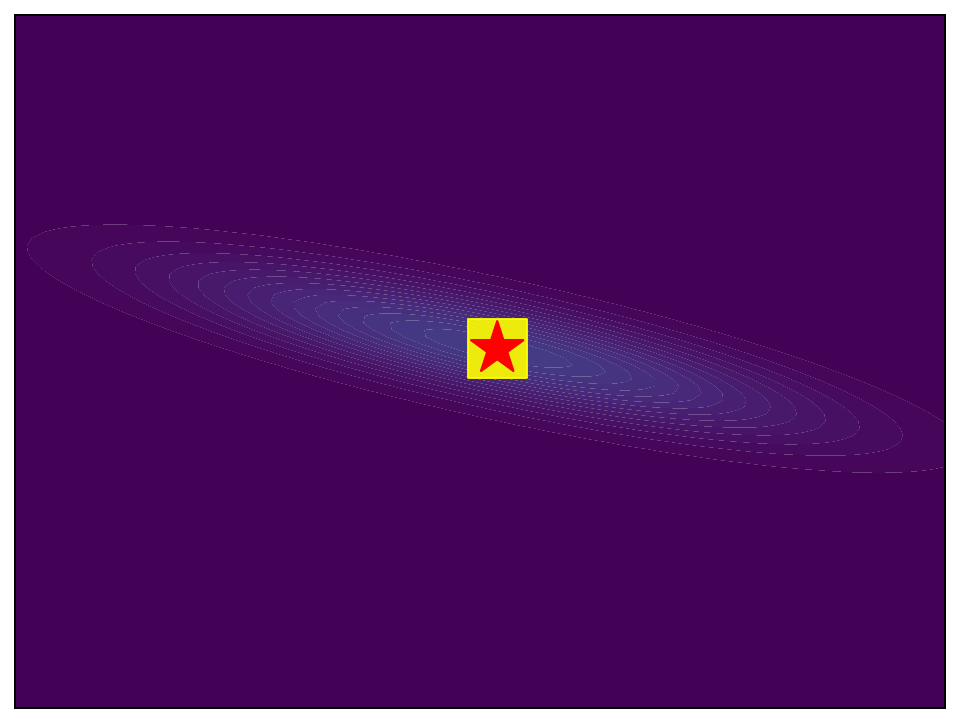} &
\includegraphics[scale=0.22]{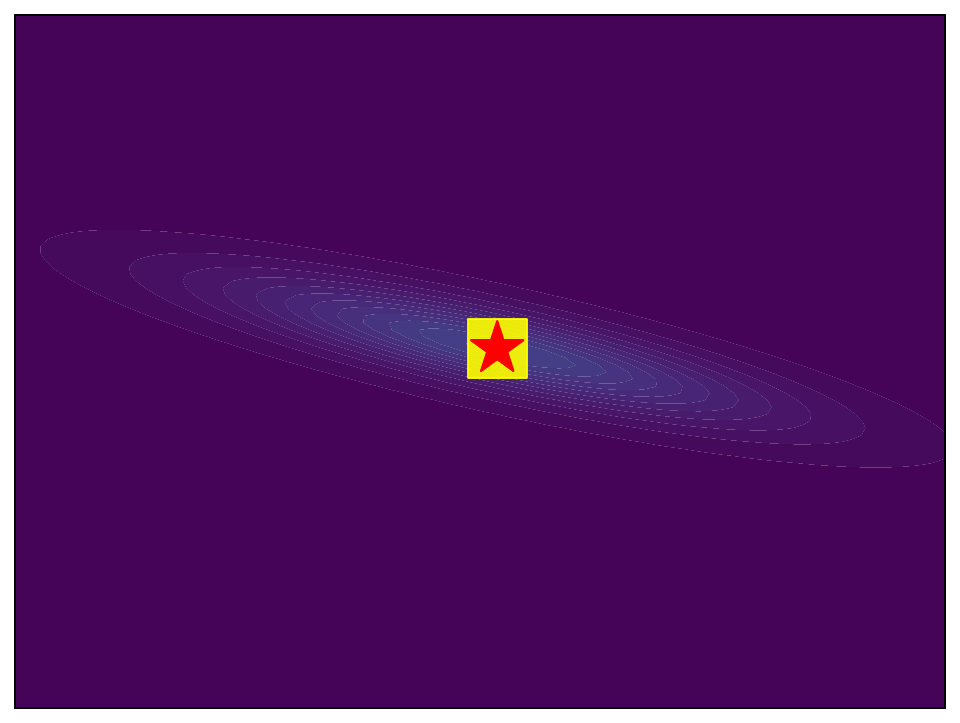} &
\includegraphics[scale=0.22]{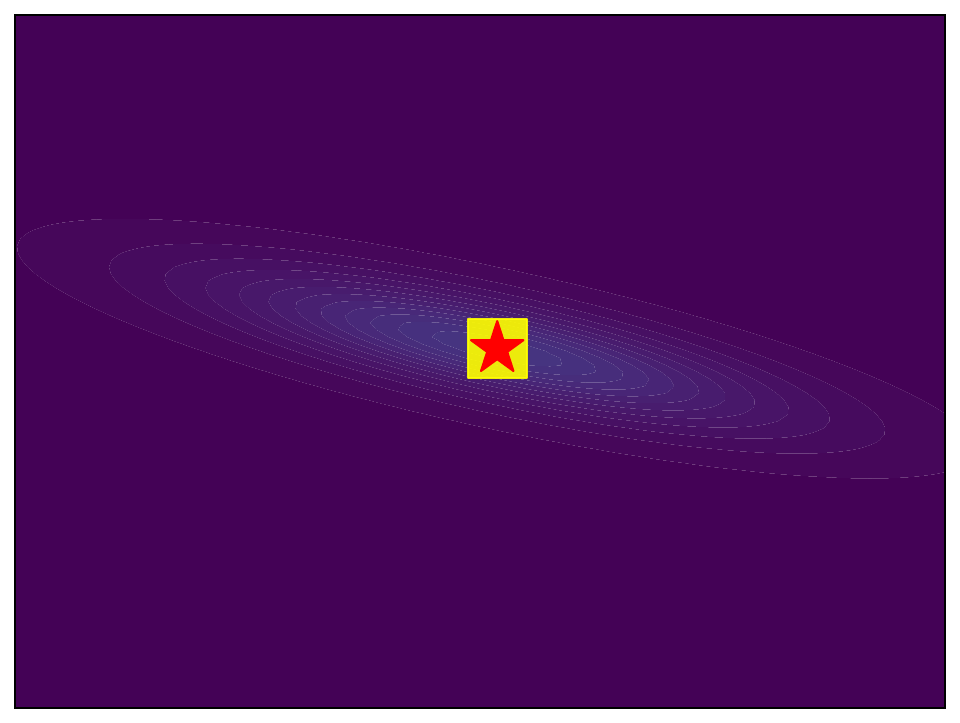} &
\includegraphics[scale=0.22]{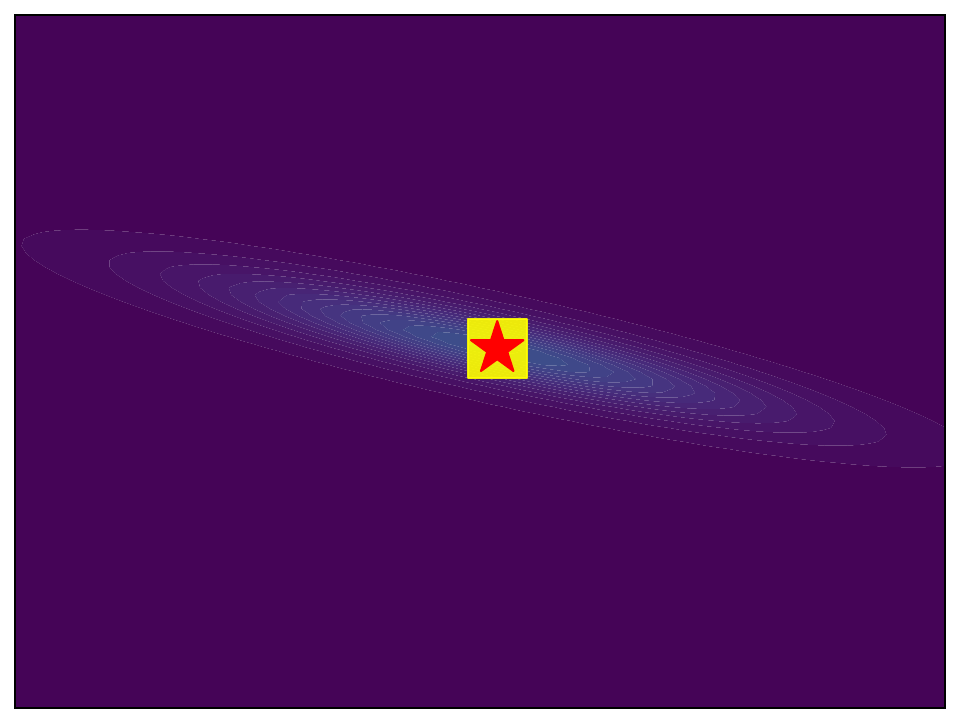} \\
$\Delta\vmu$ & 
$\textcolor{ForestGreen}{\boldsymbol{1.5 \cdot 10^{-4}}}^{\boldsymbol{*}}$ ~(\ref{theorem:theorem_fkl_1}, \ref{theorem:theorem_fkl_2}) & $\textcolor{orange}{\boldsymbol{2.2 \cdot 10^{-4}}}$ & $\textcolor{ForestGreen}{\boldsymbol{1.5 \cdot 10^{-4}}}$ ~(\ref{theorem:theorem_alpha_1},~\ref{theorem:theorem_alpha_2}) & $\textcolor{red}{\boldsymbol{9.7 \cdot 10^{-1}}}$ \\

$\Delta$Corr & 
$\textcolor{ForestGreen}{\boldsymbol{1.7 \cdot 10^{-3}}}^{\boldsymbol{*}}$ & $\textcolor{orange}{\boldsymbol{7.0 \cdot 10^{-3}}}$ & $\textcolor{orange}{\boldsymbol{7.1 \cdot 10^{-3}}}$ & $\textcolor{red}{\boldsymbol{6.5 \cdot 10^{-1}}}$ \\
\includegraphics[scale=0.22]{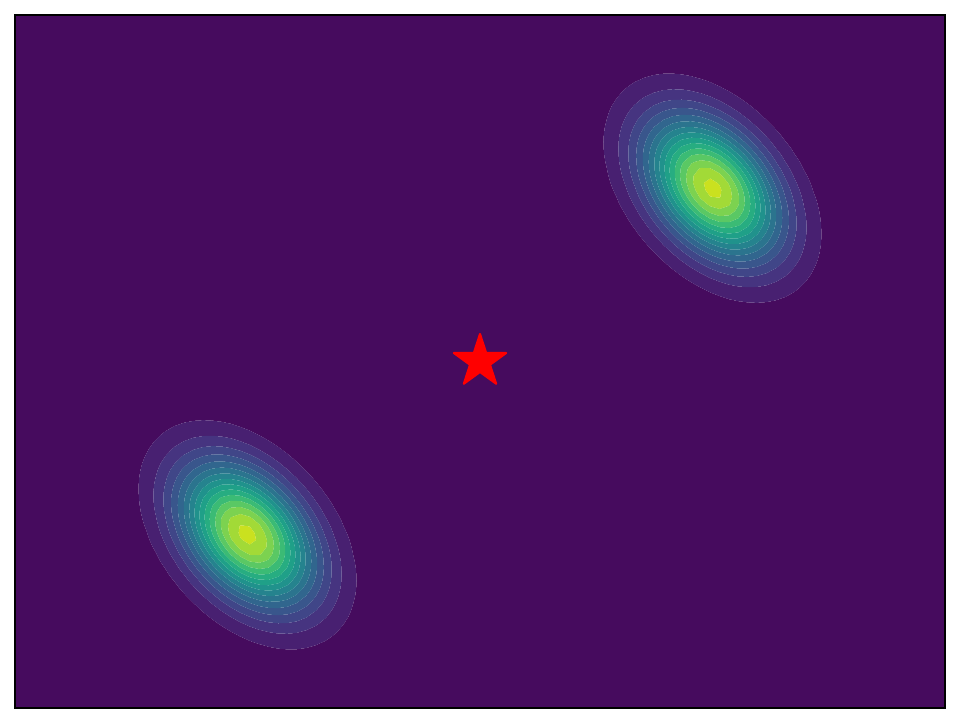} 
& 
\includegraphics[scale=0.22]{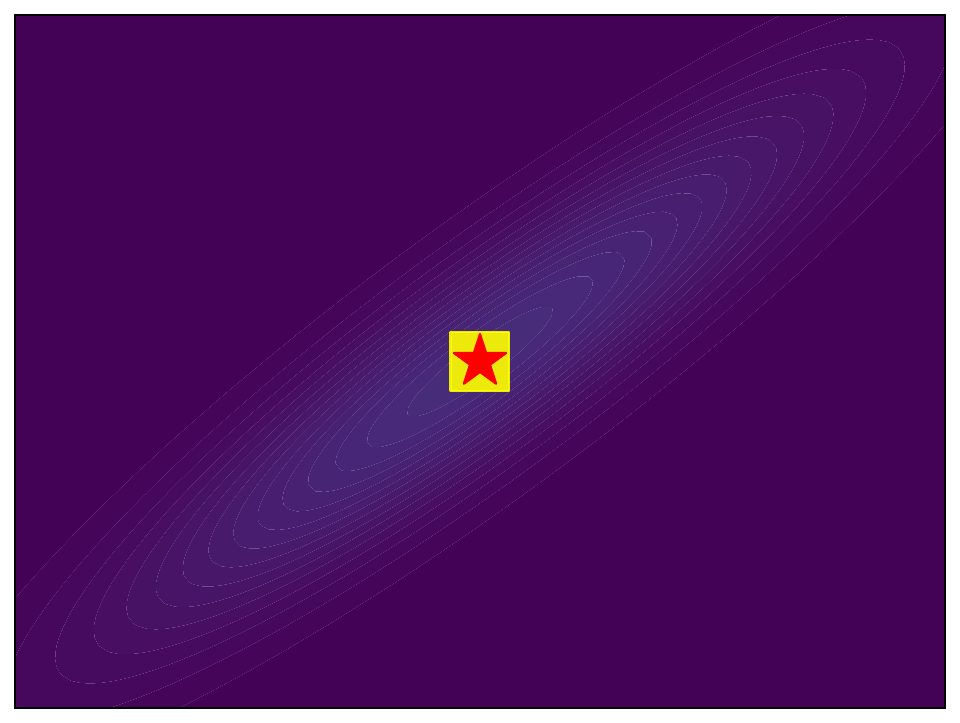} &
\includegraphics[scale=0.22]{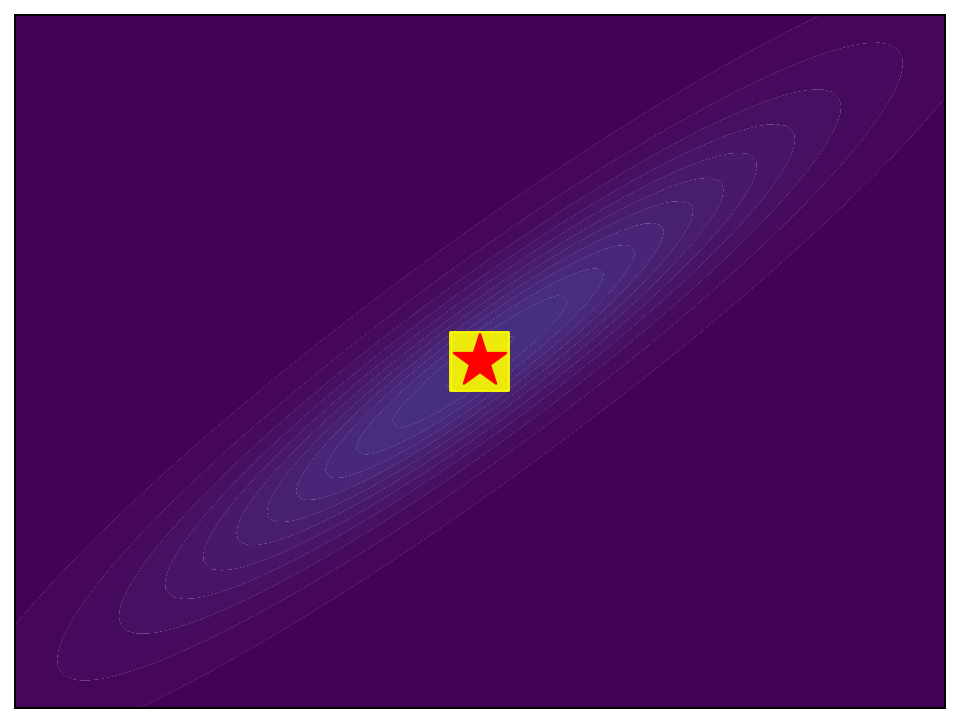} &
\includegraphics[scale=0.22]{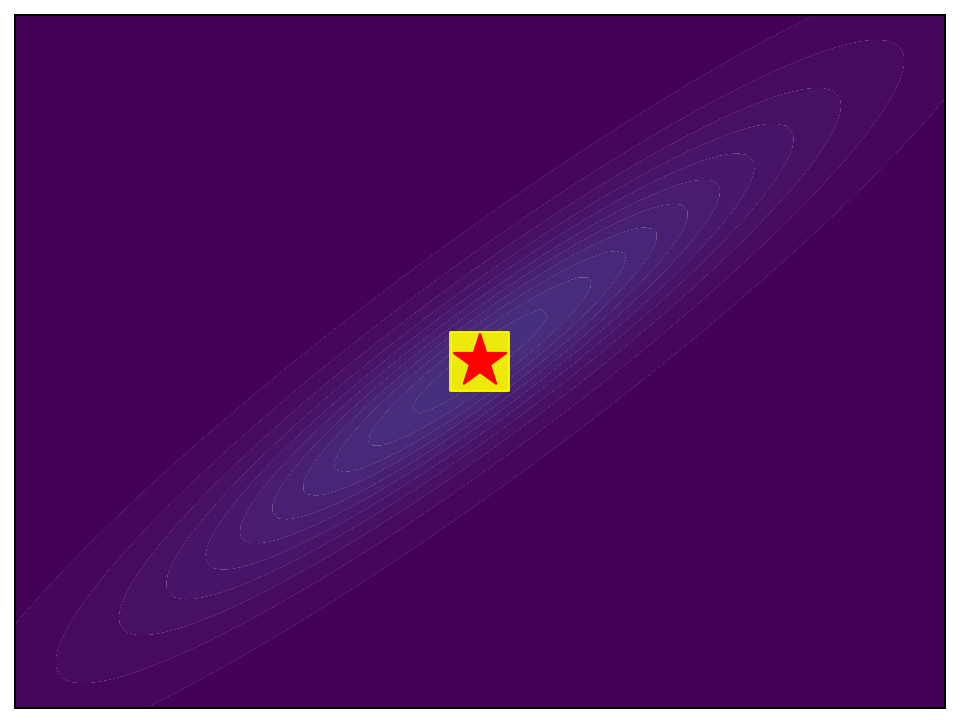} & \includegraphics[scale=0.22]{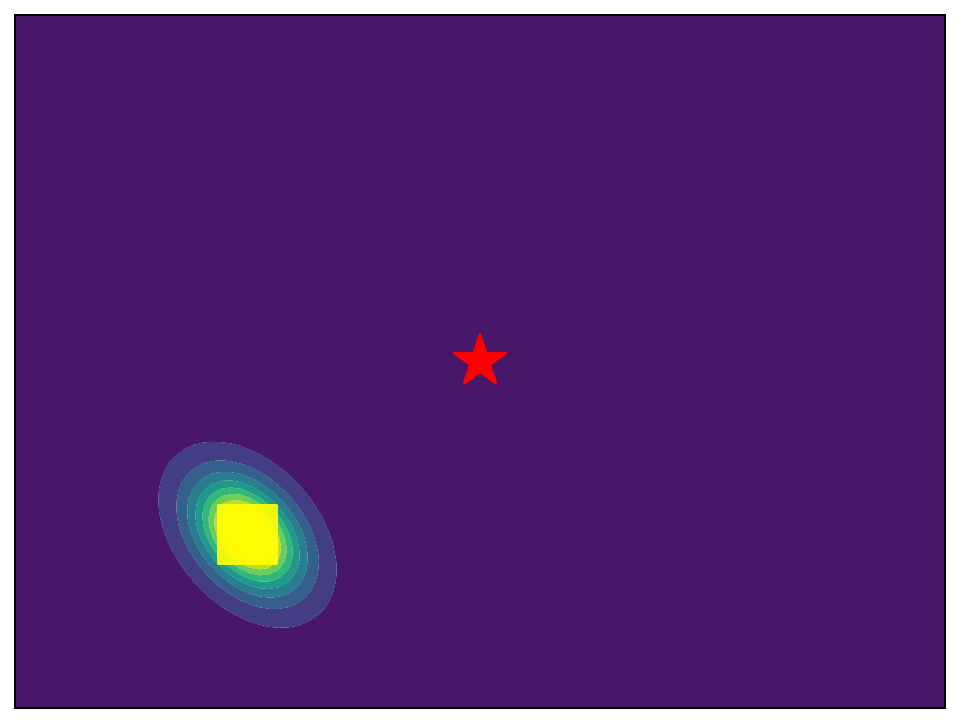}
\\
$q$ & \multicolumn{1}{c}{Gaussian} & \multicolumn{1}{c}{Student-t} & \multicolumn{1}{c}{Gaussian} & \multicolumn{1}{c}{Student-t}\\
$\Delta\vmu$ & 
$\textcolor{ForestGreen}{\boldsymbol{1.7 \cdot 10^{-4}}}^{\boldsymbol{*}}$ ~(\ref{theorem:theorem_fkl_1}, \ref{theorem:theorem_fkl_2}) & $\textcolor{orange}{\boldsymbol{2.5 \cdot 10^{-4}}}$ & $\textcolor{ForestGreen}{\boldsymbol{5.3 \cdot 10^{-4}}}$ ~(\ref{theorem:theorem_alpha_1},~\ref{theorem:theorem_alpha_2}) & $\textcolor{orange}{\boldsymbol{2.4 \cdot 10^{-3}}}$ \\
$\Delta$Corr & 
$\textcolor{ForestGreen}{\boldsymbol{1.5 \cdot 10^{-5}}}^{\boldsymbol{*}}$ ~(\ref{theorem:theorem_fkl_corr}) & $\textcolor{orange}{\boldsymbol{1.5 \cdot 10^{-5}}}$ & $\textcolor{ForestGreen}{\boldsymbol{6.5 \cdot 10^{-6}}}$ ~(\ref{theorem:theorem_alpha_corr}) & $\textcolor{orange}{\boldsymbol{6.9 \cdot 10^{-5}}}$ \\
\includegraphics[scale=0.22]{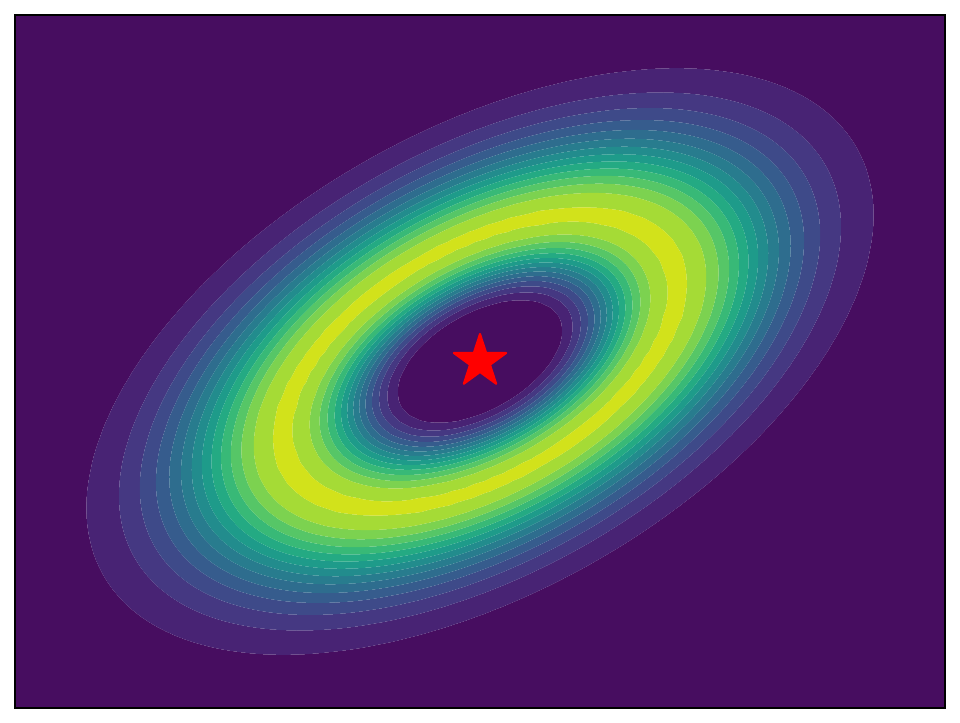} &
\includegraphics[scale=0.22]{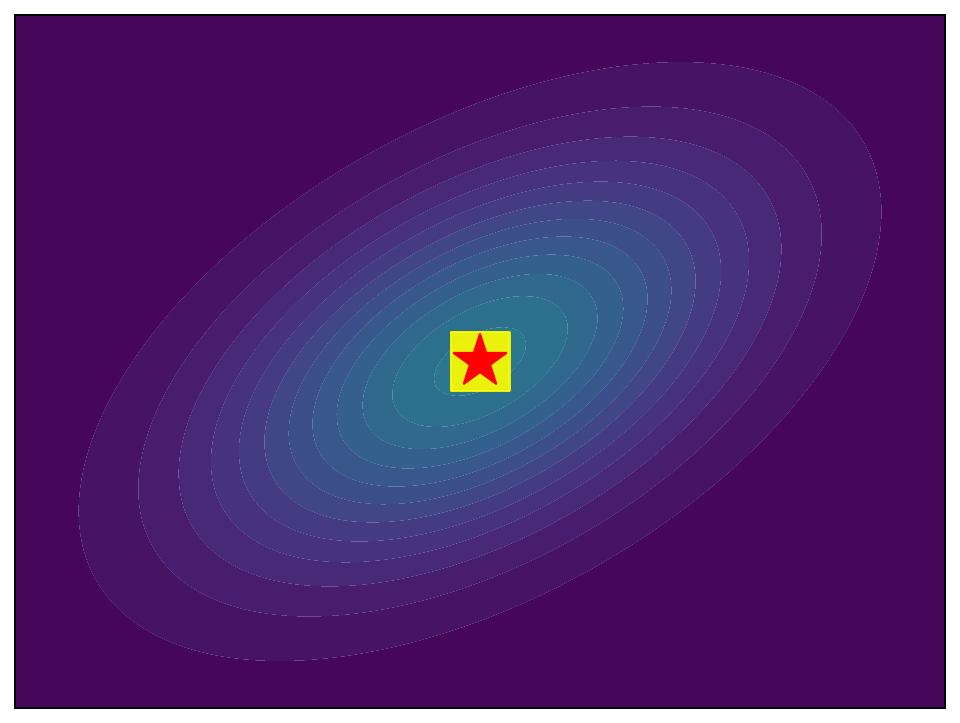} &
\includegraphics[scale=0.22]{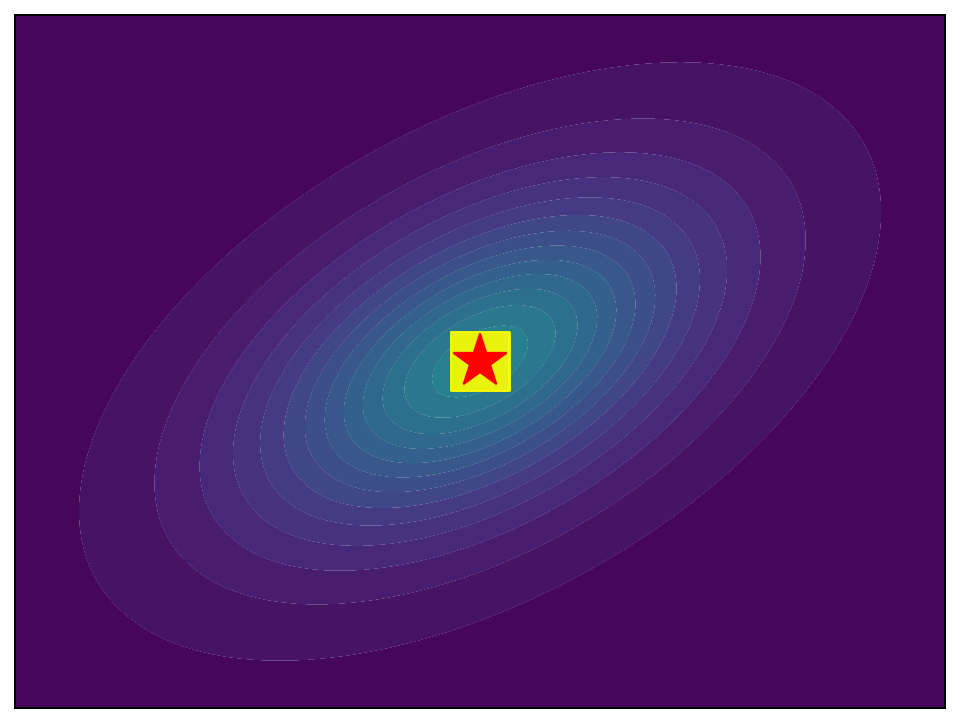} &
\includegraphics[scale=0.22]{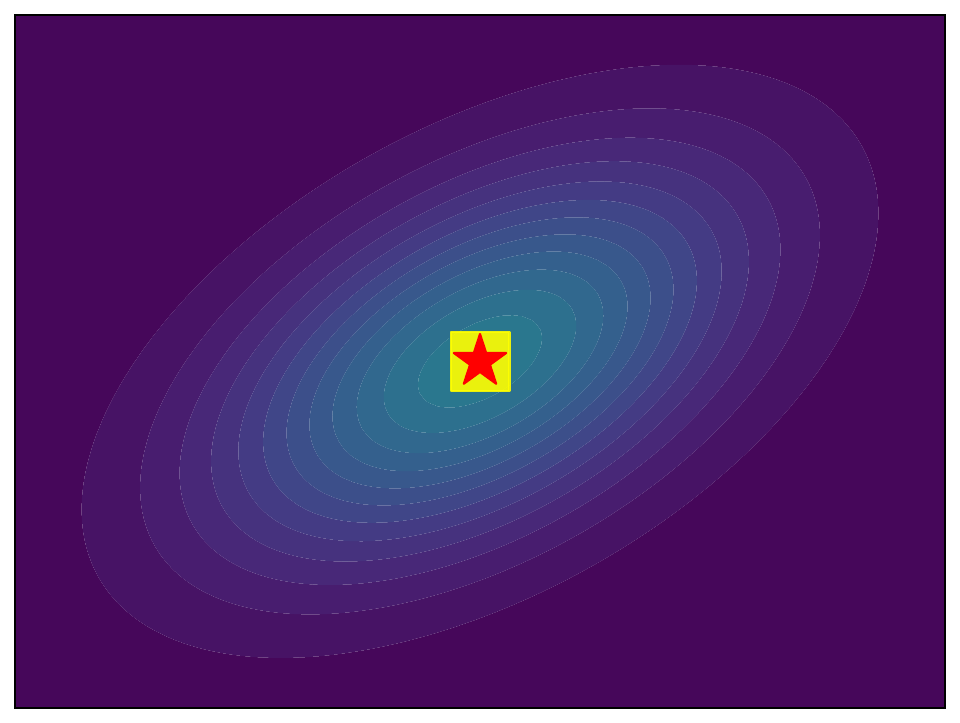} & \includegraphics[scale=0.22]{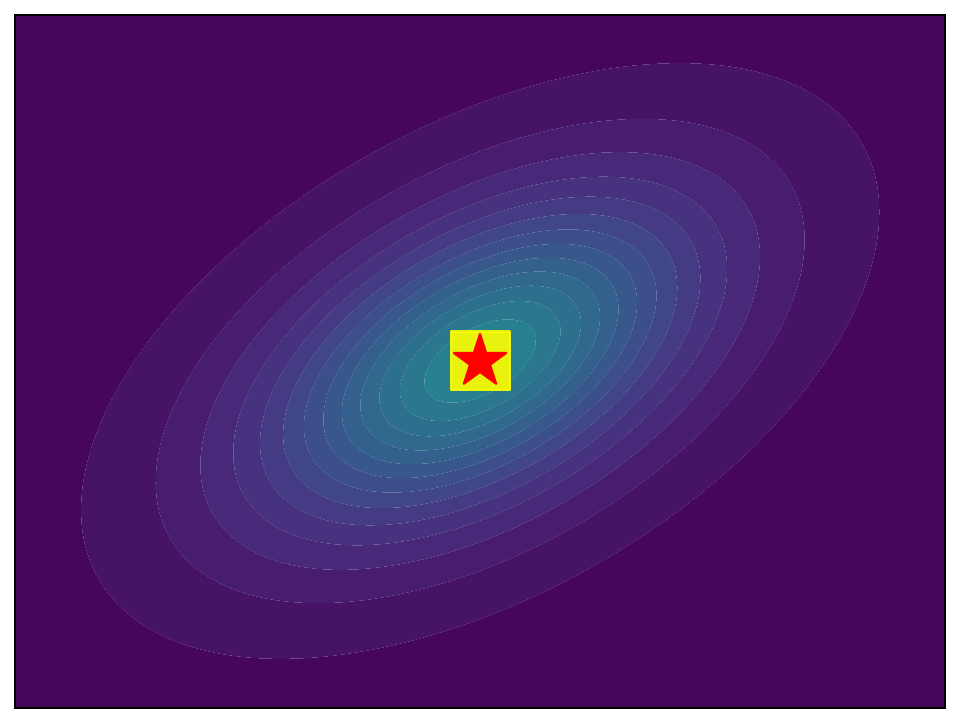} &
\\
$\Delta\vmu$ & 
$\textcolor{ForestGreen}{\boldsymbol{1.6 \cdot 10^{-4}}}^{\boldsymbol{*}}$ & $\textcolor{red}{\boldsymbol{9.7 \cdot 10^{-2}}}$ & $\textcolor{red}{\boldsymbol{8.7 \cdot 10^{-2}}}$ & $\textcolor{red}{\boldsymbol{1.2 \cdot 10^{-1}}}$ \\
$\Delta\text{Corr}$ & 
$\textcolor{ForestGreen}{\boldsymbol{7.3 \cdot 10^{-5}}}^{\boldsymbol{*}}$ & $\textcolor{orange}{\boldsymbol{8.7 \cdot 10^{-3}}}$ & $\textcolor{orange}{\boldsymbol{5.6 \cdot 10^{-3}}}$ & $\textcolor{red}{\boldsymbol{1.3 \cdot 10^{-2}}}$ \\
\includegraphics[scale=0.22]{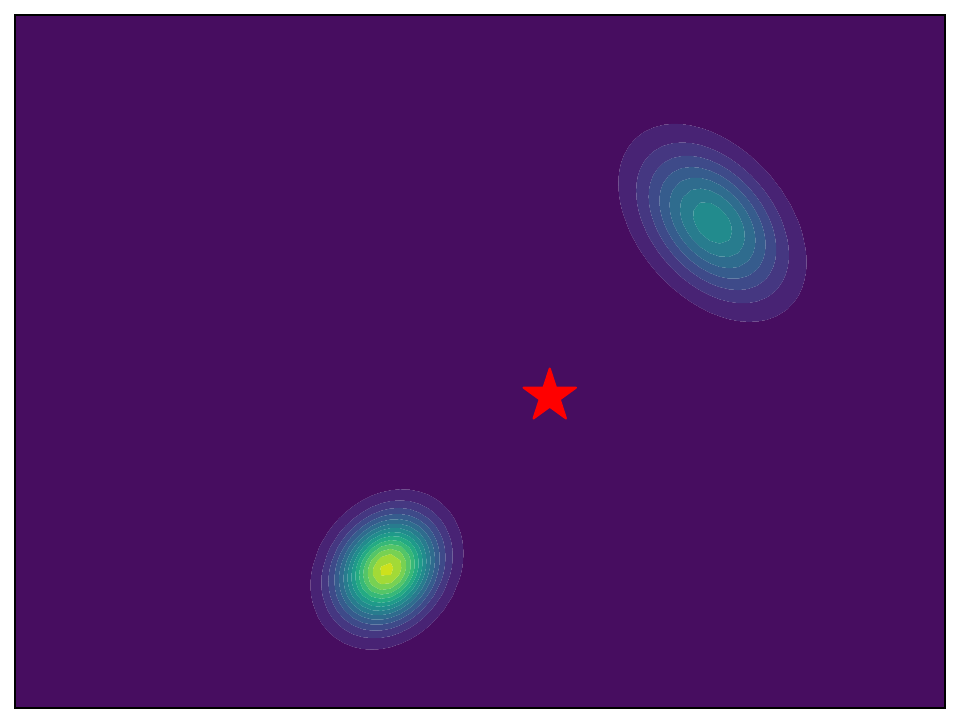} &
\includegraphics[scale=0.22]{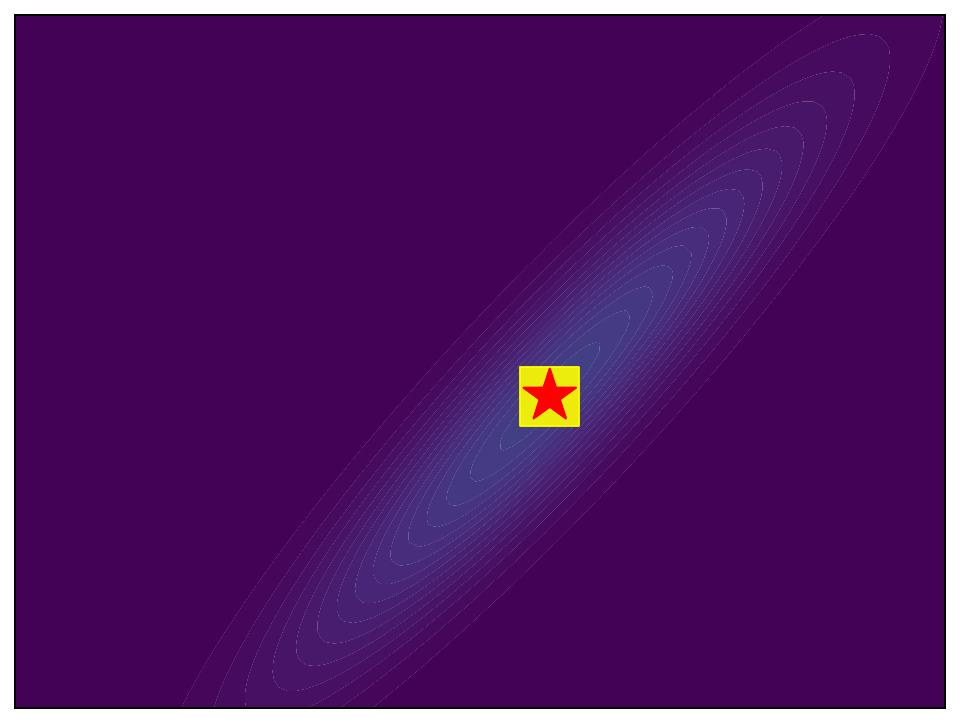} &
\includegraphics[scale=0.22]{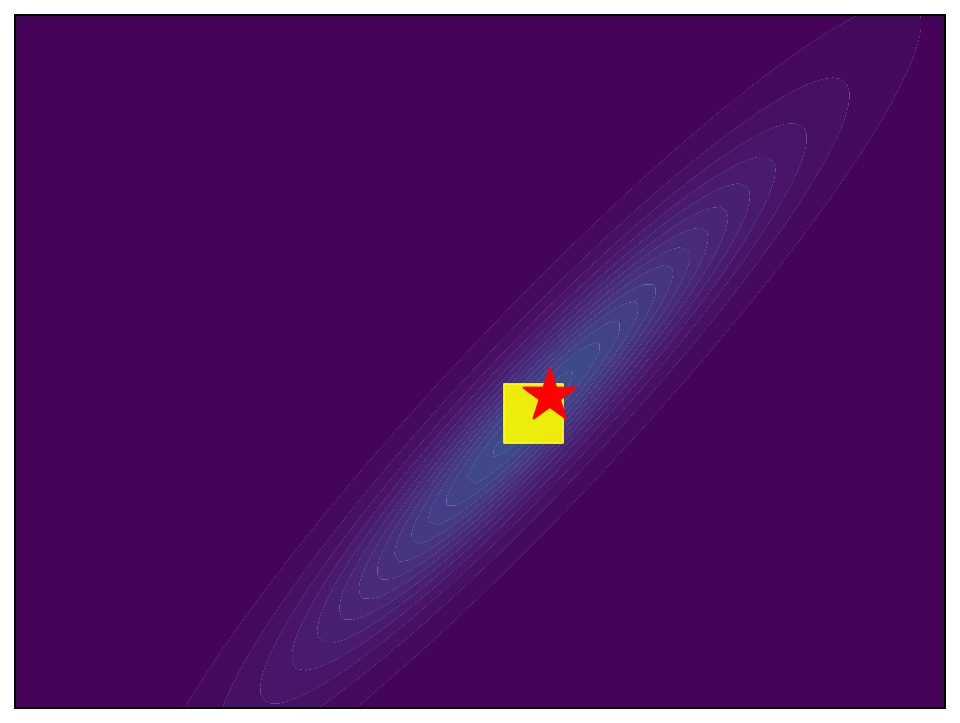} &
\includegraphics[scale=0.22]{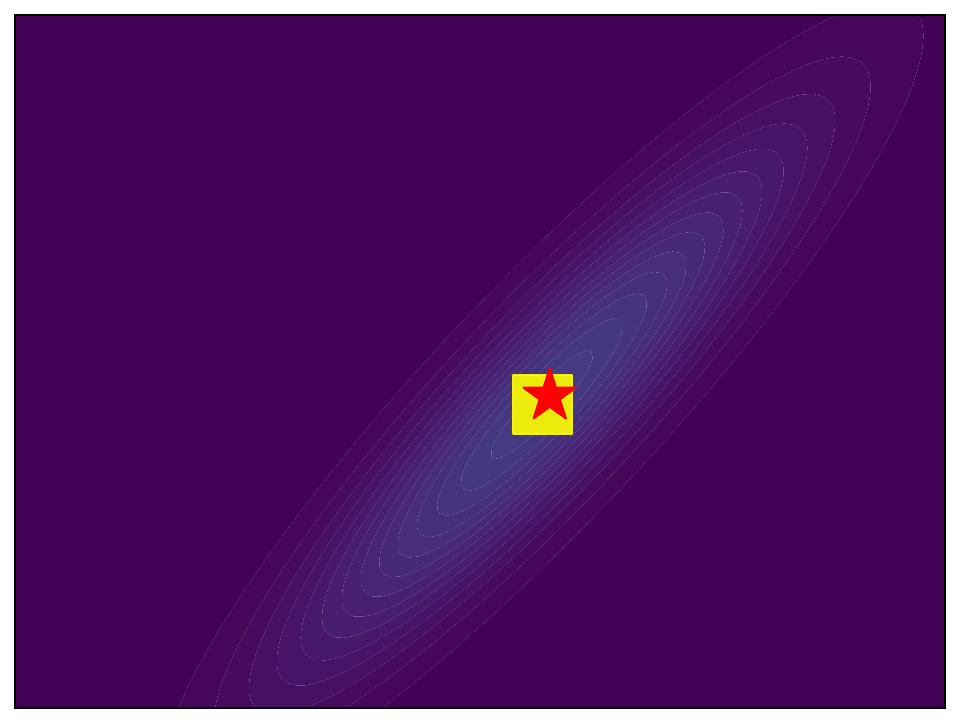} & 
\includegraphics[scale=0.22]{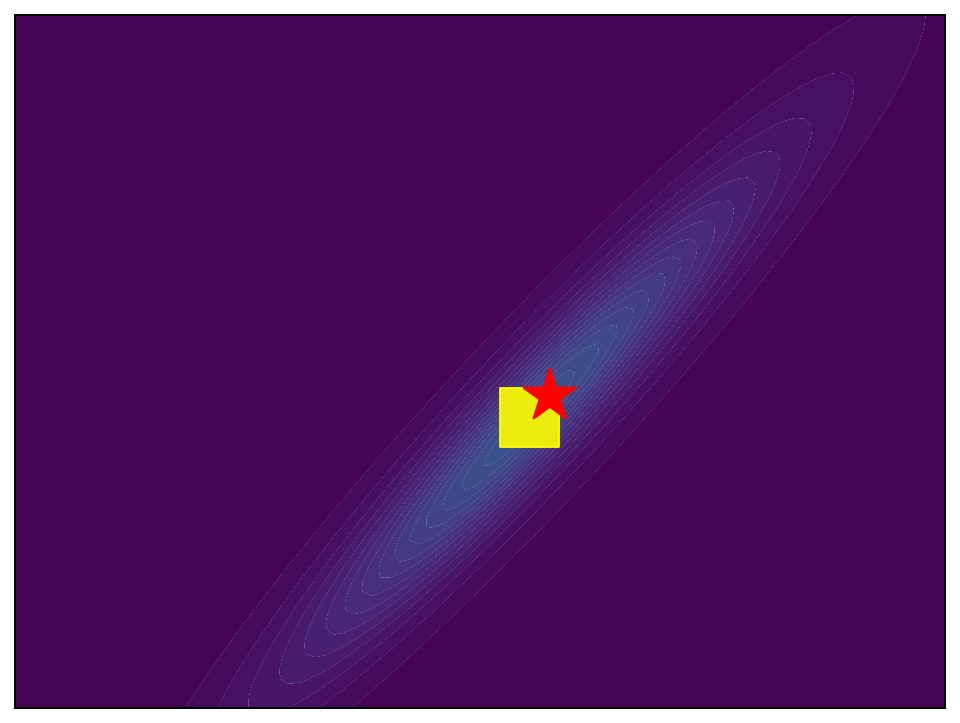} 
\\
\bottomrule
\end{tabular}
}
\end{minipage}\hfill\begin{minipage}{.25\textwidth}
\caption{\textbf{We provide theoretical guarantees to recover the location and correlation of multi-modal targets which {correspond to low error in practice}}. The mean of the target $p$ is indicated via a red star.
The mean of $q$ is denoted by a yellow square. The errors $\Delta\vmu$ and $\Delta\mathsf{Corr}$ are measured as described in \cref{sec:learning}. 
We distinguish 3 cases (1) \textbf{\textcolor{\cguarantee}{low error}} ($< 10^{-2}$) \textbf{\textcolor{\cguarantee}{as predicted by our guarantees}} (see corresponding theorems in parentheses) or previous results (indicated via *), (2) \textbf{\textcolor{\csuccess}{no guarantee but low error in practice}}, (3) \textbf{\textcolor{\cfailure}{no guarantee and high error}} ($> 10^{-2}$).}
\label{fig:results_simulations}
\end{minipage}
\end{figure}

In practice, $\mathcal{Q}$ is therefore often \emph{misspecified}, i.e., $p \notin \mathcal{Q}$ \citep{wang2019variational}. In such cases, the choice of $\mathcal{Q}$ and $D$ can greatly affect the resulting approximation of the target (see \cref{fig:results_simulations} for examples). While $\mathcal{Q}$ is not expressive enough to fully capture $p$, we might still be able to capture informative summary statistics of the target, such as its mean and correlations between the variables. 
This raises the question: \emph{Under which conditions on $p$, $\mathcal{Q}$, and $D$ are we \textbf{guaranteed} to capture relevant characteristics of the target despite misspecification?} 

Most existing results studying the quality of VI approximations are \emph{asymptotic} \citep{hall2011asymptotic, zhang2020convergence, wang2019frequentist, yang2020alpha, domke2023provable, bertholom2024asymptotics,pati2018statistical}, relying, for instance, on the Bernstein-von-Mises theorem \citep{vd1998asymptotic}, or providing post-hoc diagnostics \citep{yao2018yes,huggins2020validated}. Guarantees for \emph{exact} recovery of target statistics are still sparse and often assume \emph{Gaussian} approximations: \citet{wainwright2008graphical} show that the optimal Gaussian approximation w.r.t. the FKL exactly matches the target mean and covariance. \citet{katsevich2024approximation} provide error bounds on the mean and covariance for Gaussian approximations, but do not guarantee exact recovery. Studying proposals beyond Gaussians is relevant in practice, as surrogates with heavier tails can be preferable for expectation estimation via importance sampling \citep{yao2018yes,delyon2021safe,korba2022adaptive}.

Recently, \citet{margossian2024variational} exploited \emph{symmetries in the target distribution} to provide sufficient conditions for the \emph{exact recovery of the target mean and correlation matrix} when $\mathcal{Q}$ is a location-scale family. Symmetries can arise naturally in Bayesian posteriors when either the likelihood or the prior exhibits symmetries and dominates the other (for instance, when only little data is available) \citep{margossian2024variational}. Their initial results considered the \emph{reverse Kullback-Leibler divergence (RKL)}, which is a standard objective in VI \citep{blei2017variational}, but known for its mode-seeking behavior, which can lead to poor target coverage \citep{giordano2018covariances,margossian2025variational}.
\citet{margossian2025variational} subsequently extended their analysis to $f$-divergences \citep{renyi1961measures}, which 
generalize other popular divergences beyond the RKL, such as the \emph{forward KL divergence (FKL)} \citep{kullback1997information, wainwright2008graphical, jerfel2021variational, campbell2019universal} and \emph{$\alpha$-divergence ($\alpha$-DIV)} \citep{cichocki2010families, li2016renyi, daudel2021mixture, guilmeau2023variational, guilmeau2024adaptive}. Unfortunately, only the RKL was shown to provably attain a unique minimizer under their conditions, while the remaining divergences are only guaranteed to reach a stationary point of the location parameter at the target mean. 
This begs the question: \emph{Can we retrieve similar guarantees for VI when using the FKL and $\alpha$-DIV?}

We fill this gap by extending the analysis of
\citet{margossian2024variational,margossian2025generalized} and providing the following contributions:
\textbf{(1)} In \cref{sec:mean_recovery}, we derive \emph{complementary sufficient conditions} for the exact recovery of the mean when optimizing the FKL and $\alpha$-DIV. 
We find that we can guarantee exact recovery of the mean with the FKL when placing mild assumptions on the base distribution of the location-scale family. For the $\alpha$-DIV, we establish a more fine-grained criterion that additionally depends on the value of $\alpha$. Notably, our analysis, unlike previous results \citep{margossian2024variational, margossian2025generalized}, does not rely on log-concavity of the target, and hence also applies to multi-modal targets, such as mixture distributions \citep{neal1992bayesian,marin2005bayesian}, allowing us to provide guarantees for a wider range of realistic targets. \textbf{(2)} In \cref{sec:corr_recovery}, we provide sufficient conditions for simultaneously recovering the correlation matrix of the target. Akin to \citet{margossian2024variational,margossian2025generalized}, we place additional symmetry requirements on the target for correlation recovery, but still alleviate the assumption of log-concavity. 
\textbf{(3)} Lastly, in \cref{sec:simulations}, we discuss how our guarantees can inform the choice of the variational family in practice, and show how optimization can fail to recover target characteristics when our sufficient conditions are not met.

\section{Location-scale VI under symmetries}\label{sec:loc_scale_vi}

\begin{figure*}[t]
\resizebox{\textwidth}{!}{
\begin{tabular}{cccc}
{\footnotesize asymmetric} & \footnotesize even symmetric & \footnotesize spherically symmetric & \footnotesize elliptically symmetric\\
\includegraphics[scale=0.17]{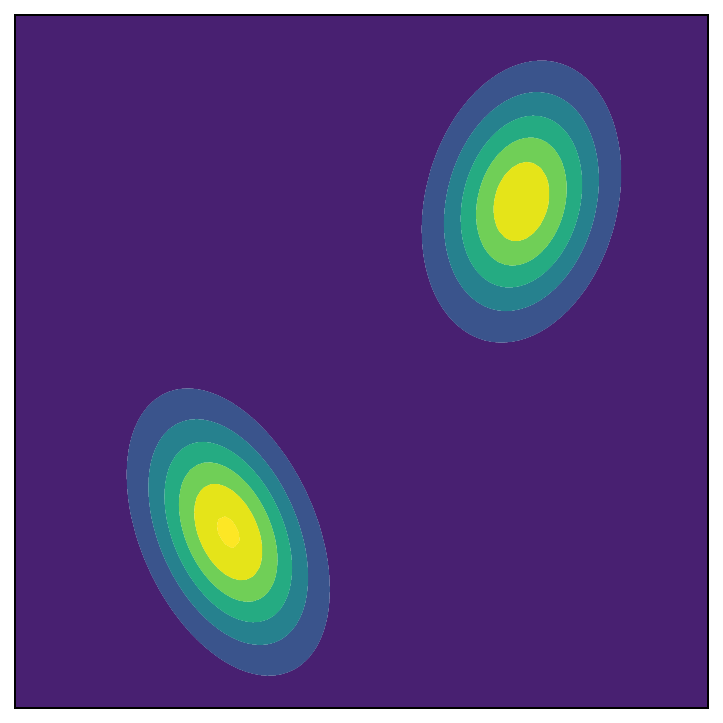}
&
\includegraphics[scale=0.17]{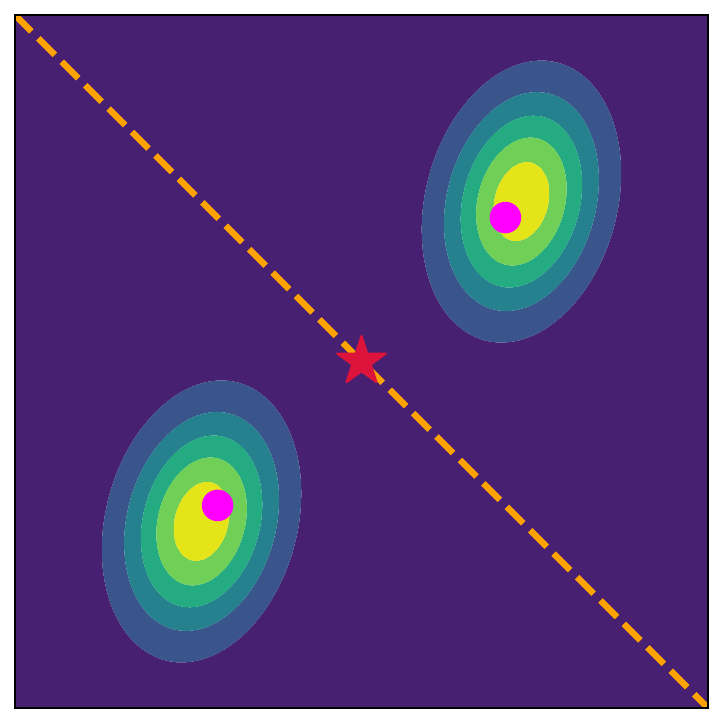}
&
\includegraphics[scale=0.17]{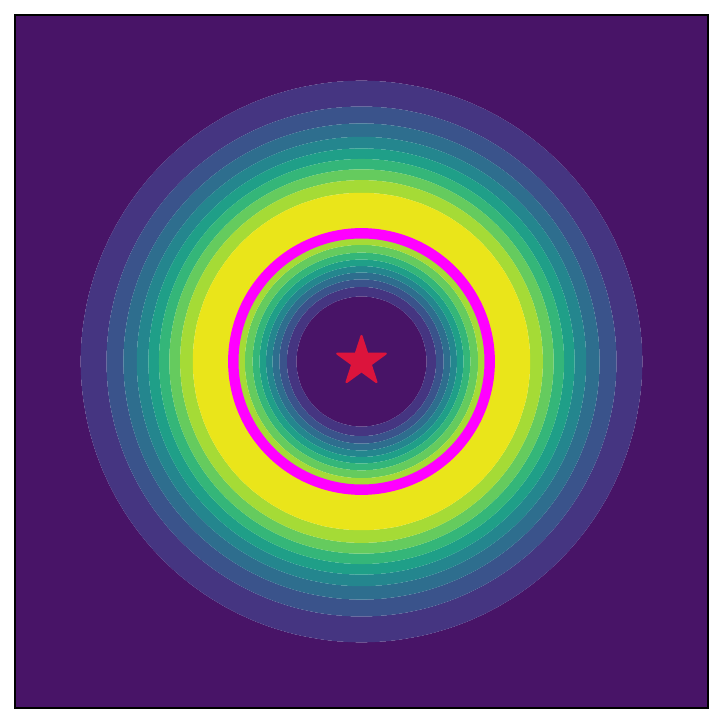}
&
\includegraphics[scale=0.17]{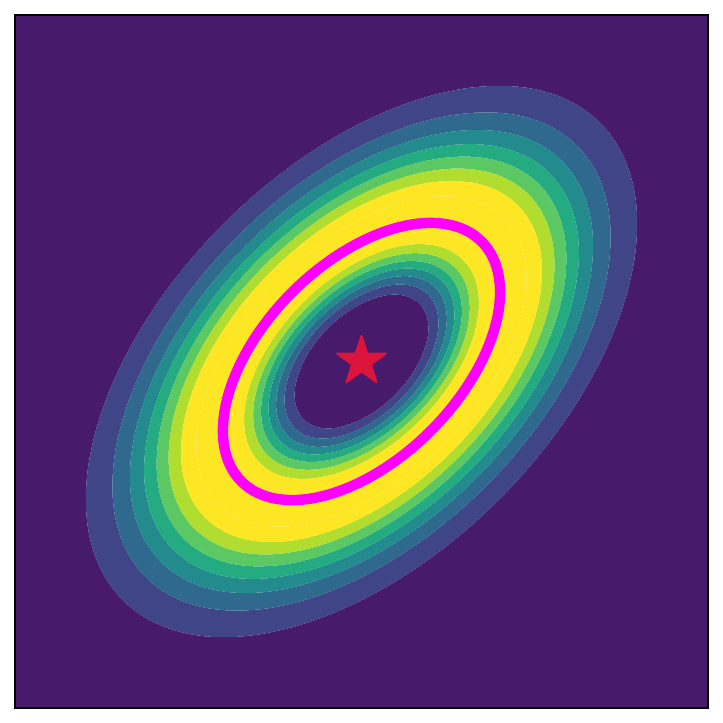}
\end{tabular}
}
\caption{\emph{\textbf{Illustration of the symmetries enabling our theoretical results}.} A point of \emph{even symmetry} is illustrated via a red star, and the orange dashed line illustrates $f(x)=-x$, along which points are mirrored. The \emph{spherically symmetric} ring  was transformed into the \emph{elliptically symmetric} ellipse via \cref{eq:p_ell}. Points (and sets of points) in pink attain the same density values within the respective figure.} \label{fig:symmetry_illustration}
\end{figure*}

We consider a standard variational inference setting, where given a target $p$, we aim to find the best tractable surrogate within the variational family, $q^{\star} \in \mathcal{Q}$ by optimizing a divergence of choice $D$:
\begin{equation}\label{eq:vi_setup}
q^{\star} = \text{argmin}_{q \in \mathcal{Q}}\, D(p, q).
\end{equation}

Our analysis focuses on $\alpha$-DIVs \citep{cichocki2010families}, defined as 
\begin{align}\label{def:alpha_main_paper}
\tag{$\alpha$-DIV}
D_\alpha(p, q):= \frac{1}{\alpha(\alpha-1)} \int \left[\left(\frac{p(\vx)}{q(\vx)}\right)^{\alpha} -1\right]q(\vx)d\vx,
\end{align}
with $\alpha > 0$ and $\alpha\neq 1$. The $\alpha$-DIV is a flexible family of divergences: For $\alpha=\frac{1}{2}$, we obtain a divergence that is proportional to the squared Hellinger distance \citep{le2000asymptotics}, while $\alpha=2$ recovers the $\chi^2$-divergence \citep{csiszar1967information,dieng2017variational}. In the limiting case of $\alpha \rightarrow 0$, \cref{def:alpha_main_paper} recovers the RKL, $D_{\text{RKL}}= \int \log\left(q(\vx)/p(\vx)\right)q(\vx)d\vx$, while $\alpha \rightarrow 1$ gives the FKL, $D_{\text{FKL}}= \int \log\left(p(\vx)/q(\vx)\right)p(\vx)d\vx$. Following \citet{margossian2024variational, margossian2025generalized}, we study \emph{location-scale families} as variational families.

\begin{definition}[Location-Scale Families]\label{def:loc_scale}
A member of a location-scale family can be expressed as 
\begin{equation}\label{eq:loc_scale}
q_{\vnu, S}(\vx) = q_{\vzero}{\Big(}S^{-\frac{1}{2}}(\vx-\vnu){\Big)}|S|^{-\frac{1}{2}},
\end{equation}
where $\vnu\in \mathbb{R}^d$ is the location parameter and $S\in \mathbb{S}_{++}^d$ is a positive definite scale matrix. A \emph{location family}, denoted by $q_{\vnu}$, only varies $\vnu$ while $S$ is fixed and equal for all members of the family.
\end{definition}
Location-scale families include popular distributions, such as Gaussian, Laplace, and Student-t distributions. For most location-scale families, $\vnu$ corresponds to the mean of $q_{\vnu}$, but this is not always the case. For instance, the mean of a Cauchy distribution is undefined, and $\vnu$ corresponds to its median (and mode) instead. For our analysis, $S^{\frac{1}{2}}$ denotes the principal square root. We use $\supp(p) = \{\vx \in \mathbb{R}^d: p(\vx) > 0\}$ to denote the \emph{support of a density} $p$. Akin to \citet{margossian2024variational, margossian2025generalized}, our analysis places symmetry conditions on $q_{\vzero}$ and the target $p$, which we illustrate in \cref{fig:symmetry_illustration}, and formally introduce next. 

\begin{definition}[Even Symmetry]\label{def:even_odd}
We say a function $h: \mathbb{R}^d \rightarrow \mathbb{R}$ is \textbf{even symmetric} around $\vnu \in \mathbb{R}^d$ if for all $\vx \in \mathbb{R}^d$,
$h(\vnu + \vx)= h(\vnu - \vx).$
\end{definition}

\cref{fig:symmetry_illustration} illustrates even symmetry on the example of a mixture distribution. For every point on the left-hand side of the orange dashed line, there exists a ``mirrored'' point on the opposite side of the line with the exact same density value. 
We have now introduced all ingredients for \cref{assumptions_mean}, the core assumptions underlying the results by \citet{margossian2025generalized} and our analysis for \textbf{\emph{exact recovery of the target mean}}.  The result of \citet{margossian2025generalized} is stated for general $f$-divergences, defined as $D_f(p, q)=\int f({p(\vx)}/{q(\vx)})q(\vx)d\vx$, which subsume the RKL, FKL, and $\alpha$-DIV for particular choices of the convex function $f$ (see \cref{app:f_div} for details). Under \cref{assumptions_mean} and mild regularity conditions on $p$ and $q$, \citet{margossian2025generalized} arrive at \cref{theorem:charles_f}. 
\begin{restatable}{assumption}{ASSUMPTIONSM}
\label{assumptions_mean}
Let $q_{\vnu}$ be a member of a location family with fixed scale matrix $S$ and let $p$ be the target. We make the following assumptions: The base distribution $q_{\vzero}$ is even symmetric around $\vzero$. The target $p$ is even symmetric around its mean $\vmu$. Lastly, 
$D_{f}(p, q_{\vnu})$ is well-defined for all  values of $\vnu$, and $\vmu$ is finite.
\end{restatable}

\begin{theorem}[Mean recovery with the RKL \citep{margossian2025generalized}]\label{theorem:charles_f} 
Under \ref{assumptions_mean}, a stationary point of $D_f(p,q_{\vnu})$ occurs at $\vnu = \vmu$.
Furthermore, if $\varphi(\vv) := f \circ \exp(\vv)$ is convex and
strictly decreasing, and $p$ somewhere-strictly log
concave over $\mathbb{R}^d$, then $\vnu = \vmu$ is a unique minimizer of $D_f(p,q_{\vnu})$.
\end{theorem}

\cref{theorem:charles_f} guarantees a \emph{stationary point} of $D_{f}(p, q_{\vnu})$ at $\vnu=\vmu$. Notably, this partial result 
poses no additional requirements on $\varphi$, and hence applies to general $f$-divergences, such as the FKL and $\alpha$-DIV. The sufficient conditions guaranteeing a \emph{unique minimizer} are more restrictive: On the one hand, (somewhere strict) log-concavity of $p$ implies a unimodal target. Moreover, the requirements on $\varphi$ are only fulfilled by the RKL among the $f$-divergences considered by \citet{margossian2025generalized}. Nevertheless, they found that the remaining $f$-divergences often located the correct mean in practice, even for targets that violate log-concavity, such as the Student-t distribution (cf. Fig. 2 in \citet{margossian2025generalized}). In \cref{sec:mean_recovery}, we provide \emph{complementary sufficient conditions for exact recovery of the mean with the FKL and $\alpha$-DIV}. Notably, our results do not assume $p$ to be log-concave.

In order to additionally guarantee \emph{\textbf{exact recovery of correlations}}, we need to place a stronger set of assumptions on both the target and the variational family, which we summarize as \cref{assumptions_corr}. First, we introduce the concepts of \textbf{\emph{spherical and elliptical symmetry}}.
\begin{definition}[Spherical and Elliptical Symmetry]\label{def:ellip}
We say a function $h: \mathbb{R}^d \rightarrow \mathbb{R}$ is \textbf{spherically symmetric} if $h(\vx) = h(\vx')$ whenever $||\vx||=||\vx'||$. We further say $h$ is \textbf{elliptically symmetric} if there exists a positive definite matrix $M \in \mathbb{S}_{++}^d$ and a vector $\vmu \in \mathbb{R}^d$, such that $h(\vzeta)$ with $\vzeta = M^{-\frac{1}{2}}(\vx - \vmu)$ is spherically symmetric.
\end{definition} 

Conveniently, whenever $p$ is elliptically symmetric, we can express it in terms of a spherically symmetric base density $p_{\vzero}$ and a positive definite matrix $M \in \mathbb{S}_{++}^d$, which is directly related to the correlation matrix of $p$ via $\text{Corr}_p[x_i, x_j]= M_{ij}/\sqrt{M_{ii}M_{jj}}$, 
\begin{align}
p(\vx) = p_\vzero\left(M^{-\frac{1}{2}}(\vx-\vmu)\right)|M^{-\frac{1}{2}}| \label{eq:p_ell}.
\end{align}
Moreover, since $p_\vzero$ is spherically symmetric, there exists a function $g: \mathbb{R}^+_0 \rightarrow \mathbb{R}^+_0$, such that $g(||\vx||) = p_{\vzero}(\vx)$. Note that elliptical symmetry around a point $\vmu$ implies even symmetry around $\vmu$. Hence, \cref{assumptions_corr} below entails \cref{assumptions_mean}. Elliptical symmetry is, for instance, present in multivariate Gaussian and Student-t distributions, but we can also construct more intricate shapes fulfilling the property, such as the elliptic density in \cref{fig:symmetry_illustration}, by choosing a more expressive $p_{\vzero}$. By exploiting \cref{eq:p_ell}, \citet{margossian2025variational} arrive at \cref{theorem:charles_f_correlations} under conditions \cref{assumptions_corr}.
\begin{restatable}{assumption}{ASSUMPTIONSC}
\label{assumptions_corr}
Let $q_{\vnu, S}$ be a member of a location-scale family and let $p$ be the target. We make the following assumptions: The base distribution $q_{\vzero}$ is spherically symmetric. The target $p$ is elliptically symmetric with $M \in \mathbb{S}_{++}^d$ and mean $\vmu \in \mathbb{R}^d$. 
Lastly, $D_{f}(p, q_{\vnu, S})$ is well-defined and finite for all considered values of $\vnu$ and $S$.
\end{restatable}

\begin{theorem}[Mean and correlation recovery with the RKL \citep{margossian2025generalized}]\label{theorem:charles_f_correlations}
Assume \cref{assumptions_corr} and \textbf{additionally} that (i) $p$ is somewhere strictly log-concave over $\mathbb{R}^d$ and (ii) $\varphi(\vu) = f \circ\exp(\vu)$ is convex and strictly decreasing and (iii) $g(\vx) = p_\vzero(||\vx||)$ is everywhere continuously differentiable with $|g'(0)| < \infty$. Then, $D_f (p, q_{\vnu, S})$ has a unique minimizer with respect to the location-scale parameters $\vnu, S$ at $\vnu = \vmu$ and $S = \gamma^2M$ for some $\gamma > 0$. 
\end{theorem}
First, we remark that despite the similarities of \cref{eq:p_ell} and \cref{eq:loc_scale}, \cref{theorem:charles_f_correlations} allows for  misspecification of the variational family since the base distributions $q_{\vzero}$ and $p_{\vzero}$ can differ. Once again, however, \cref{theorem:charles_f_correlations} is limited to the RKL and log-concave targets. 
In \cref{sec:corr_recovery}, we provide sufficient conditions on $p$ and $\mathcal{Q}$ under which we can guarantee simultaneous exact recovery of the target mean and correlations with the FKL and $\alpha$-DIV. 

\section{Guarantees for the FKL and $\alpha$-divergence}\label{sec:our_gaurantees}
We now state our core theoretical results. 
The full proofs are given in \cref{app:proofs}. 
Notably, none of our theorems requires log-concavity of $p$, hence, they apply to \emph{multi-modal} and \emph{heavy-tailed} targets. 
Our conditions primarily concern the \emph{base distribution}  (\cref{def:loc_scale}). See \cref{tab:dist-sym} for a summary of common location-scale distributions, and whether their base distributions fulfill properties related to our sufficient conditions.

\subsection{Exact recovery of the mean}\label{sec:mean_recovery}
In this section, we state our results for exact mean recovery for the FKL and $\alpha$-\text{DIV} for location families. Throughout the analysis, we treat the scale parameter $S$ as fixed. When the conditions in our theorems are met, the target mean will be recovered for \emph{any} assignment of $S$. As a result, our guarantees for exact mean recovery also apply when the variational family is further restricted to a simple \emph{mean-field approximation} \citep{blei2017variational, xing2012generalized}, which is widely used practice for computational efficiency.

We begin by analyzing the \textbf{FKL}. First, note that optimizing the FKL w.r.t. the location parameter $\vnu$ results in the following objective:
\begin{align}\tag{FKL}
D_{\text{FKL}}(p, q_{\vnu})=\int\log{\Big(}\frac{p(\vx)}{q_{\vnu}(\vx)}\Big{)}p(\vx)d\vx = \int-\log {\Big(}q_{\vzero}{\Big(}S^{-\frac{1}{2}}(\vx-\vnu){\Big)}{\Big)}p(\vx)d\vx + C,\label{eq:FKL}
\end{align}
where we plug in the definition of a member of a location family (\cref{def:loc_scale}) and $C$ is an additive constant in $\vnu$ (see \cref{app:f_div} for details).
Viewing the FKL in this form gives rise to our \cref{theorem:theorem_fkl_1}.

\begin{restatable}{theorem}{FKLSC}
\label{theorem:theorem_fkl_1}
Under \cref{assumptions_mean}, $D_{\text{FKL}}(p, q_{\vnu})$ has a unique minimizer at $\vnu = \vmu$ if \wkl is  strictly convex.
\end{restatable}

Intuitively, the above condition ensures that $D_{\text{FKL}}(p, q_{\vnu})$ is strictly convex in $\vnu$. Combined with the insight from \cref{theorem:charles_f} that $D_{\text{FKL}}(p, q_{\vnu})$ has a stationary point at $\vnu=\vmu$, we conclude that $D_{\text{FKL}}(p, q_{\vnu})$ has a unique minimizer at $\vnu=\vmu$. Interestingly, \citet{margossian2024variational} already hinted at the strict convexity of general integral expressions akin to \cref{eq:FKL} in their Proposition 12, but to the best of our knowledge never used this result to state the existence of a unique minimizer for divergences besides the RKL. \cref{theorem:theorem_fkl_1} also entails the classical result that
the optimal Gaussian approximation w.r.t. the FKL will match the mean of the target \citep{wainwright2008graphical}. However, it is more widely applicable: For instance, also other location-scale families such as the \emph{logistic distribution} are covered by \cref{theorem:theorem_fkl_1} (see \cref{app:loc_scale_families} for the definition and visualization). 
Next, we state another result for the FKL that allows us to \emph{relax strict convexity} of \wkl at the cost of placing a condition on the support of $p$.
\begin{restatable}{theorem}{FKLC}
\label{theorem:theorem_fkl_2}
Under \cref{assumptions_mean}, $D_{\text{FKL}}(p, q_{\vnu})$ has a unique global minimizer at $\vnu = \vmu$ if (1) \wkl is convex and strictly increasing in $||\vx||$ and (2) \csupp.
\end{restatable}

Compared to \cref{theorem:theorem_fkl_1}, \cref{theorem:theorem_fkl_2} places different assumptions on both $p$ and $q$: A Laplace distribution can fit into the framework of \cref{theorem:theorem_fkl_2} but not \cref{theorem:theorem_fkl_1}. However, \cref{theorem:theorem_fkl_2} tells us that VI with a Laplace family can fail to obtain a unique minimizer at $\vmu$ when $p$ is not supported around $\vmu$. In \cref{sec:simulations}, we illustrate this phenomenon on a concrete example targeting a mixture of uniforms with disjoint support. 

The condition that \wkl is strictly increasing in $||\vx||$ is fulfilled by most standard location-scale families, as shown in \cref{fig:wkl_curves}. We note however that it is possible to construct functions $f: \mathbb{R}^d\rightarrow \mathbb{R}$ such that $f$ is even symmetric and convex, but \emph{not} strictly increasing in $||\vx||$, such as $f(x_1, x_2) = x_1^4 + x_2^6$. 
At the same time, not every spherically symmetric function is strictly increasing in $||\vx||$. For instance, the \emph{uniform distribution} is not covered by our guarantees. We now provide intuition as to why \cref{theorem:theorem_fkl_2} holds.
\begin{figure}[t!]
\begin{minipage}{.5\textwidth}
    \begin{tabular}{cc}
\includegraphics[scale=0.24]{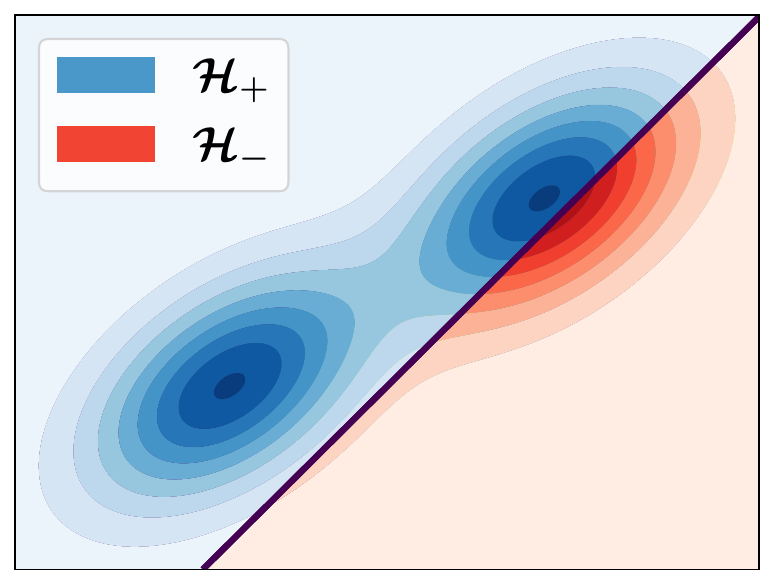}&
\includegraphics[scale=0.24]{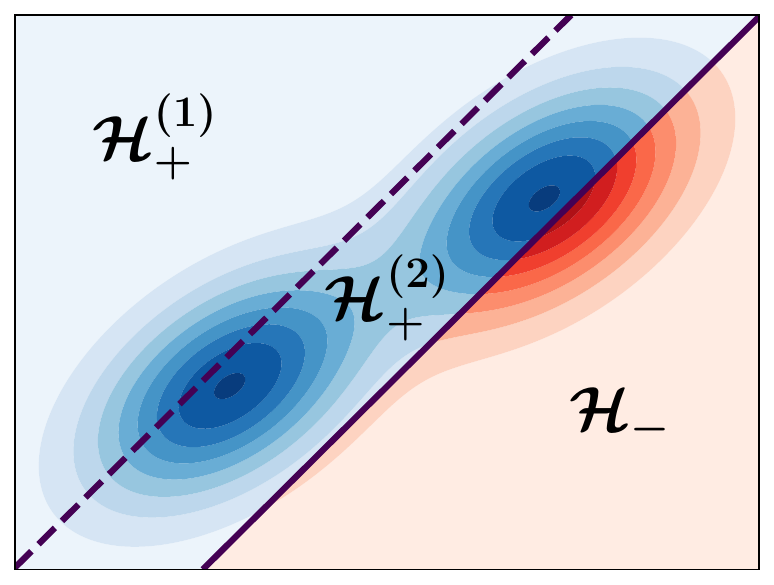}
\end{tabular}
\end{minipage}\hfill\begin{minipage}{.48\textwidth}
    \caption{For \cref{theorem:theorem_fkl_2} and \cref{theorem:theorem_alpha_2}, we exploit the symmetry of $p(\vmu+\vtau)$ by splitting its domain  into $\mathcal{H}_+$ (blue) and $\mathcal{H}_-$ (red), where $\Delta_{\text{FKL}}$ (\cref{eq:delta_FKL}) \emph{increases} over the first and \emph{decreases} over the second (L). We further partition $\mathcal{H}_+$ into $\mathcal{H}_+^{(1)}$ and $\mathcal{H}_+^{(2)}$, where $\mathcal{H}_+^{(1)}$ mirrors $\mathcal{H}_-$ (R). We set $\nu'=(1.5, -0.9)$ and $q_\vzero$ as a standard Gaussian.}\label{fig:proof_sketch}
\end{minipage}
\begin{center}
\end{center}
\end{figure}

\textbf{Intuition for Theorem \ref{theorem:theorem_fkl_2}.} To prove \cref{theorem:theorem_fkl_2} in \cref{app:proofs}, we show directly that every alternative assignment to $\vnu =\vmu$ must result in a strictly larger value of $D_{\text{FKL}}$. For brevity, let  $w:= \wkl$, $\vnu' \neq \vzero$ and $\vtau := \vx-\vmu$. Consider the \emph{difference in $D_{\text{FKL}}$} (\cref{eq:FKL}) \emph{for alternative assignments of the location parameter} $\vnu=\vmu$ and $\vnu = \vmu +\vnu'$:
\begin{align}
\Delta_{\text{FKL}}:=D_{\text{FKL}}(p, q_{\vmu+\vnu'}) - D_{\text{FKL}}(p, q_{\vmu})=&\int {\Big[}w(S^{-\frac{1}{2}}(\vtau-\vnu')) - w(S^{-\frac{1}{2}}\vtau){\Big]}p(\vmu+\vtau)d\vtau.  \label{eq:delta_FKL}
\end{align}
Remember that $w(\vz)$ is assumed strictly increasing in $||\vz||$.
Therefore, we can distinguish points that increase $\Delta_{\text{FKL}}$ from those that decrease $\Delta_{\text{FKL}}$ via the condition $||S^{-\frac{1}{2}}(\vtau+\vnu')|| > ||S^{-\frac{1}{2}}\vtau||$. We denote the resulting sets of points, which form half-spaces for strictly radially increasing $w$, by $\mathcal{H}_+$ and $\mathcal{H}_-$ respectively (see \cref{fig:proof_sketch} (L)). 
We further split $\mathcal{H}_+$ into disjoint sets $\mathcal{H}_+^{(1)}$ and $\mathcal{H}_+^{(2)}$, such that $\mathcal{H}_+^{(1)}$ ``mirrors'' $\mathcal{H}_-$: for every $\vtau \in \mathcal{H}_+^{(1)}$, it holds that $-\vtau \in \mathcal{H}_-$, and vice versa (see \cref{fig:proof_sketch} (R)). Moreover, by even symmetry of $p$ around $\vmu$, we have $p(\vmu + \vtau) = p(\vmu - \vtau)$. When convexity is not strict, it can happen that the contributions of $\mathcal{H}_+^{(1)}$ and $\mathcal{H}_-$ to $\Delta_{\text{FKL}}$ cancel. In such cases, we need to rely on $\mathcal{H}_+^{(2)}$ to ensure $\Delta_{\text{FKL}} > 0$. When $p$ is not supported around $\vmu$ on a set of positive measure, the contribution of $\mathcal{H}_+^{(2)}$ to $\Delta\text{FKL}$ can be exactly $0$, and hence the minimizer might not be unique. In \cref{fig:exps_mean}, we show a practical example of this case, where a Laplace proposal (for which \wkl is convex but not strictly) fails to distinguish between alternative assignments of $\vnu$ due to a lack of support around the true mean.

Our construction also provides a different way to prove the existence of stationary points than 
\cref{theorem:charles_f}. Moreover, it sheds further light on \cref{theorem:theorem_fkl_1}: When $\wkl$ is strictly convex, the contribution of $\mathcal{H}_{+}^{(1)}$ to $\Delta\text{FKL}$ will be strictly greater than the decrease induced by $\mathcal{H}_{-}$. Hence, we do not need to rely on $\mathcal{H}_+^{(2)}$ to ensure a strictly positive value of $\Delta\text{FKL}$ in \cref{theorem:theorem_fkl_1}.

By the same line of reasoning, we arrive at analogous results for the $\boldsymbol{\alpha}$\textbf{-DIV} (\cref{def:alpha_main_paper}), which for fixed $S$ simplifies to
\begin{align}
D_{\alpha}(p, q_{\vnu}) 
&= |S|^{-\frac{1-\alpha}{2}} \int \frac{1}{\alpha(\alpha-1)}q_{\vzero}^{1-\alpha}{\Big(}S^{-\frac{1}{2}}(\vx-\vnu){\Big)}p^{\alpha}(\vx)d\vx + C,\label{eq:a_div_proof}
\end{align}
where $\alpha > 0$ and $\alpha \neq 1$ and $C$ again denotes a constant in $\vnu$. Note that $|S|^{-\frac{1-\alpha}{2}}$ is positive and fixed, and $p^{\alpha}$ preserves non-negativity, support, and even symmetry of $p$ around $\vmu$. Hence, \cref{eq:a_div_proof} closely resembles \cref{eq:FKL} in structure. In fact, we find that replacing the requirements on \wkl in \cref{theorem:theorem_fkl_1,theorem:theorem_fkl_2} with the same conditions on \wa gives rise to more general statements for the $\alpha$-DIV as detailed below.
\begin{restatable}{theorem}{ASC}
\label{theorem:theorem_alpha_1}
Under \cref{assumptions_mean}, $D_{\alpha}(p, q_{\vnu})$ has a unique global minimizer at $\vnu = \vmu$ if \wa is strictly convex.
\end{restatable}

\begin{restatable}{theorem}{AC}
\label{theorem:theorem_alpha_2}
Under \cref{assumptions_mean}, $D_{\alpha}(p, q_{\vnu})$ has a unique global minimizer at $\vnu = \vmu$ if \wa is convex, strictly increasing in $||\vx||$ and (2) \csupp.
\end{restatable}

In practice, we find that the requirements on \wa are governed by the chosen $\alpha$ more so than $q_{\vzero}$. \cref{fig:exps_mean} gives examples where the same location-scale family succeeds and fails to recover the target mean depending on the value of $\alpha$. As a general rule of thumb, $\alpha > 1$ is favorable for exact mean recovery. \cref{fig:wa_curves} shows \wa for varying values of $\alpha$ for common location-scale families. Notably, even for distributions for which $q_{\vzero}$ itself is not log-concave, such as the Student-t, it can be possible to find values of $\alpha$, such that our sufficient conditions are fulfilled.

\textbf{Partial symmetries.} In practice, the target might only be symmetric  along a \emph{subset} of variables. A prime example for such a density is Neal's Funnel \citep{neal2003slice} but similar posterior geometries can also arise naturally in general hierarchical Bayesian models (see \citep{margossian2025generalized} for a detailed discussion). 
In \cref{app:partial}, we show that our results for mean recovery can be extended to the setting where $p$ is only partially even symmetric. In this scenario, under  similar conditions as outlined in \cref{theorem:theorem_fkl_1,theorem:theorem_fkl_2,theorem:theorem_alpha_1,theorem:theorem_alpha_2}, we can guarantee exact mean recovery along the symmetric dimensions if we additionally assume a block-diagonal structure of $S$, separating symmetric and non-symmetric variables.

\subsection{Exact recovery of correlations}\label{sec:corr_recovery}

We now present our results for exact recovery of the correlation matrix. The statements closely resemble the results of \citet{margossian2024variational} for the RKL, but we are once again able to alleviate the assumption that $p$ is (somewhere strictly) log-concave. While we still make the assumption that $p$ is elliptically symmetric, dropping log-concavity opens up a wider range of densities, including heavy-tailed targets or targets with deep valleys around their mean (\cref{fig:results_simulations}, third row). This allows us to give guarantees for a wider range of misspecifications. The full proofs are given in \cref{app:corr}. Remember that below $M$ is directly related to the correlation matrix of $p$, as described in \cref{eq:p_ell}.

\begin{restatable}{theorem}{FKLCORR}
\label{theorem:theorem_fkl_corr}
Assume \cref{assumptions_corr}. Moreover, assume that all conditions for either  \cref{theorem:theorem_fkl_1} or \cref{theorem:theorem_fkl_2} are satisfied, and hence we are guaranteed to recover the target mean. If additionally $g(||\vx||) = \log q_\vzero(\vx)$ is almost everywhere continuously differentiable with $|g'(0)| < \infty$ or \limc, then $D_{\text{FKL}}(p, q_{\vnu, S})$ has a unique minimizer at $(\vmu, \gamma^2M)$ for some $\gamma >0$.
\end{restatable}

\begin{restatable}{theorem}{ACORR}
\label{theorem:theorem_alpha_corr}
Assume \cref{assumptions_corr}. Moreover, assume that all conditions for either \cref{theorem:theorem_alpha_1} or \cref{theorem:theorem_alpha_2} are satisfied, and hence we are guaranteed to recover the target mean. If additionally (i) $g(||\vx||) = q_\vzero^{1-\alpha}(\vx)$ is almost everywhere continuously differentiable with $|g'(0)| < \infty$  or \limca, (ii) $\alpha > 1$, and (iii) \wkl is convex, 
then $D_{\alpha}(p, q_{\vnu, S})$ has a unique minimizer at $(\vmu, \gamma^2M)$ for some $\gamma >0$.
\end{restatable}

Notably, our result for the \textbf{FKL} \cref{theorem:theorem_fkl_corr} does not require us to put strong additional assumptions on the variational family. The conditions on $g$ are fulfilled for most common location scale families. Note that we replace the condition $|g'(0)| < \infty$  with \limc for densities that do not have a well-defined derivative at the origin. Since our analysis relies on the derivative being increasing and non-negative, the singularity at the origin does not affect our results. 

\begin{figure}
\resizebox{\textwidth}{!}{
\begin{tabular}{ccccc}
& FKL & FKL & $\alpha = 1.1$ & $\alpha = 0.3 $\\
Target ($p_1$) & Gaussian (\cref{theorem:theorem_fkl_1} \yes) & Laplace (\cref{theorem:theorem_fkl_1} \no) & Gaussian (\cref{theorem:theorem_alpha_1} \yes) &  Gaussian (\cref{theorem:theorem_alpha_1} \no)\\
\includegraphics[scale=0.2]{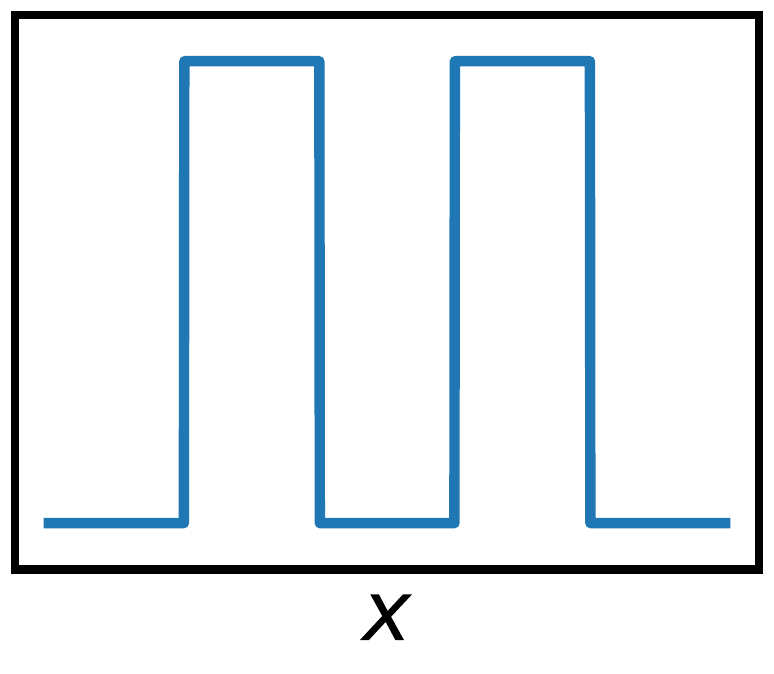}
& 
\includegraphics[scale=0.2]{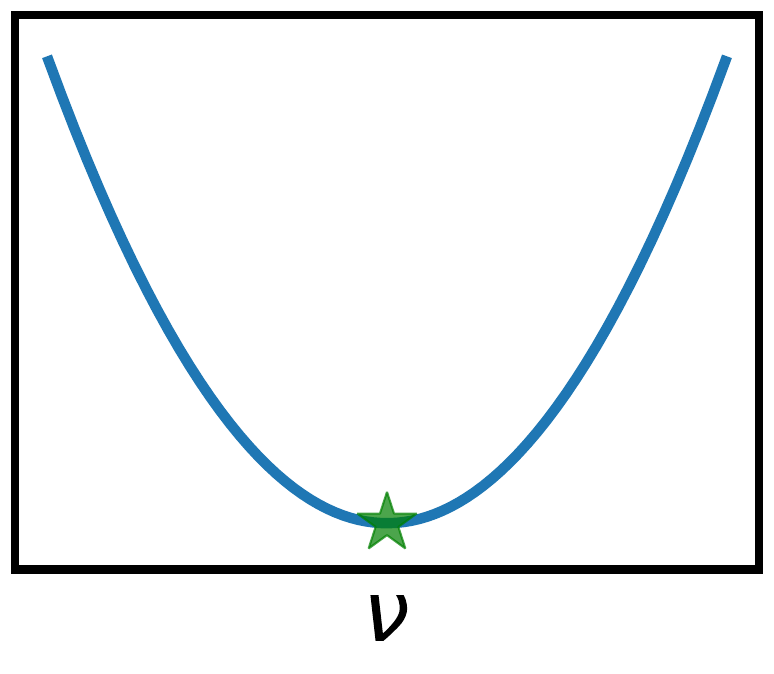}
&
\includegraphics[scale=0.2]{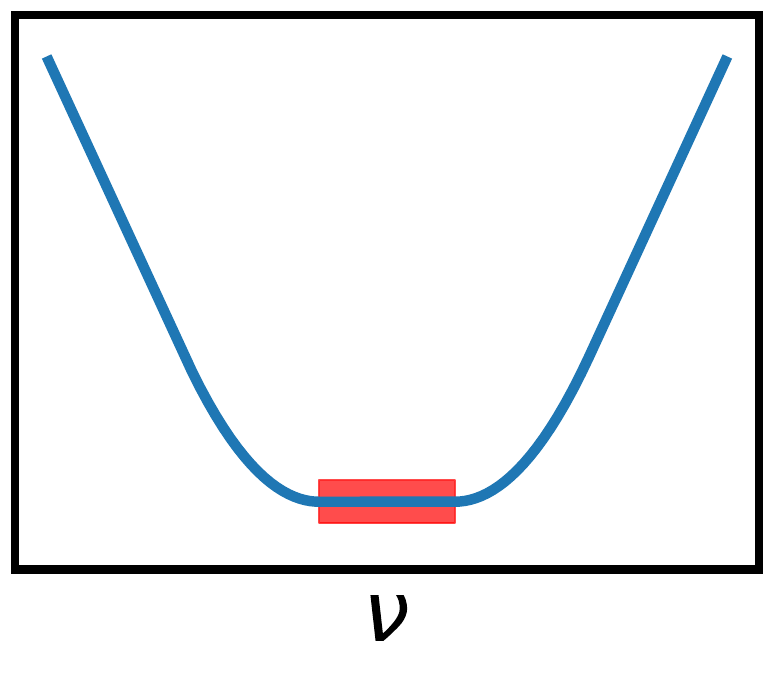}
&
\includegraphics[scale=0.2]{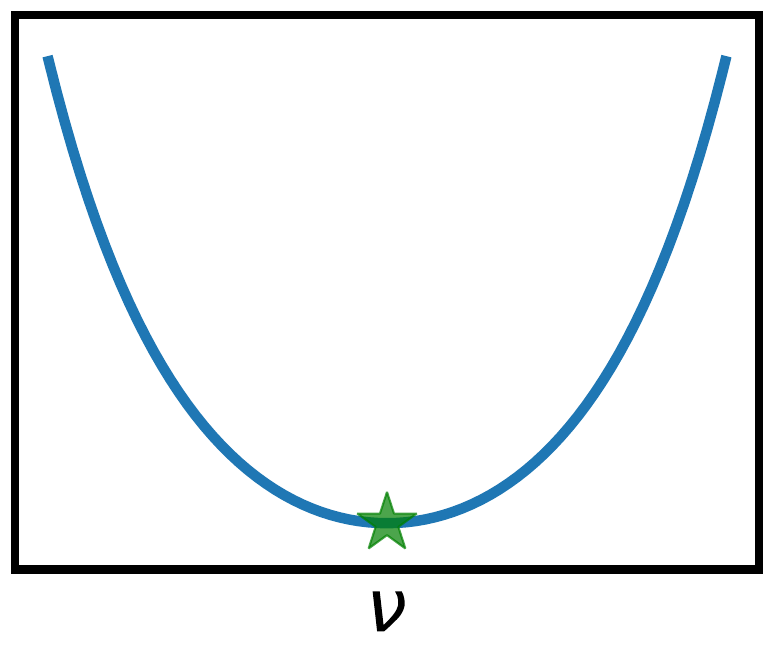} &
\includegraphics[scale=0.2]{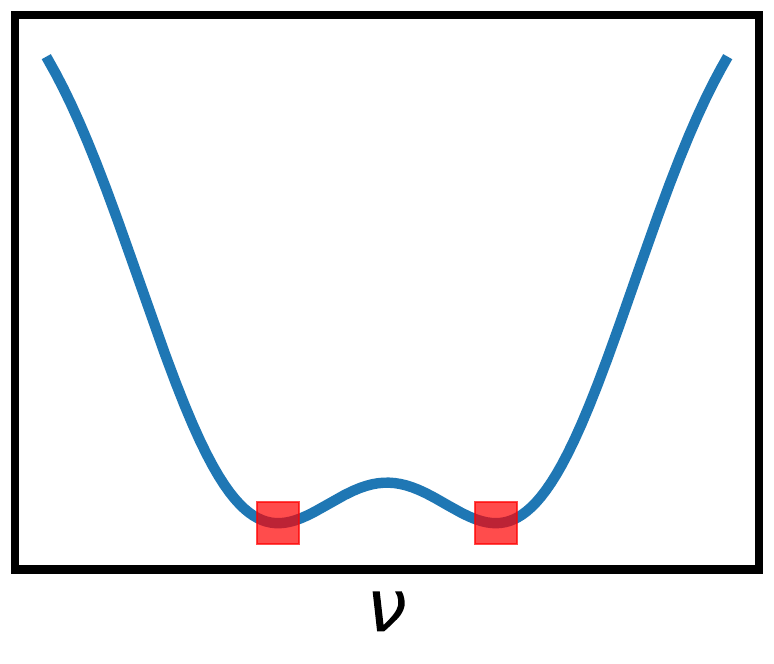}
\\
& FKL & FKL & $\alpha = 2.0$ & $\alpha =0.5$\\
Target ($p_2$) & 
Laplace (\cref{theorem:theorem_fkl_2} \yes) &
Student-t (\cref{theorem:theorem_fkl_2} \no) & Student-t (\cref{theorem:theorem_alpha_2} \yes) & Student-t (\cref{theorem:theorem_alpha_2} \no)\\
\includegraphics[scale=0.2]{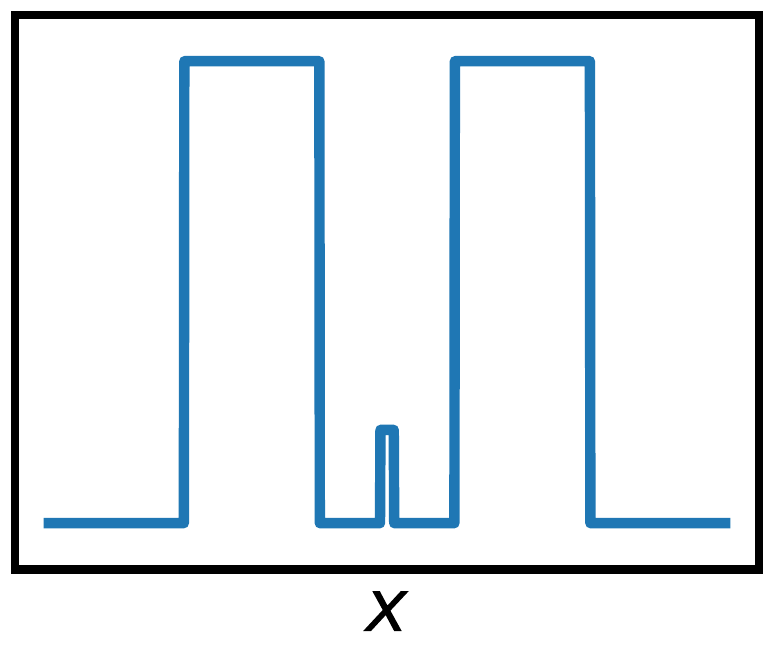} &
\includegraphics[scale=0.2]{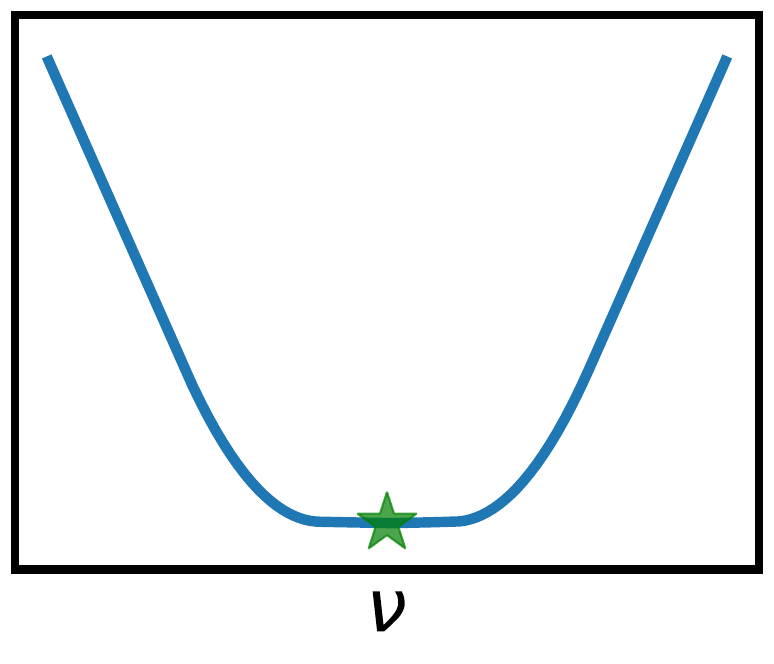} &
\includegraphics[scale=0.2]{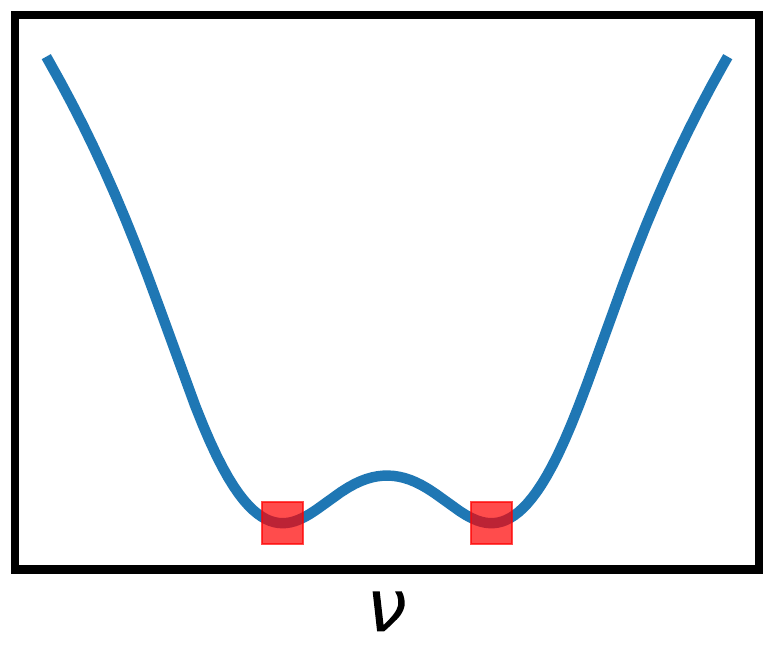} &
\includegraphics[scale=0.2]{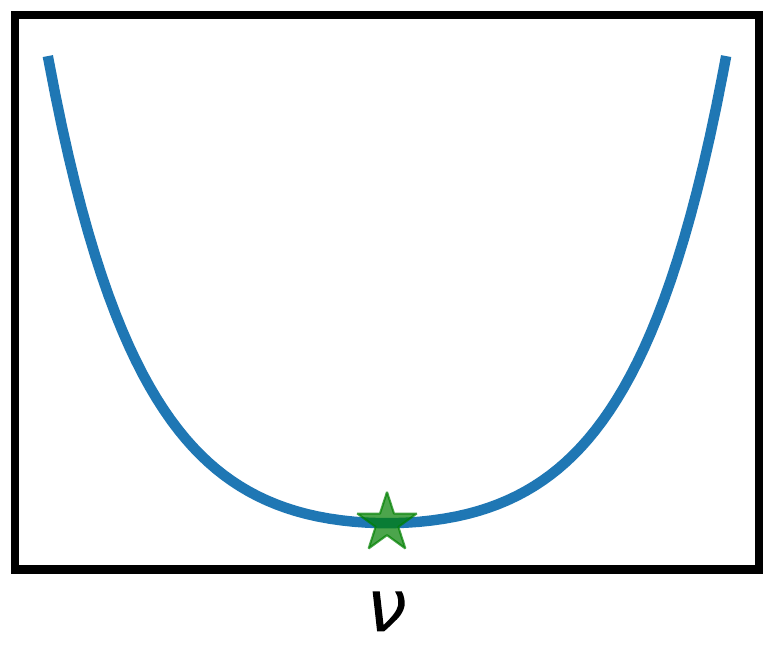} &
\includegraphics[scale=0.2]{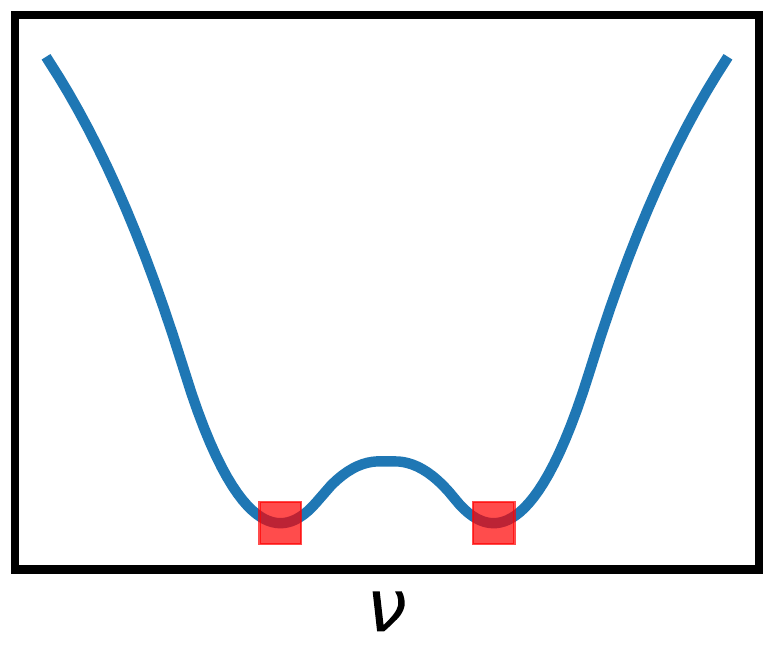}
\end{tabular}
}
\caption{\textbf{Our sufficient conditions guarantee a unique global optimum at the true mean of the target. When they are violated, optimization may fail to locate the correct mean.} Illustration of settings that \emph{comply with} (\yes) and \emph{violate} (\no) the sufficient conditions provided by our theorems for mean recovery (\cref{sec:mean_recovery}). The first figure depicts the target density $p$. The remaining figures show the divergence between $p$ and $q_{\vnu}$ as a function of $\vnu$. When our sufficient conditions are met, we expect a unique global minimizer of the divergence at the targets true mean (\emph{indicated via a green star}). If our conditions are violated, we cannot guarantee the former (\emph{indicated via a red square/bar}). The concrete settings are described in \cref{sec:simulations_analytic}. We use $5$ degrees of freedom for the Student-t.
}\label{fig:exps_mean}
\end{figure}

For the $\boldsymbol{\alpha}\text{\textbf{-Div}}$, we again obtain a more fine-grained criterion. However, compared to \cref{theorem:theorem_alpha_1} and \cref{theorem:theorem_alpha_2}, we now explicitly decouple conditions on $\alpha$ and $q_{\vzero}$. Note that these conditions are consistent with the requirements for mean recovery, albeit stronger: While an appropriately high value of $\alpha$ can guarantee exact mean recovery for a Student-t (as we also show empirically in \cref{fig:exps_mean}), we do not provide a guarantee for exact correlation recovery with the Student-t.

\section{What happens in practice}\label{sec:simulations}
To illustrate the practical implications of our derived sufficient conditions, we now present examples where our theoretical results can aid the choice of variational family and $\alpha$ value. Experimental details and target specifications are in \cref{app:simulations_setup}.

\subsection{Symbolic integration for mean recovery}\label{sec:simulations_analytic}
We first consider scenarios for which the exact computation of the divergence is tractable. We construct the following \emph{targets}: We define $p_1$ as a 
one-dimensional, symmetric mixture of uniform distributions, such that $X \sim0.5 \cdot \text{Unif}(X; [-9, -3])+0.5\cdot\text{Unif}(X; [3, 9])$. The mean of this distribution ($\mu=0$) is not contained in the support of $p$, i.e., $\mu \notin \text{supp}(p)$. To create a target $p_2$ for which $\mu \in \text{supp}(p)$, we mix $p_1$ with an additional uniform centered at $0$: $X \sim 0.99 \cdot p_1(X) + 0.01\cdot\text{Unif}(X; [-0.3, 0.3])$. Our results are summarized in \cref{fig:exps_mean}. 

\textbf{\cref{theorem:theorem_fkl_1}.} We are guaranteed to obtain a unique minimizer of the FKL at $\nu =\mu$ \emph{if} we choose our variational family, such that \wkl is strictly convex. The variational family $\mathcal{Q}_1 = \mathcal{N}(\nu, 4)$ complies with this condition, and we obtain a unique minimizer at $\nu =\mu$ in practice. 
Next, we \emph{violate} the sufficient condition in \cref{theorem:theorem_fkl_1} by choosing $\mathcal{Q}_2 = \text{Laplace}(\nu, 4)$. We do not obtain a \emph{unique} global minimizer. In fact, the objective obtains the exact same value for every $\nu \in [-3, 3]$. \cref{app:case_1_2_details} provides the full mathematical explanation for this phenomenon. Intuitively, for $\nu \in [-3, 3]$, we are in the setting where the integral over $\mathcal{H}_+^{(2)}$ equals $0$ (see \cref{fig:proof_sketch}).

\textbf{\cref{theorem:theorem_fkl_2}.} As $p_2$ is supported around $\mu$, we do not require strict convexity of \wkl. As a result, we can reuse $\mathcal{Q}_2=\text{Laplace}(\nu, 4)$, which indeed gives a unique minimizer of the FKL for $p_2$. To showcase that we \emph{cannot} guarantee a unique minimizer when $-\log q_0$ is not convex, we use a Student-t variational family with $5$ degrees of freedom $\mathcal{Q}_3 = \text{Student}_5(\nu, 1)$. We observe that the FKL has a stationary point at $\mu=0$ as per \cref{theorem:charles_f} \citep{margossian2025generalized}, but it is a local maximizer.

\textbf{\cref{theorem:theorem_alpha_1}.} We now move to our guarantees for $\alpha$-DIVs. We use $\mathcal{Q}_1 = \mathcal{N}(\nu, 4)$. Note that $q_0$  \emph{itself} being strictly log-concave does not guarantee a unique minimizer; the criterion depends on the interplay between $q_{0}$ and the chosen value of $\alpha$. We compare  $\alpha=1.1$, for which \wao is strictly convex and optimization succeeds, and $\alpha=0.3$, which results in a function \wao that is not convex. The second case is not covered by any of our guarantees, and we obtain a local maximizer at $\mu=0$. Further values of $\alpha$ and visualizations of \wao are in \cref{app:extra_alpha_case_3}. 

\textbf{\cref{theorem:theorem_alpha_2}.} Lastly, we discuss \cref{theorem:theorem_alpha_2}, which relaxes \cref{theorem:theorem_alpha_1} in the sense that \wao is required to be convex, but not necessarily strictly, while imposing the additional condition that \csupp. We find that for common location families, it is usually not the case that \wao is convex but not strictly convex (see \cref{fig:wa_curves}). We nevertheless provide an example on the target $p_2$ with $\mathcal{Q}_3=\text{Student-t}(\nu, 4)$ to show that the Student-t can succeed to recover the mean with the $\alpha$-DIV when our conditions are satisfied, despite not being covered by any guarantee when used with the FKL. We show $\alpha=2.0$ and $\alpha=0.5$, which correspond to the $\chi^2$ divergence and squared Hellinger distance respectively. As expected, we obtain a unique minimizer at the correct mean for $\alpha=2.0$. For $\alpha=0.5$, we instead observe a a local maximum at $\mu$.

\subsection{Learning under misspecification}\label{sec:learning} 
We conduct experiments using a standard VI setup based on stochastic gradient descent (SGD). The aim of these experiments is to show that whenever our sufficient conditions are satisfied, we can recover the corresponding target statistics with low error, despite the noise introduced by stochastic optimization. We optimize the location and scale parameters simultaneously. 

\textbf{Setup.} All of our experiments in 2D use the Adam optimizer \citep{kingma2014adam} with default parameters and learning rate \lr. For the RKL and $\alpha$-DIV, we learn via reparameterization \citep{kucukelbir2017automatic} when the target has full support and use REINFORCE \citep{williams1992simple} otherwise. For optimizing the FKL, we use i.i.d. samples from $p$ (see \cref{app:simulations_setup} for the used sampling schemes).
We train for a maximum of $10^4$ steps using $10^5$ samples for each update. For the Student-t, we fix the degrees of freedom to $5$. The reported error metrics are computed w.r.t. the checkpoint that achieved the lowest training divergence. Following \citet{margossian2024variational}, we measure the error in the mean estimate as $\Delta\vmu:= \frac{1}{d}\sum_{j=1}^d|\nu_j-\mu_j|/\text{max}(\sqrt{\mathbb{V}_p[x_j]}, |\mu_j|)$ and the error of the correlation matrix as $\Delta\text{Corr}:= \frac{1}{d^2}\sum_{j=1}^d\sum_{i=1}^d|\text{Corr}_q[x_j, x_i]-\text{Corr}_p[x_j, x_i]|$. 

Below, we introduce a diverse set of targets for which we study VI under misspecification. Similar to \cref{sec:simulations_analytic}, we present examples in pairs of cases which are (1) (partially) covered by our guarantees and (2) not covered by any of our guarantees. 
\cref{fig:results_simulations} summarizes our insights.
We provide results for additional combinations of divergences and variational families in \cref{app:full_sgd_results}, including results for the RKL. \cref{tab:target_stats} gives an overview of our targets and their compatibility with our theorems.

\textbf{Results.} On a high level, we observe that whenever our theorems guarantee exact recovery of the mean or correlation matrix in theory, we also observe low error in practice. At the same time, we find that when our conditions are not satisfied, optimization might either succeed or fail to capture the target characteristics. This is not at odds with our statements since we provide sufficient conditions. We now give a more fine-grained discussion of interesting cases for each target.

\textbf{MoU.} First, we discuss a two-dimensional mixture of uniforms with disjoint support, akin to the targets used in \cref{sec:simulations_analytic}. For the \textbf{FKL}, we assess whether (1) a Gaussian correctly locates the mean as indicated via \cref{theorem:theorem_fkl_1} and (2) whether a Student-t succeeds in recovering the mean despite the lack of a guarantee. 
Interestingly, we find that both is the case. For the \textbf{$\boldsymbol{\alpha}$-DIV}, we compare a Student-t surrogate learned with $\alpha=2.0$, which complies with \cref{theorem:theorem_alpha_1}, and $\alpha=0.5$, for which we do not have a guarantee. In both cases, we get a similar fit to the FKL and a good match for the mean.

\textbf{GMM.} Next, we target an \emph{even symmetric} GMM, which has support on all of $\mathbb{R}^2$. The target is \emph{not elliptically symmetric}, and hence we do not expect correlation recovery, except when we optimize a Gaussian with the \textbf{FKL} \citep{wainwright2008graphical}. We observe that also a Student-t distribution optimized with the FKL recovers both mean and correlation well, despite the lack of a formal guarantee. Moreover, we find that a Student-t distribution learned via the $\boldsymbol{\alpha}$\textbf{-DIV} with $\alpha=2.0$ closely matches the target mean, as guaranteed by \cref{theorem:theorem_alpha_1,theorem:theorem_alpha_2}, while for $\alpha=0.5$, it fails to capture the mean. 

\textbf{Ellipse.} Next, we discuss an \emph{elliptically symmetric} target. In this setting, we are \emph{guaranteed} a unique global minimizer at the \emph{correct mean and correlation} with the \textbf{FKL} when optimizing a Gaussian family via \cref{theorem:theorem_fkl_1} and \cref{theorem:theorem_fkl_corr}, and we indeed observe low error in practice. At the same time, also a Student-t distribution learned via the FKL captures the characteristics well, despite the lack of a guarantee. For the $\boldsymbol{\alpha}$\textbf{-DIV}, we guarantee exact recovery of mean and correlation with a Gaussian for $\alpha=2.0$ via \cref{theorem:theorem_alpha_1,theorem:theorem_alpha_2} and \cref{theorem:theorem_alpha_corr}. The approximation indeed achieves low error in practice. Interestingly, also pairing $\alpha=0.5$ with a Student-t recovers the statistics well, despite not fulfilling our sufficient conditions.

\textbf{What happens on asymmetric targets?} Lastly, we construct an \emph{asymmetric GMM target} which is neither even symmetric nor elliptically symmetric and therefore is not covered by any of our guarantees. Indeed, we find that all configurations shown in \cref{fig:results_simulations} fail to match the target mean and correlation, except for the Gaussian fit with the FKL, for which we expect mean and correlation recovery irregardless of symmetry \citep{wainwright2008graphical}.

\textbf{What happens on higher-dimensional targets?} In \cref{app:full_sgd_results}, we provide additional results for higher-dimensional versions of the targets \emph{GMM} and \emph{Ellipse} for $d\in \{4, 8, 16\}$. Our main observations prevail: When our sufficient conditions are satisfied, we recover the corresponding target characteristics with low error. We note however that optimizing the $\alpha$-DIV with $\alpha=2.0$ was challenging in higher dimensions. We found that estimating an informative gradient for the $\alpha$-DIV in the early stages of training was difficult, causing learning to stagnate unless carefully initialized. This is a known problem when optimizing the $\alpha$-DIV with a high value of $\alpha$ in higher dimensions \citep{geffner2020difficulty}, and how to improve the stability of $\alpha$-DIV optimization in practice is an interesting open question. 

\section{Discussion \& future work}\label{sec:discussion}
In this work, we have extended previous guarantees for robust VI with location-scale families under target symmetries to the FKL and $\alpha$-divergence. Notably, our analysis relaxes the assumption that the target needs to be log-concave, which allows us to provide guarantees for multi-modal targets. We demonstrate the practical impact of our guarantees on a diverse set of target densities with varying support and symmetries, and show how and why optimization can fail when our conditions are violated. A natural future direction is to extend our analysis to more expressive variational families such as mixtures of location-scale families. 

We acknowledge that our analysis has the following limitations: First of all, the conditions we provide are only \emph{sufficient}
- investigating whether they are also necessary is interesting future work. Secondly, our guarantees cover the FKL and $\alpha$-DIV, which can be notoriously hard to optimize in practice \citep{jerfel2021variational, geffner2020difficulty}. As we discuss in \cref{sec:learning}, even when we are guaranteed a strictly convex objective for the $\alpha$-DIV via our theorems, the Monte Carlo estimate of the $\alpha$-DIV can have high variance, especially when the variational approximation is poorly initialized.
Moreover, optimizing the FKL w.r.t. an unnormalized target (and without access to i.i.d. target samples) requires the use of self-normalized importance sampling \citep{mcbook} or Markov Chain Monte Carlo (MCMC) \citep{robert2004monte}, which results in a biased estimator \citep{mcbook}. It is unclear whether our guarantees also apply to this setting. Lastly, our guarantees for the $\alpha$-DIV focus on $\alpha > 1$. Finding guarantees for the $\alpha$-DIV with $\alpha \in (0, 1)$ is still an open question.

\subsubsection*{Acknowledgements}
LZ and AV were supported by the ``UNREAL: Unified Reasoning Layer for Trustworthy ML'' project (EP/Y023838/1) selected by the ERC and funded by UKRI EPSRC. The authors are grateful to Nicola Branchini and Maximilian Pixner for useful feedback on the draft.

\subsubsection*{Contributions}
LZ had the initial idea and is responsible for all theoretical derivations, plots and experiments in the paper. LZ wrote the paper with help from AV, who supervised all stages of the project.

\bibliographystyle{plainnat}
\bibliography{bib}

\appendix
\clearpage
\section{Definitions}\label{app:defs}

\begin{table*}
\caption{Base distributions of common location-scale families with location $0$ and unit scale. For the multivariate Laplace distribution, we report its symmetric formulation \citep{eltoft2006multivariate}. $K_{b}$ denotes the modified Bessel function of the second kind \citep[Ch. 9]{abramowitz1948handbook}, $\Gamma$ denotes the Gamma function, and DoF refers to degrees of freedom.} 
\label{tab:dist-sym}
\begin{center}
\resizebox{\textwidth}{!}{
\begin{tabular}{llllll}
\toprule
Distribution & $q_{\vzero}(\vz)$ & $-\log q_{\vzero}$ convex & $-\log q_{\vzero}$ strictly convex & $-\log q_{\vzero}$ str. incr. in $||\vx||$ & $q_{\vzero}$ spherically symmetric\\ \midrule
Gaussian & 
$(2\pi)^{-d/2}\exp\left(-\frac{1}{2}\vz^T\vz\right)$ 
& \yes
&\yes
&\yes
&\yes
\\\midrule
Logistic ($d=1)$ &  $\frac{\exp(-z)}{(1+\exp(-z))^2}$
& \yes
& \yes
& \yes 
& \yes
\\ \midrule
Laplace ($d=1$) & $\frac{1}{2}\exp(-|z|)$
& \yes
& \no
& \yes 
& \yes
\\ \midrule
Laplace ($d > 1$) & $\frac{2}{(2\pi)^{d/2}}\left(\frac{1}{2}\vz^T\vz\right)^{\frac{b}{2}}K_{b}\left(\sqrt{2\vz^T\vz}\right)$, with $b=\frac{2-d}{2}$
& \no 
& \no
& \yes 
& \yes
\\ \midrule
Student-t ($\text{DoF} =v$) & $\frac{\Gamma((v+d)/2)}{\Gamma(v/2)v^{d/2}\pi^{d/2}} \left[1+ \frac{1}{v}\vz^T\vz\right]^{-(v+d)/2}$
& \no
& \no
& \yes 
& \yes
\\ \midrule
Cauchy & $\frac{\Gamma((1+d)/2)}{\Gamma(1/2)\pi^{d/2}} \left[1+ \vz^T\vz\right]^{-(1+d)/2}$
& \no
& \no
& \yes 
& \yes
\\ \midrule
Uniform $(d=1)$ & $\begin{cases} \frac{1}{2}, z \in  (-1, 1)\\ 0, \text{else} \end{cases}$
& \yes
& \no
& \no
& \yes
\\
\bottomrule
\end{tabular}
}
\end{center}
\end{table*}

\subsection{Common location-scale families}\label{app:loc_scale_families}
Below, we provide definitions for common location-scale families. Remember that, by definition, members of a location-scale family can be expressed as 
\begin{equation}
q_{\vnu, S}(\vx) = q_{\vzero}\left(S^{-\frac{1}{2}}(\vx-\vnu)\right)|S|^{-\frac{1}{2}},
\end{equation}
where $\vnu\in \mathbb{R}^d$ is the location parameter and $S\in \mathbb{S}_{++}^d$ is a positive definite scale matrix. 

\begin{figure*}
\begin{tabular}{ccccc}
Gaussian & Logistic & Laplace & Student-t & Cauchy\\
\includegraphics[scale=0.2]{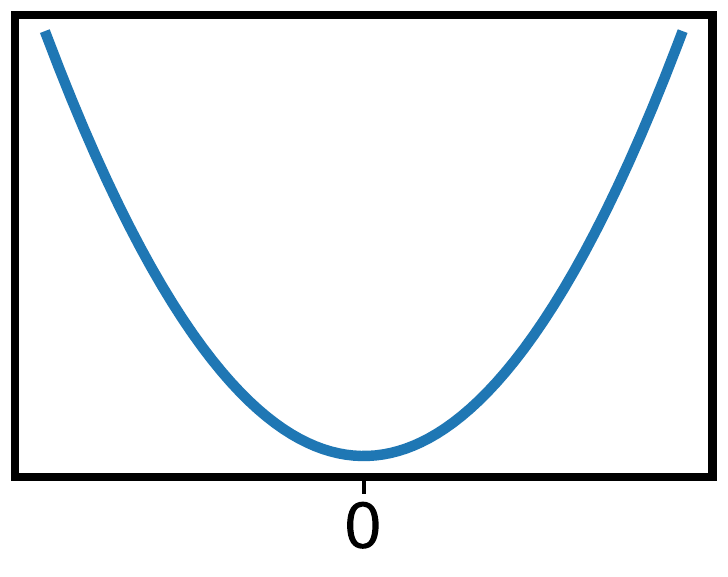}
& 
\includegraphics[scale=0.2]{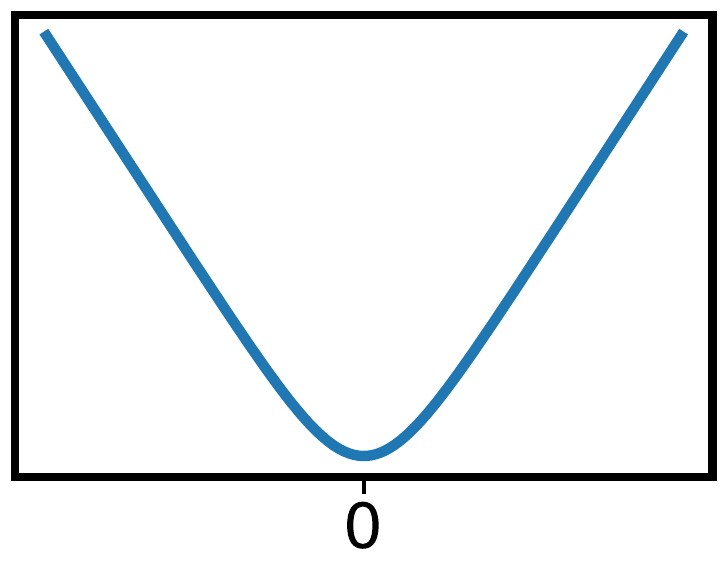}
&
\includegraphics[scale=0.2]{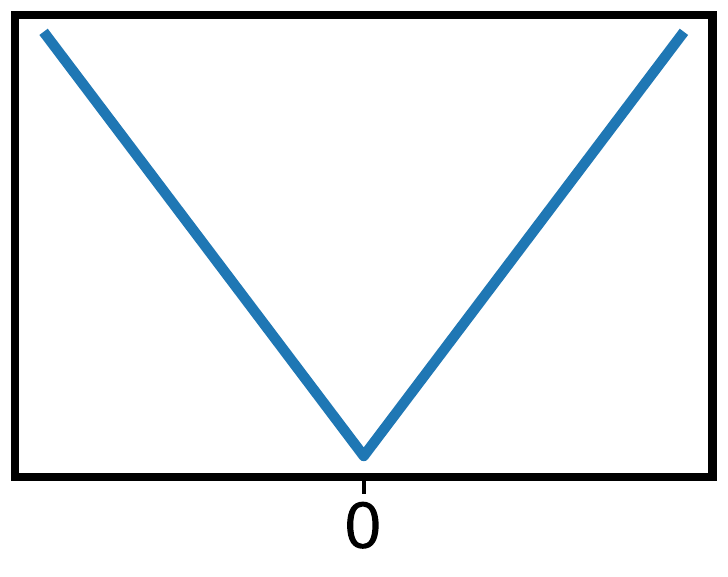}
&
\includegraphics[scale=0.2]{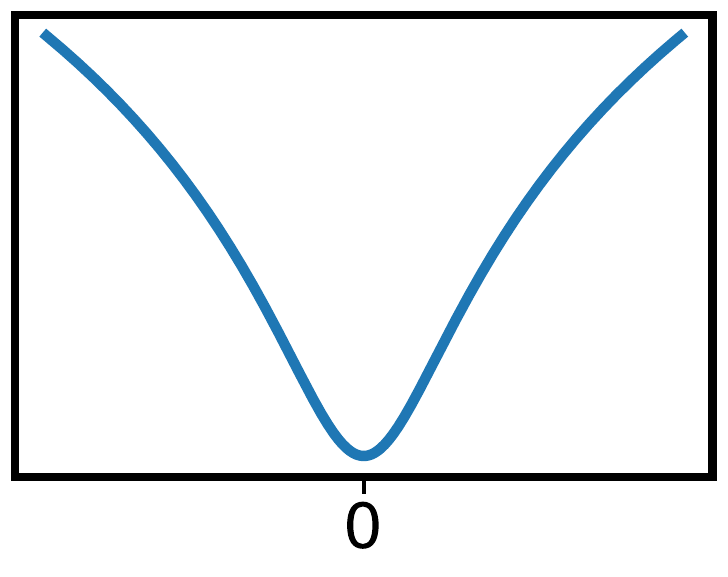}
&
\includegraphics[scale=0.2]{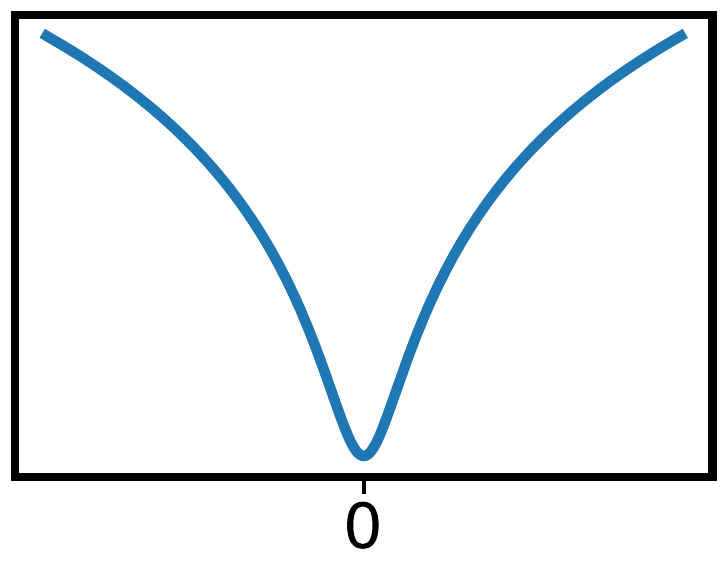}
\end{tabular}
\caption{Visualization of $w(x) = \wkl(x)$ for common location families. We depict the Student-t distribution with $5$ degrees of freedom.}\label{fig:wkl_curves}
\end{figure*}

\begin{figure*}
\begin{center}
\begin{tabular}{ccccc}
\multicolumn{5}{c}{Gaussian}\\
$\alpha=0.1$ & $\alpha=0.5$ & $\alpha=0.9$ & $\alpha=1.1$ & $\alpha=2.0$\\
\includegraphics[scale=0.2]{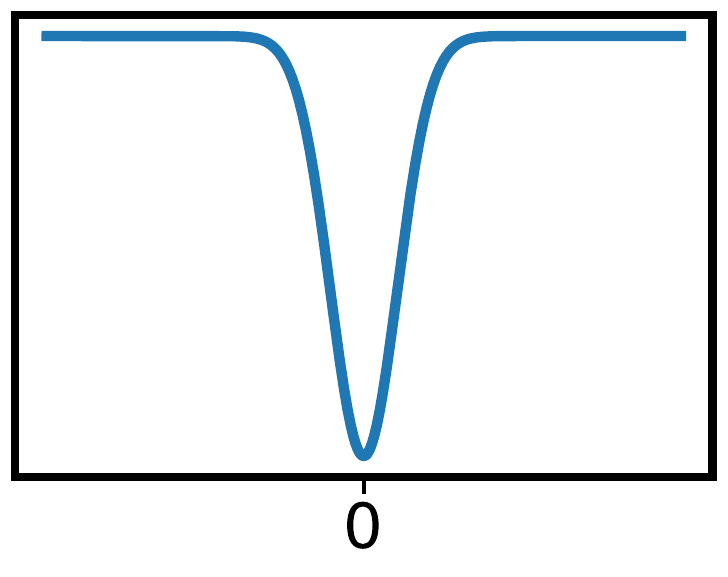}
&
\includegraphics[scale=0.2]{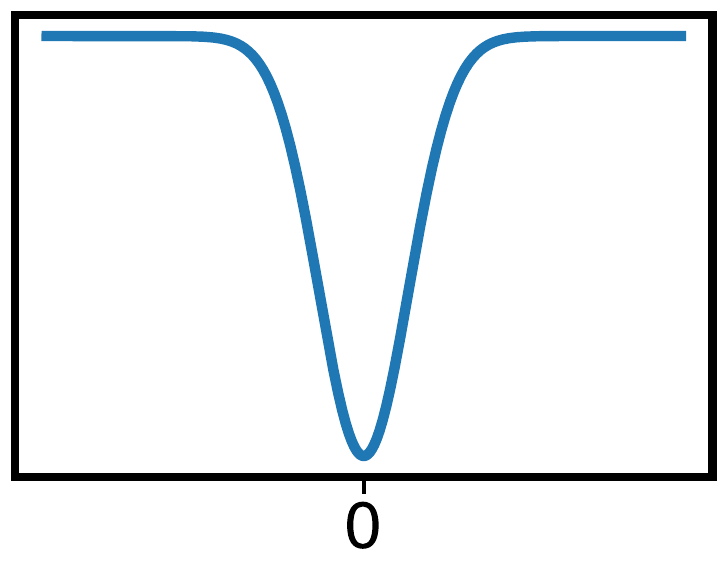}
&
\includegraphics[scale=0.2]{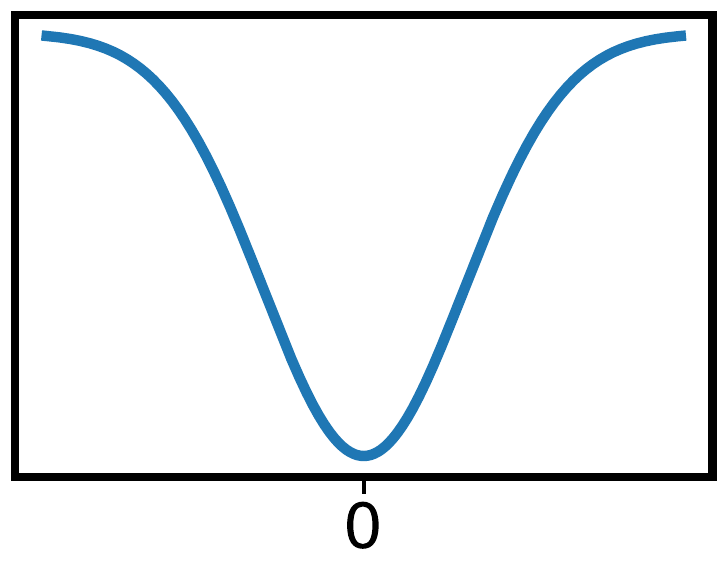}
&
\includegraphics[scale=0.2]{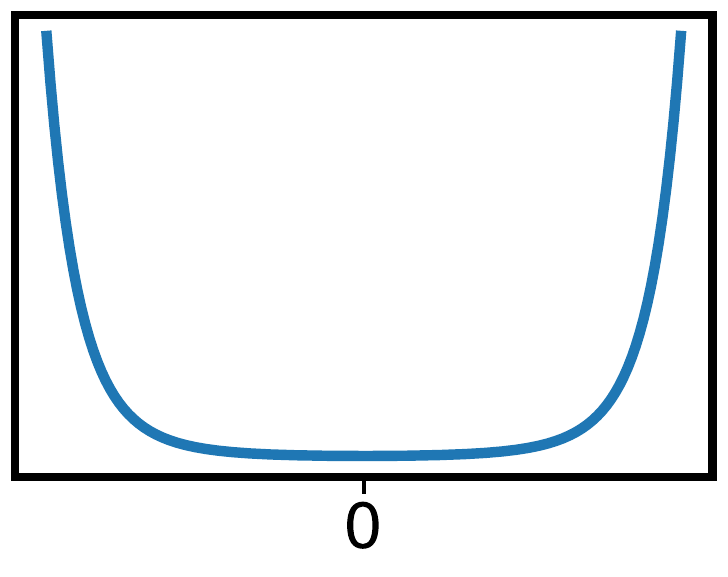}
&
\includegraphics[scale=0.2]{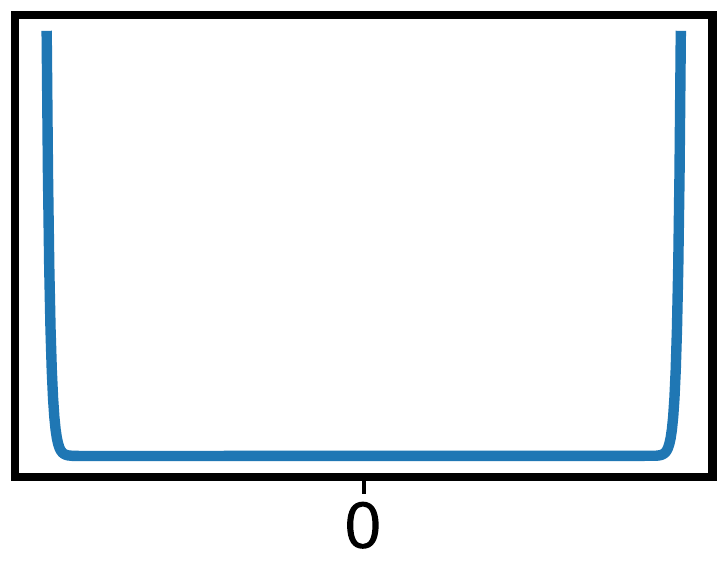}\\
\multicolumn{5}{c}{Laplace}\\
$\alpha=0.1$ & $\alpha=0.5$ & $\alpha=0.9$ & $\alpha=1.1$ & $\alpha=2.0$\\
\includegraphics[scale=0.2]{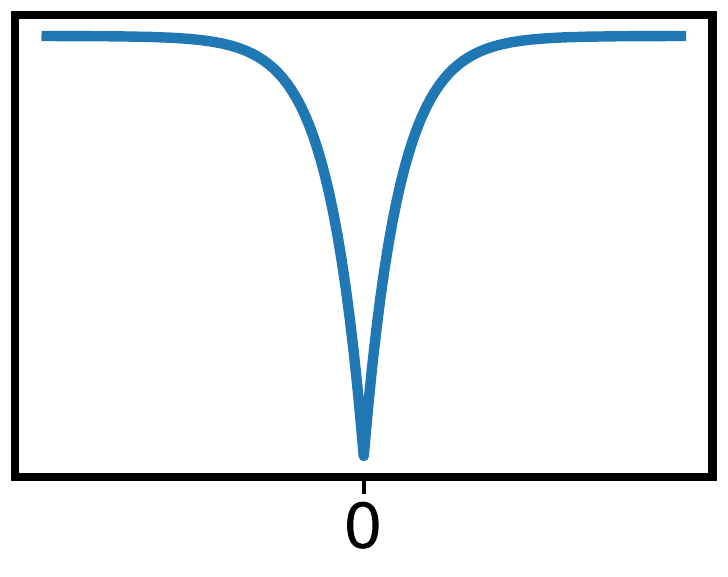}
&
\includegraphics[scale=0.2]{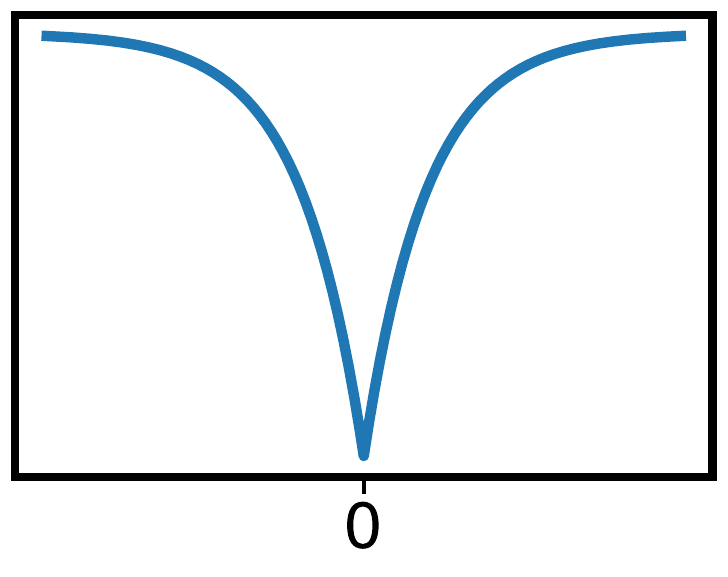}
&
\includegraphics[scale=0.2]{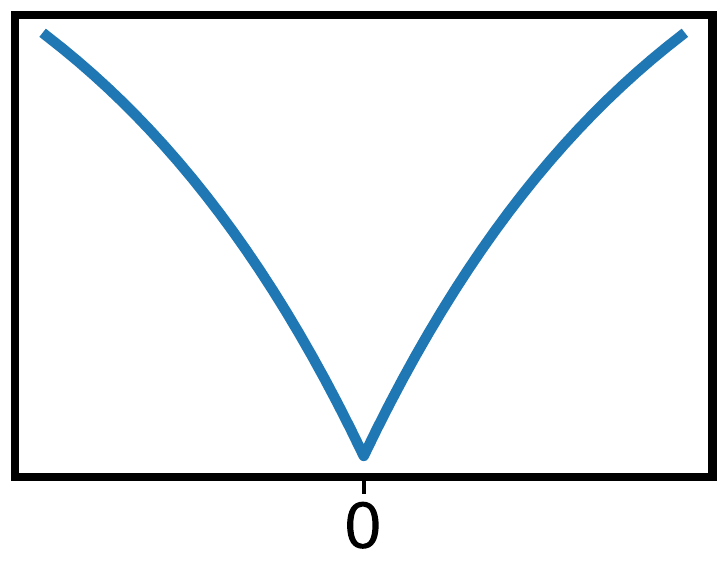}
&
\includegraphics[scale=0.2]{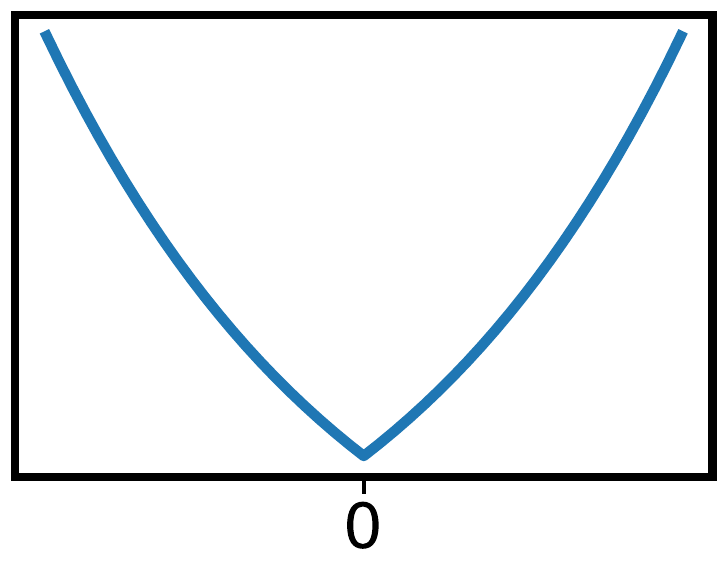}
&
\includegraphics[scale=0.2]{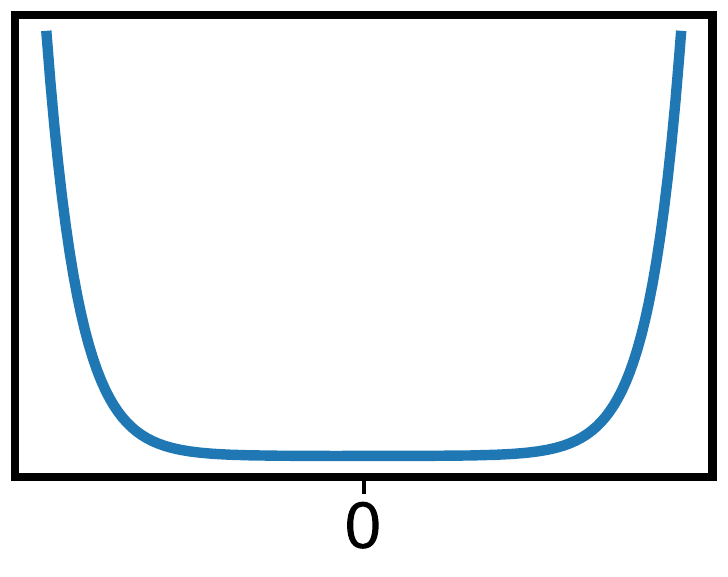}\\
\multicolumn{5}{c}{Logistic}\\
$\alpha=0.1$ & $\alpha=0.5$ & $\alpha=0.9$ & $\alpha=1.1$ & $\alpha=2.0$\\
\includegraphics[scale=0.2]{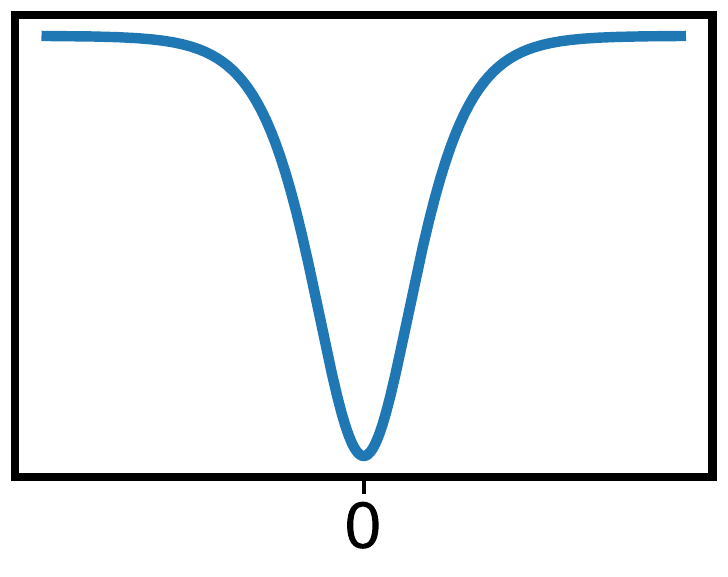}
&
\includegraphics[scale=0.2]{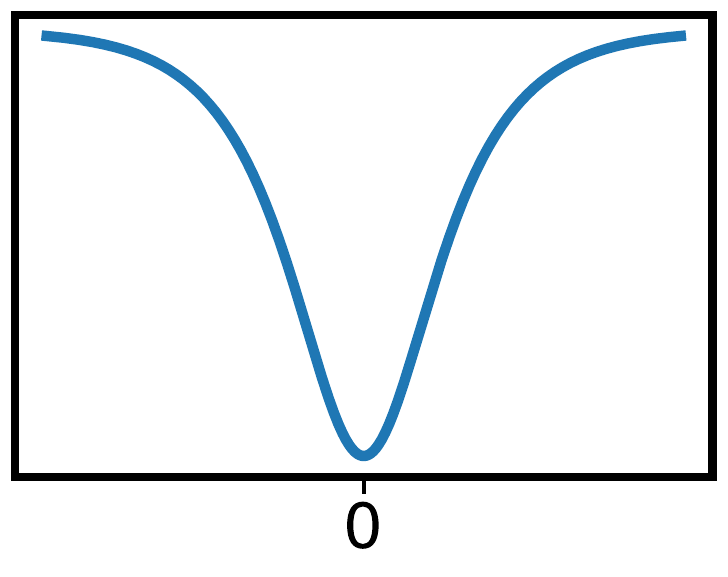}
&
\includegraphics[scale=0.2]{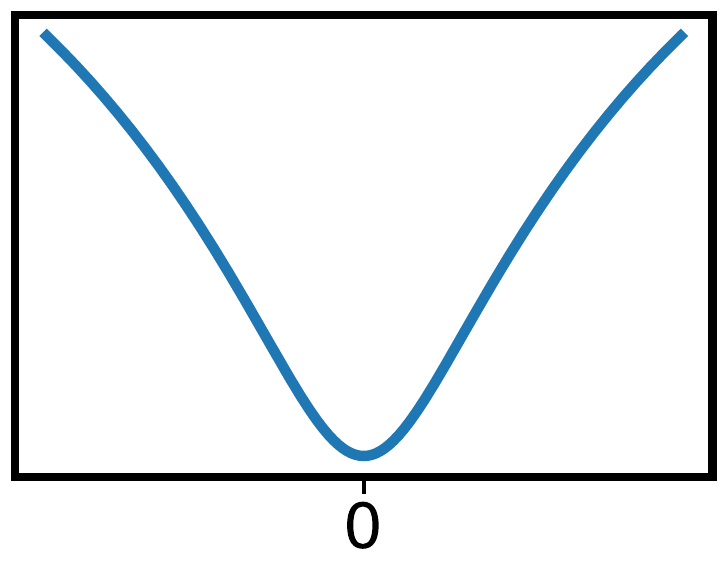}
&
\includegraphics[scale=0.2]{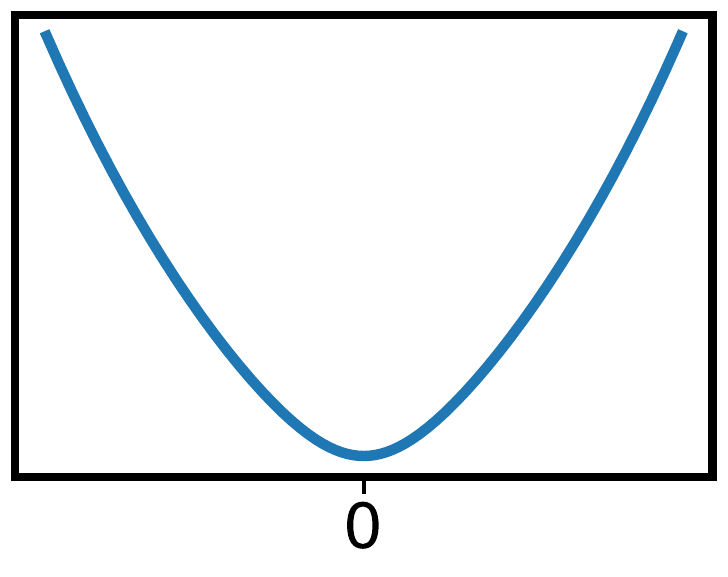}
&
\includegraphics[scale=0.2]{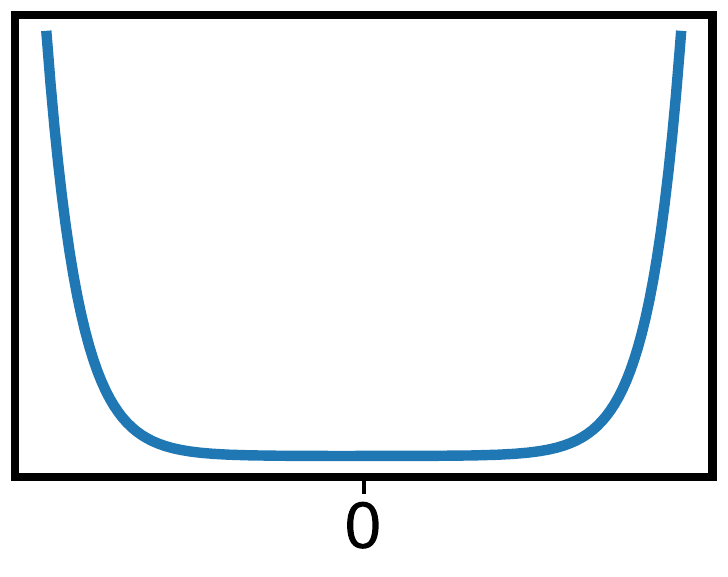}\\
\multicolumn{5}{c}{Student-t (5 degrees of freedom)}\\
$\alpha=0.1$ & $\alpha=0.5$ & $\alpha=0.9$ & $\alpha=1.1$ & $\alpha=2.0$\\
\includegraphics[scale=0.2]{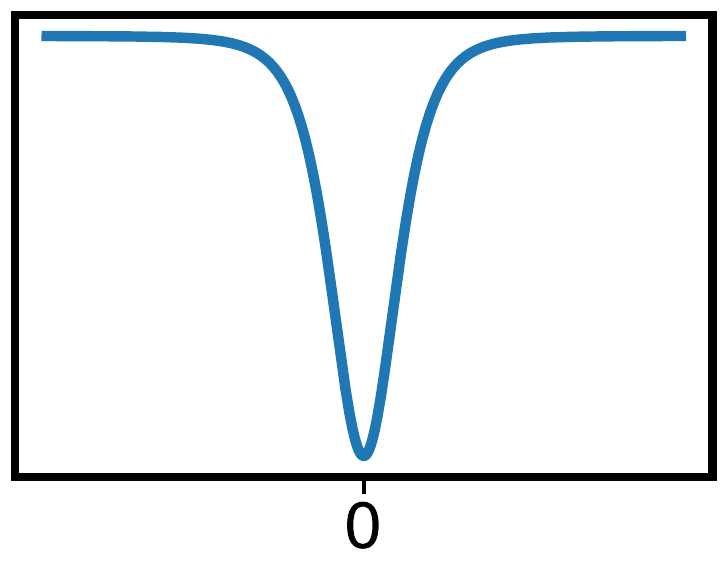}
&
\includegraphics[scale=0.2]{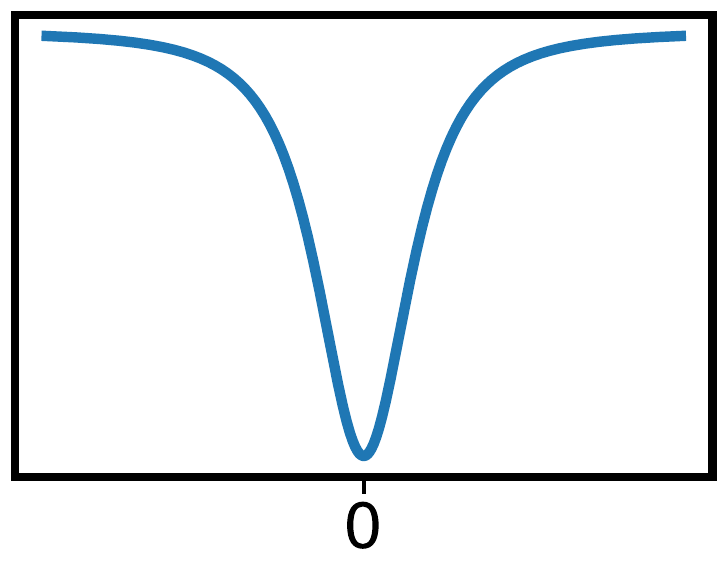}
&
\includegraphics[scale=0.2]{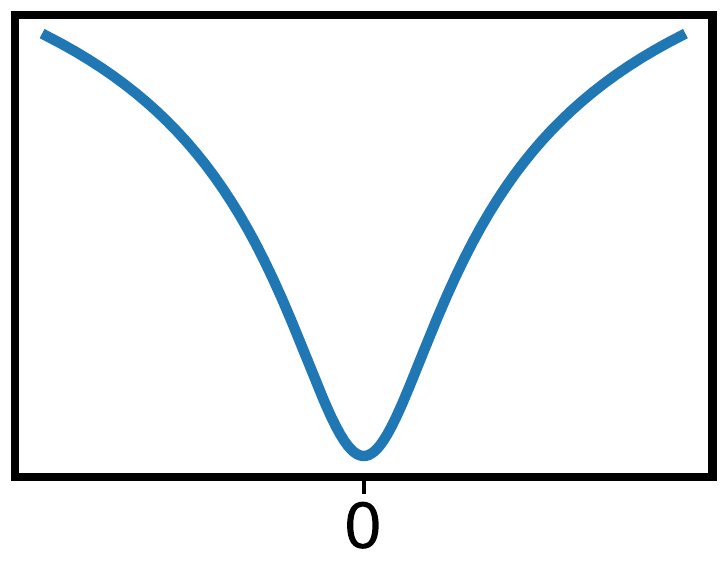}
&
\includegraphics[scale=0.2]{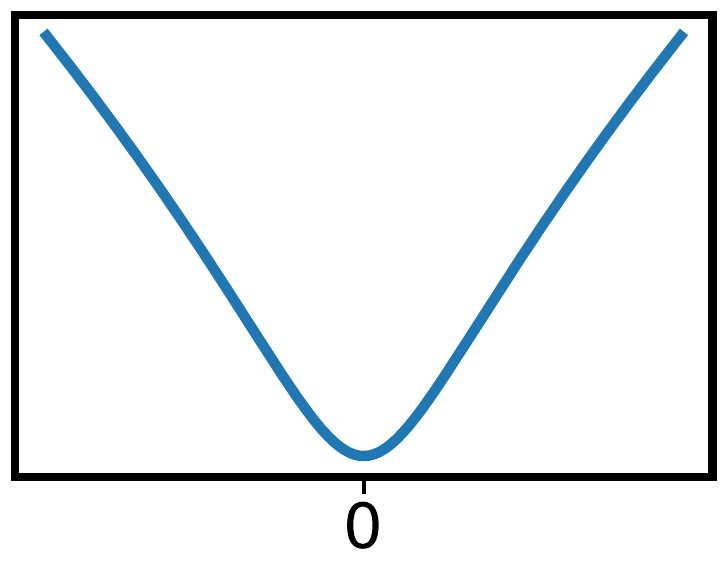}
&
\includegraphics[scale=0.2]{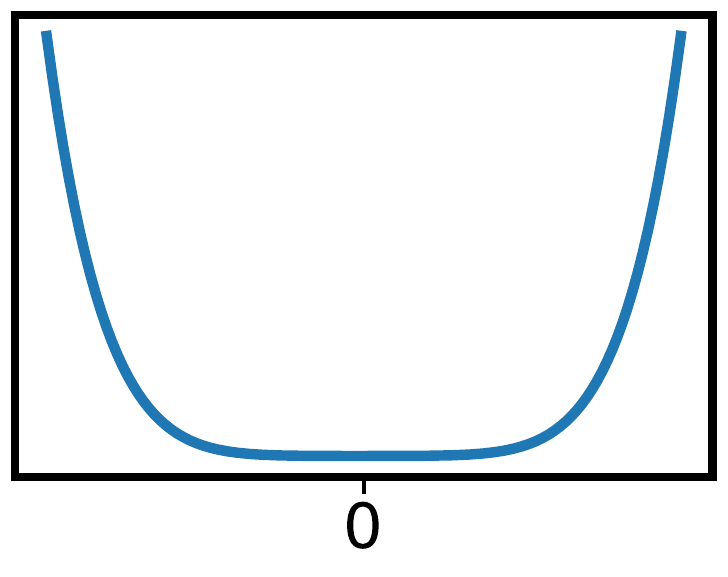}\\
\multicolumn{5}{c}{Cauchy}\\
$\alpha=0.1$ & $\alpha=0.5$ & $\alpha=0.9$ & $\alpha=1.1$ & $\alpha=2.0$\\
\includegraphics[scale=0.2]{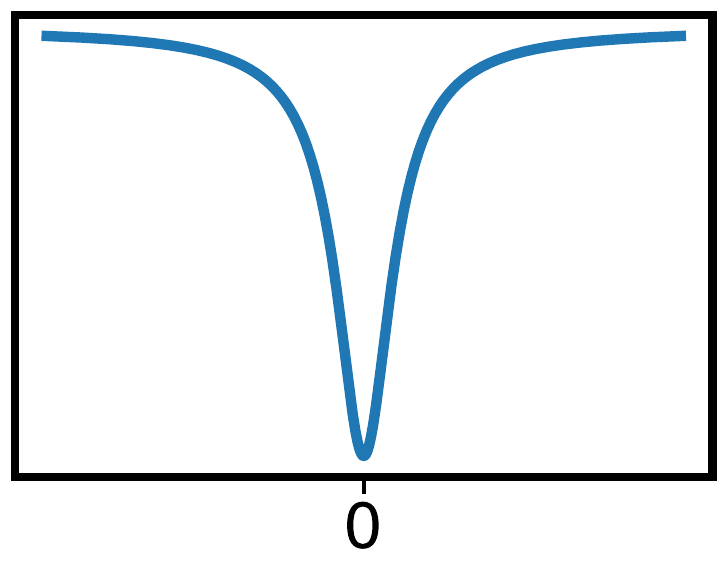}
&
\includegraphics[scale=0.2]{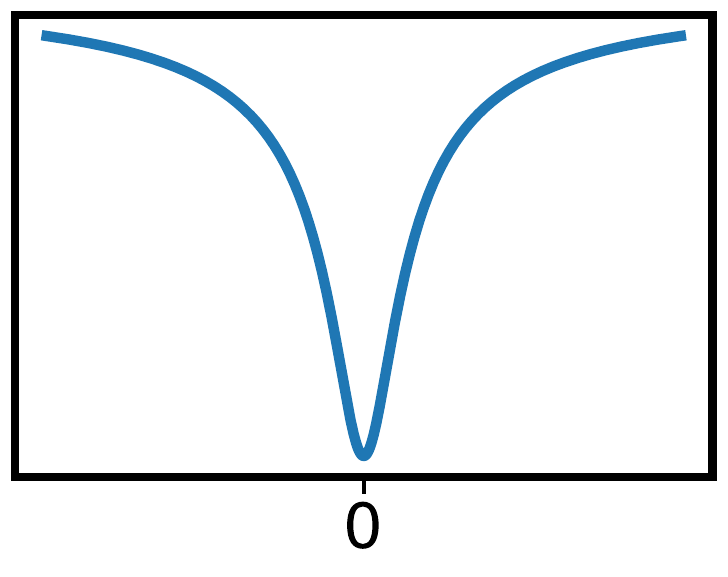}
&
\includegraphics[scale=0.2]{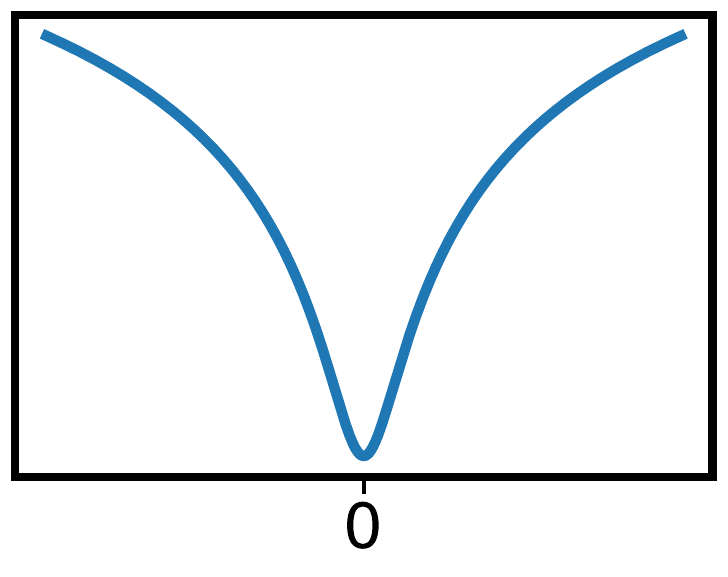}
&
\includegraphics[scale=0.2]{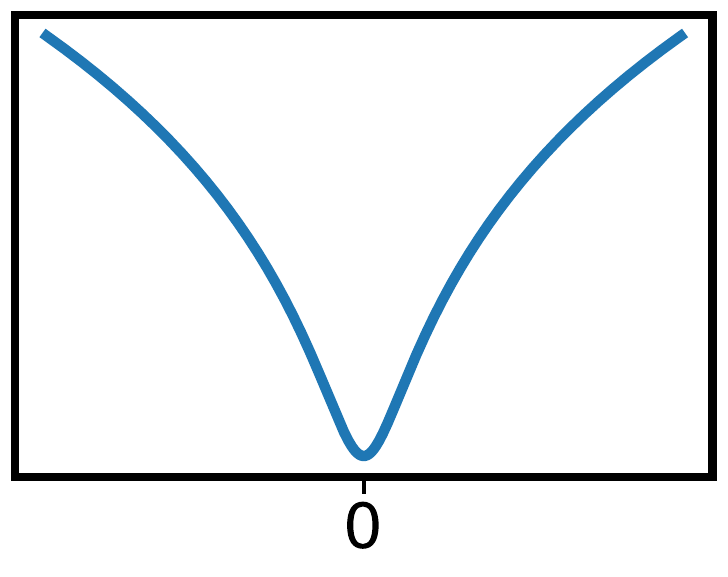}
&
\includegraphics[scale=0.2]{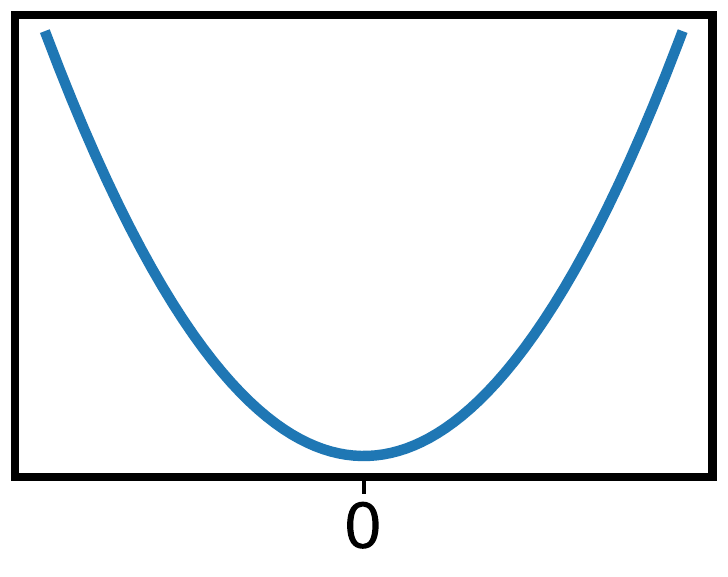}\\
\end{tabular}
\end{center}
\caption{Visualization of $w(x) = \wa(x)$ for common location families. Choosing $\alpha > 1$ tends to result in strictly convex $w$, complying with \cref{theorem:theorem_alpha_1} (except for the heavy-tailed Student-t and Cauchy, which require an even  larger value of $\alpha$). For $\alpha$ close to $0$, the resulting weighting functions are not convex and hence do not comply with our theorems. We do not observe a clear case where \wa is convex but not strictly.}\label{fig:wa_curves}
\end{figure*}

\cref{tab:dist-sym} gives an overview of the base distributions $q_{\vzero}$ of commonly used location-scale families and whether they fulfill properties related to our theorems. In \cref{fig:wkl_curves}, we show how $\wkl$, which is central to our \cref{theorem:theorem_fkl_1} and \cref{theorem:theorem_fkl_2}, behaves for common location families. We fix the scale to $1$. In \cref{fig:wa_curves}, we provide illustrations of $\wa$, as used in our \cref{theorem:theorem_alpha_1} and \cref{theorem:theorem_alpha_2}, for varying values of $\alpha$. We find that choosing a high values of $\alpha$ results in functions $\wa$ that comply with our theorems for mean recovery. 

\newpage

\subsection{$f$-divergences}\label{app:f_div}
An $f$-divergence \citep{renyi1961measures} is given as
\begin{align}\label{eq:f_div}
D_f(p, q)=\int f{\Big(}\frac{p(\vx)}{q(\vx)}{\Big )}q(\vx)d\vx
\end{align}
where $f: \mathbb{R}^+ \rightarrow \mathbb{R}$ is convex, and $f(1)=0$, and $\lim_{u \rightarrow0^+}f(u) = f(0)$. Many common divergences can be expressed as an $f$-divergence, including the RKL, FKL, and $\alpha$-DIV (see \citep{margossian2025generalized} for more examples). Beyond standard assumptions on $f$, the analysis by \citet{margossian2025generalized} requires $f$ and $\varphi$ to be differentiable.

\subsubsection{Reverse KL divergence}\label{app:rkl}
Choosing  $f(\vu)=-\log(\vu)$ recovers the RKL, given as
\begin{align}\label{eq:RKL}
\tag{RKL}
D_{\text{RKL}}(p,q_{\vnu})
&= \int\log{\Big(}\frac{q_{\vnu}(\vx)}{p(\vx)}\Big{)}q_{\vnu}(\vx)d\vx.
\end{align}

Notably, the $f$ defining the RKL, and its associated $\varphi(\vu)=-\vu$, fulfill the conditions for a unique minimizer specified in \cref{theorem:charles_f} (under the additional condition that \cref{assumptions_mean} holds and the target is somewhere strictly log-concave).

\subsubsection{Forward KL divergence}\label{app:fkl}
Choosing  $f(\vu)=\vu\log(\vu)$ recovers the FKL, given as
\begin{align}\label{eq:FKL_app}
\tag{FKL}
D_{\text{FKL}}(p,q_{\vnu})
&= \int\log{\Big(}\frac{p(\vx)}{q_{\vnu}(\vx)}\Big{)}p(\vx)d\vx.
\end{align}

\subsubsection{$\alpha$-divergence} Choosing $f(\vu)= \frac{\vu^{\alpha}-1}{\alpha(\alpha-1)}$ recovers the $\alpha$-divergence \citep{cichocki2010families},
\begin{align}\label{eq:alpha_div}
\tag{$\alpha$-DIV}
D_{\alpha}(p,q_{\vnu})
&= \frac{1}{\alpha(\alpha-1)}\int {\Big[}\Big{(}\frac{p(\vx)}{q_{\vnu}(\vx)}\Big{)}^{\alpha}-1{\Big]}q_{\vnu}(\vx) d\vx,
\end{align}
where $\alpha > 0$ and $\alpha \neq 1.$

\section{Proofs}\label{app:proofs}
In this section, we present the proofs for our theorems given in \cref{sec:our_gaurantees}. First, we prove some useful lemmas (\cref{app:lemmas}) to simplify the proofs of our main results.
\subsection{Lemmas}\label{app:lemmas}
\begin{lemma}\label{lemma:strict_composition}
Let $f: \mathbb{R}^d \rightarrow \mathbb{R}$ be strictly convex. Further, let $g(\vx) = A\vx + \vb$ be an injective affine mapping with $A \in \mathbb{R}^{d\times d}$, $\vb \in \mathbb{R}^d$. Then $f \circ g$ is strictly convex.
\end{lemma}

\paragraph{Proof.} As convexity is preserved under composition with an affine mapping \citep[Ch. 3.2.2.]{boyd2004convex}, we have $\forall \vx_1, \vx_2 \in \mathbb{R}^d, \vx_1 \neq \vx_2$, $\forall \lambda \in  (0, 1)$:
\begin{align}
f(\lambda g(\vx_1) + (1-\lambda)g(\vx_2)) \leq \lambda f(g(\vx_1)) + (1-\lambda)f(g(\vx_2)). \label{eq:conv}
\end{align}
Further, $f$ is strictly convex, i.e., $\forall \vx_1, \vx_2 \in \mathbb{R}^d, \vx_1\neq \vx_2$, $\forall \lambda \in (0, 1)$:
\begin{align}
f(\lambda \vx_1 + (1-\lambda)\vx_2) < \lambda f(\vx_1) + (1-\lambda)f(\vx_2).
\end{align}
Moreover, $g$ is injective, i.e., $\forall \vx_1,\vx_2 \in \mathbb{R}^d: ~\vx_1 \neq \vx_2 \implies g(\vx_1) \neq g(\vx_2)$. Hence, the inequality in \cref{eq:conv} is strict.

\begin{lemma}\label{lemma:strict_integral}
Let $f_{\vx}: \mathbb{R}^d \rightarrow \mathbb{R}$ be strictly convex in $\vnu$. Further, let $p(\vx)\geq 0 ~\forall~ \vx \in \mathbb{R}^d$. Then,
\begin{align*}
h(\vnu)=\int_{\mathbb{R}^d} f_{\vx}(\vnu)p(\vx)d\vx
\end{align*}
is strictly convex in $\vnu$.
\end{lemma}

\paragraph{Proof.} First of all, note that it suffices to take the integral over the support of $p$, as $p(\vx)=0$ otherwise by definition. Hence, we have
\begin{align}
h(\vnu)=\int_{\mathbb{R}^d} f_{\vx}(\vnu)p(\vx)d\vx = \int_{\supp(p)} f_{\vx}(\vnu)p(\vx)d\vx.
\end{align}

By strict convexity of $f_{\vx}$ in $\vnu$ we have $\forall \vnu_1, \vnu_2 \in \mathbb{R}^d, \vnu_1\neq \vnu_2$, $\forall \lambda \in 
(0, 1)$:
\begin{align}
f_{\vx}(\lambda \vnu_1+(1-\lambda)\vnu_2) &< \lambda f_{\vx}(\vnu_1)+(1-\lambda)f_{\vx}(\vnu_2).\label{eq:conv_2}
\end{align}

Now, for $\vx \in \text{supp($p$)}$, it holds that $p(\vx)> 0$, and hence multiplying \cref{eq:conv} with $p(\vx)$ on both sides preserves strict convexity, i.e., $\forall \vnu_1, \vnu_2 \in \mathbb{R}^d, \vnu_1\neq \vnu_2$, $\forall \lambda \in (0, 1)$:
\begin{align}
f_{\vx}(\lambda \vnu_1+(1-\lambda)\vnu_2)p(\vx) &< \lambda f_{\vx}(\vnu_1)p(\vx) + (1-\lambda) f_{\vx}(\vnu_2)p(\vx). \label{eq:lemma_2}
\end{align}
Integrating both sides of \cref{eq:lemma_2}, we obtain $\forall \vnu_1, \vnu_2 \in \mathbb{R}^d, \vnu_1 \neq \vnu_2, \forall \lambda \in (0, 1)$:
\begin{align}
h(\lambda \vnu_1 + (1-\lambda)\vnu_2) &= \int_{\text{supp}(p)} f_{\vx}(\vnu_1+(1-\lambda)\vnu_2) p(\vx) d\vx \\
&< \lambda \int_{\text{supp}(p)} f_{\vx}(\vnu_1) p(\vx)d\vx + (1-\lambda)\int_{\text{supp}(p)}f_{\vx}(\vnu_2) p(\vx) d\vx\\
& = \lambda h(\vnu_1) + (1-\lambda) h(\vnu_2),
\end{align}
which proves strict convexity of $h$ in $\vnu$.

\subsection{Mean recovery} \label{app:proofs_mean}
In this section, we present the proofs for our guarantees for exact mean recovery. 

We now restate our core assumptions.

\ASSUMPTIONSM*

Throughout our proofs, we use that we can simplify both the FKL and $\alpha$-DIV when treating $S$ as a constant, as detailed below.
\begin{align}
D_{\text{FKL}}(p,q_{\vnu})
&= \int\log{\Big(}\frac{p(\vx)}{q_{\vnu}(\vx)}\Big{)}p(\vx)d\vx\\ \tag{Drop $-H(p)$}
&\sim \int-\log q_{\vnu}(\vx)p(\vx)d\vx\\
\tag{Plug in $q_{\vnu}$}
&= \int-\log {\Big(}q_{\vzero}{\Big(}S^{-\frac{1}{2}}(\vx-\vnu){\Big)}|S|^{-\frac{1}{2}}{\Big)}p(\vx)d\vx\\
\tag{Pull out $-\log|S|^{-\frac{1}{2}}$}
&= \int-\log {\Big(}q_{\vzero}{\Big(}S^{-\frac{1}{2}}(\vx-\vnu){\Big)}{\Big)}p(\vx)d\vx-\log|S|^{-\frac{1}{2}}\\
\tag{Drop $-\log|S|^{-\frac{1}{2}}$ and set $w(\vz):=-\log q_{\vzero}(\vz)$}
&\sim\int w{\Big(}S^{-\frac{1}{2}}(\vx-\vnu)\Big{)}p(\vx)d\vx.
\end{align}
where $w(\vz) := -\log q_{\vzero}(\vz)$.

\begin{align}
D_{\alpha}(p,q_{\vnu})
&= \frac{1}{\alpha(\alpha-1)}\int {\Big[}\Big{(}\frac{p(\vx)}{q_{\vnu}(\vx)}\Big{)}^{\alpha}-1{\Big]}q_{\vnu}(\vx) d\vx\\
&=\int \frac{1}{\alpha(\alpha-1)}q_{\vnu}^{1-\alpha}(\vx) p^{\alpha}(\vx)d\vx - \frac{1}{\alpha(\alpha-1)}\\
&\sim\int \frac{1}{\alpha(\alpha-1)}q_{\vnu}^{1-\alpha}(\vx) p^{\alpha}(\vx)d\vx
\tag{Drop $- \frac{1}{\alpha(\alpha-1)}$}\\
&=\int \frac{1}{\alpha(\alpha-1)}{\Big(}q_{\vzero}{\Big(}S^{-\frac{1}{2}}(\vx-\vnu){\Big)}|S|^{-\frac{1}{2}}{\Big)}^{1-\alpha} p^{\alpha}(\vx)d\vx \tag{Plug in $q_{\vnu}$}\\
& |S|^{-\frac{1-\alpha}{2}} \int \frac{1}{\alpha(\alpha-1)}q_{\vzero}^{1-\alpha}{\Big(}S^{-\frac{1}{2}}(\vx-\vnu){\Big)}p^{\alpha}(\vx)d\vx \\
\tag{Drop $|S|^{-\frac{1-\alpha}{2}} > 0$}
&\sim \int \frac{1}{\alpha(\alpha-1)}q_{\vzero}^{1-\alpha}{\Big(}S^{-\frac{1}{2}}(\vx-\vnu){\Big)}p^{\alpha}(\vx)d\vx \\
&:= \int w_{\alpha}(S^{-\frac{1}{2}}(\vx-\vnu)) p^{\alpha}(\vx)d\vx, \tag{Set $w_{\alpha}(\vz):=\frac{1}{\alpha(\alpha-1)}q_{\vzero}^{1-\alpha}(\vz)$}
\end{align}
where $w_{\alpha}(\vz):=\frac{1}{\alpha(\alpha-1)}q_{\vzero}^{1-\alpha}(\vz)$. Note that we do not omit $\frac{1}{\alpha(\alpha-1)}$, as it can be positive or negative depending on the chosen value of $\alpha$.

We now prove our main theorems presented in \cref{sec:mean_recovery}. We assume throughout that \cref{assumptions_mean} holds.

\subsection{Proof of Theorem 1 and Theorem 3}\label{proof:theorem_1_3}
We first proof \cref{theorem:theorem_fkl_1} and \cref{theorem:theorem_alpha_1} as they share analogous reasoning.
\subsubsection{Proof of Theorem 1}\label{proof:theorem_1}
\FKLSC*
\paragraph{Proof.}
\cref{theorem:charles_f} tells us that $D_{\text{FKL}}(p, q_{\vnu})$ has a stationary point at $\vnu=\vmu$. Therefore, if $D_{\text{FKL}}(p, q_{\vnu})$ (\cref{eq:FKL_app}) is strictly convex in $\vnu$, the minimizer will be unique and located at $\vmu$. Note that $w(\vz):=-\log (q_{\vzero}(\vz))$ is strictly convex in the setting of \cref{theorem:theorem_fkl_1} as this is our additional sufficient condition. Moreover, $g_{\vx}(\vnu):=S^{-\frac{1}{2}}(\vx-\vnu)$
is an affine mapping of $\vnu$. 
Since $S \in \mathbb{R}^{d \times d}$ is positive definite (see \cref{def:loc_scale}), so is $S^{-\frac{1}{2}}$, and hence $g$ is bijective \citep[Ch. 4]{bhatia2009positive}. As a result, $w(g_{\vx}(\vnu))$ is strictly convex in $\vnu$ by \cref{lemma:strict_composition}.
Moreover, since $p$ is a density, we have $p(\vx) \geq 0$ for all $\vx$ and hence $D_{\text{FKL}}(p, q_{\vnu}) = \int_{\mathbb{R}^d} w(g_{\vx}(\vnu))p(\vx)d\vx +C$ is strictly convex in $\vnu$ by \cref{lemma:strict_integral}. As a result, we have a unique global minimizer of $D_{\text{FKL}}(p, q_{\vnu})$ at $\vnu=\vmu$.

\subsubsection{Proof of Theorem 3}\label{proof:theorem_3}
\ASC*
\paragraph{Proof.} We can use the same reasoning as for proving \cref{theorem:theorem_fkl_1}. \cref{theorem:charles_f} tells us that $D_{\alpha}(p, q_{\vnu})$ has a stationary point at $\vnu=\vmu$. The function $w(\vz):=\wa(\vz)$ is strictly convex in the setting of \cref{theorem:theorem_fkl_1} as this is our additional sufficient condition. Moreover, $g_{\vx}(\vnu):=S^{-\frac{1}{2}}(\vx-\vnu)$ is an affine mapping of $\vnu$. Since $S \in \mathbb{R}^{d \times d}$ is positive definite (see \cref{def:loc_scale}), so is $S^{-\frac{1}{2}}$, and hence $g$ is bijective \citep[Ch. 4]{bhatia2009positive}. Hence, we conclude $w(g_{\vx}(\vnu))$ is strictly convex in $\vnu$ by applying \cref{lemma:strict_composition}. Moreover, since $p$ is a density, we have $p^{\alpha}(\vx) \geq 0$ for all $\vx \in \text{supp}(p)$ and for any $\alpha \in \mathbb{R}$, and hence $D_{\alpha}(p, q_{\vnu}) \sim \int_{\mathbb{R}^d} w(g_{\vx}(\vnu))p^{\alpha}(\vx)d\vx$ is strictly convex in $\vnu$ by \cref{lemma:strict_integral}. As a result, we have a unique global minimizer of $D_{\alpha}(p, q_{\vnu})$ at $\vnu=\vmu$.

\subsection{Proof of Theorem 2 and Theorem 4}

Next, we provide a proof for and \cref{theorem:theorem_fkl_2} and \cref{theorem:theorem_alpha_2}. We prove the following more general statement, which is agnostic to the specific form of $w$ and hence can be applied to both the  FKL and $\alpha$-DIV.
\begin{theorem}\label{theorem:theorem_universal_convex}
Let $q_{\vnu}$ be a location family with a base distribution that is even symmetric around the origin (in line with \cref{assumptions_mean}). Moreover, let $I: \mathbb{R}^d\rightarrow\mathbb{R}^+_{0}$ be even symmetric around $\vmu$. Let $w = h \circ q_{\vzero}$ with $h: \mathbb{R}^+_0 \rightarrow \mathbb{R}$. Lastly, assume
$D(p, q_{\vnu})$ is well-defined for all  values of $\vnu$, and $\vmu$ is finite. Then, $D(p, q_{\vnu}) = \int w(S^{-\frac{1}{2}}(\vx-\vnu))I(\vx)d\vx$ has a unique global minimizer at $\vnu = \vmu$ if \textbf{additionally}
\begin{itemize}
    \item $w$ is convex and strictly increasing in $||\vx||$
    \item \csuppI 
\end{itemize}
\end{theorem}

\paragraph{Proof.}  Let $\vnu' \in \mathbb{R}^d$ and $\vnu' \neq \vzero$. We want to show that $\forall \vnu': D(p, q_{\vmu+\vnu'}) - D(p, q_{\vmu}) > 0$.  

Let $\vtau := \vx-\vmu$ (and hence $\vx = \vtau  + \vmu$). 
\begin{align}
D(p, q_{\vmu+\vnu'}) - D(p, q_{\vmu})
&= \int_{\mathbb{R}^d}[w(S^{-\frac{1}{2}}(\vx-\vmu-\vnu'))-w(S^{-\frac{1}{2}}(\vx-\vmu))]I(\vx)d\vx \tag{Def. of $q_{\vnu}$}\\
&= \int_{\mathbb{R}^d}[w(S^{-\frac{1}{2}}(\vtau-\vnu'))-w(S^{-\frac{1}{2}}\vtau)]I(\vtau + \vmu)d\vtau.  \tag{$\vtau := \vx-\vmu$}
\end{align}

Under the assumption that $w$ is strictly increasing in $||\vx||$, it holds that $w(S^{-\frac{1}{2}}(\vtau-\vnu')) \geq w(S^{-\frac{1}{2}}\vtau)$ iff $||S^{-\frac{1}{2}}(\vtau-\vnu')|| \geq ||S^{-\frac{1}{2}}\vtau||$. This condition corresponds to a half-space defined w.r.t. the hyperplane $||S^{-\frac{1}{2}}(\vtau-\vnu')|| - ||S^{-\frac{1}{2}}\vtau||=\vzero$. Let $(\cdot)$ denote the inner product. It holds that
\begin{equation*}
||S^{-\frac{1}{2}}(\vtau-\vnu')|| \geq ||S^{-\frac{1}{2}}\vtau|| \iff \vtau \cdot S^{-1}\vnu' \leq\frac{||S^{-\frac{1}{2}}\vnu'||^2}{2}
\end{equation*}

The above follows immediately from the properties of the inner product \citep[Chapter 3]{kreyszig1991introductory}:
\fbox{
\begin{minipage}{0.9\linewidth}
\begin{align}
||S^{-\frac{1}{2}}(\vtau-\vnu')||^2 &\geq ||S^{-\frac{1}{2}}\vtau||^2\\\tag{expanded quadratic form}
||S^{-\frac{1}{2}}\vtau||^2 - 2\vtau\cdot S^{-1} \vnu' + ||S^{-\frac{1}{2}}\vnu'||^2 &\geq ||S^{-\frac{1}{2}}\vtau||^2\\
-2\vtau\cdot S^{-1} \vnu' &\geq -||S^{-\frac{1}{2}}\vnu'||^2\\
\vtau \cdot S^{-1}\vnu' &\leq\frac{||S^{-\frac{1}{2}}\vnu'||^2}{2}.
\end{align}
\end{minipage}
}

\begin{figure}[!t]
\resizebox{\textwidth}{!}{
\begin{tabular}{ccc}
$p(\vx)$ & $p(\vmu+\vx), ~\vnu' = (1.53, -0.94)^T$ &  $p(\vmu+\vx), ~\vnu'=(1.53, -0.94)^T$\\
\includegraphics[scale=0.4]{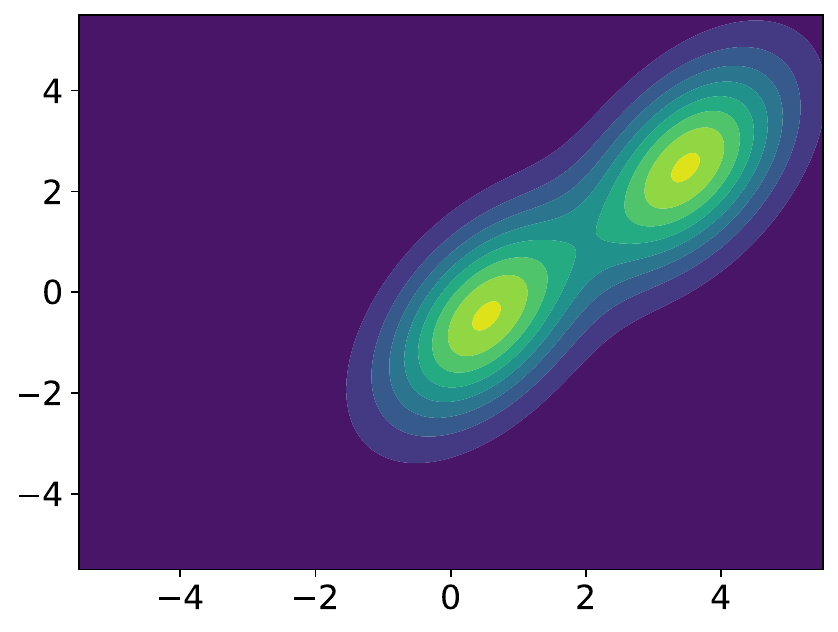}
&
\includegraphics[scale=0.4]{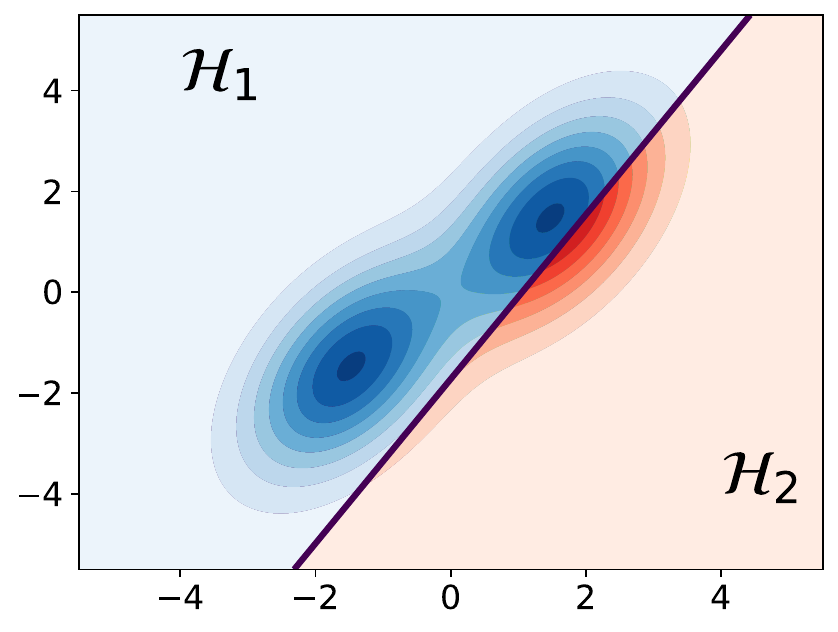}
&
\includegraphics[scale=0.4]{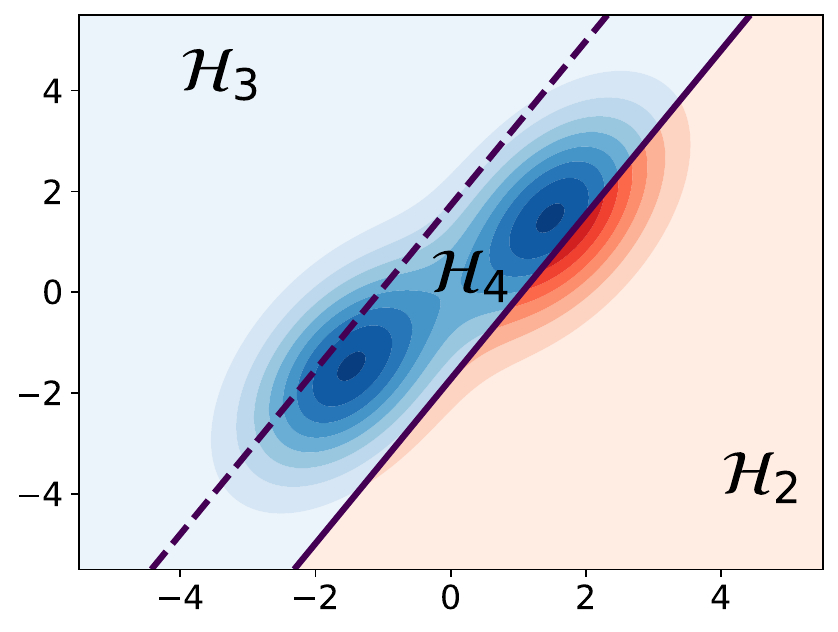}
\end{tabular}
}
\caption{Illustration of the domain partitioning used for proving \cref{theorem:theorem_universal_convex}. \textbf{(1)} A bimodal target density $p$ that is even symmetric around $\vmu=(2, 1)^T$. \textbf{(2)} The shifted target $p(\vmu+\vx)$, which is even symmetric around $\vzero$. Areas where $w(S^{-\frac{1}{2}}(\vx - \vnu')) \geq w(S^{-\frac{1}{2}}\vx)$ are depicted in blue, the rest in red. These areas correspond to $\mathcal{H}_1$ and $\mathcal{H}_2$ in the proof. \textbf{(3)} Additional partition of $\mathcal{H}_1$ into $\mathcal{H}_3$ and $\mathcal{H}_4$. In the above examples, $w$ is the negative log-density of a multivariate standard Gaussian.}\label{app_fig:hyperplanes}
\end{figure}

Let now $\mathcal{H}_1:=\{\vtau: \vtau \cdot S^{-1}\vnu'\leq\frac{||S^{-\frac{1}{2}}\vnu'||^2}{2}\}$ and $\mathcal{H}_2:=\{\vtau: \vtau \cdot S^{-1}\vnu' > \frac{||S^{-\frac{1}{2}}\vnu'||^2}{2}\}$ be a partition of the domain. Remember that on $\mathcal{H}_1$, $w(S^{-\frac{1}{2}}(\vtau-\vnu'))-w(S^{-\frac{1}{2}}\vtau) \geq0$ and on $\mathcal{H}_2$, $w(S^{-\frac{1}{2}}(\vtau-\vnu'))-w(S^{-\frac{1}{2}}\vtau) \leq 0$ by definition. \cref{app_fig:hyperplanes} provides an illustration. Using this partition, we obtain
\begin{align}
D(p, q_{\vmu+\vnu'}) - D(p, q_{\vmu})
&= \int_{\mathbb{R}^d}[w(S^{-\frac{1}{2}}(\vtau-\vnu'))-w(S^{-\frac{1}{2}}\vtau)]I(\vtau + \vmu)d\vtau\\
&= \int_{\mathcal{H}_1}[w(S^{-\frac{1}{2}}(\vtau-\vnu'))-w(S^{-\frac{1}{2}}\vtau)]I(\vtau + \vmu)d\vtau \\
&+ \int_{\mathcal{H}_2}[w(S^{-\frac{1}{2}}(\vtau-\vnu'))-w(S^{-\frac{1}{2}}\vtau)]I(\vtau + \vmu)d\vtau.
\end{align}
For our theorem to hold we hence need to show that
\begin{align}
\int_{\mathcal{H}_1}[w(S^{-\frac{1}{2}}(\vtau-\vnu'))-w(S^{-\frac{1}{2}}\vtau)]I(\vtau + \vmu)d\vtau
&\geq -\int_{\mathcal{H}_2}[w(S^{-\frac{1}{2}}(\vtau-\vnu'))-w(S^{-\frac{1}{2}}\vtau)]I(\vtau + \vmu)d\vtau.
\end{align}
We now further split $\mathcal{H}_1$ into the disjoint regions $\mathcal{H}_3:=\{\vtau: \vtau \cdot S^{-1}\vnu'<-\frac{||S^{-\frac{1}{2}}\vnu'||^2}{2}\}$ and $\mathcal{H}_4:=\{\vtau:~ -\frac{||S^{-\frac{1}{2}}\vnu'||^2}{2} ~\leq~ \vtau \cdot S^{-1}\vnu'~\leq~ \frac{||S^{-\frac{1}{2}}\vnu'||^2}{2} \}$. The rightmost plot in \cref{app_fig:hyperplanes} illustrates the resulting partition.
\begin{align}
\int_{\mathcal{H}_3}[w(S^{-\frac{1}{2}}(\vtau-\vnu'))-w(S^{-\frac{1}{2}}\vtau)]I(\vtau + \vmu)d\vtau 
&+ \int_{\mathcal{H}_4}[w(S^{-\frac{1}{2}}(\vtau-\vnu'))-w(S^{-\frac{1}{2}}\vtau)]I(\vtau + \vmu)d\vtau\\
&\geq -\int_{\mathcal{H}_2}[w(S^{-\frac{1}{2}}(\vtau-\vnu'))-w(S^{-\frac{1}{2}}\vtau)]I(\vtau + \vmu)d\vtau.
\end{align}

Note that $\int_{\mathcal{H}_4}[w(S^{-\frac{1}{2}}(\vtau-\vnu'))-w(S^{-\frac{1}{2}}\vtau)]I(\vtau + \vmu)d\vtau \geq 0$ since  $\mathcal{H}_4 \subset \mathcal{H}_1$.
Hence, it suffices to show that
\begin{align}
\int_{\mathcal{H}_3}[w(S^{-\frac{1}{2}}(\vtau-\vnu'))-w(S^{-\frac{1}{2}}\vtau)]I(\vtau + \vmu)d\vtau 
&\geq \label{eq:stat-n}
-\int_{\mathcal{H}_2}[w(S^{-\frac{1}{2}}(\vtau-\vnu'))-w(S^{-\frac{1}{2}}\vtau)]I(\vtau + \vmu)d\vtau\\
&= \int_{\mathcal{H}_2}[-w(S^{-\frac{1}{2}}(\vtau-\vnu'))+w(S^{-\frac{1}{2}}\vtau)]I(\vtau + \vmu)d\vtau 
\end{align}
To see why the above holds, pick any point $\veta \in \mathcal{H}_3$. 
By definition of $\mathcal{H}_3$, $\veta \cdot S^{-1} \vnu'~<~-\frac{||S^{-\frac{1}{2}}\vnu'||^2}{2}$. 
As a result, it holds for the mirrored point $\vomega:=-\veta$ that $\vomega \cdot S^{-1}\vnu'~ > ~\frac{||S^{-\frac{1}{2}}\vnu'||^2}{2}$, which in turn implies $\vomega \in \mathcal{H}_2$. 
Moreover, by even symmetry of $I$ around $\vmu$, $I(\vmu+\veta) = I(\vmu-\veta)=I(\vmu+\vomega)$. 

To prove \cref{eq:stat-n} above we show that
\begin{align}
w(S^{-\frac{1}{2}}(\veta-\vnu'))-w(S^{-\frac{1}{2}}\veta) 
&\geq -w(S^{-\frac{1}{2}}(-\veta-\vnu'))+w(-S^{-\frac{1}{2}}\veta). \label{eq:symm_w} 
\end{align}

We use the even symmetry of $q_{\vzero}$ around the origin to rewrite $w(S^{-\frac{1}{2}}(-\veta-\vnu'))=w(S^{-\frac{1}{2}}(\veta+\vnu'))$ and $w(-S^{-\frac{1}{2}}\veta)= w(S^{-\frac{1}{2}}\veta)$ and arrive at
\begin{align}
w(S^{-\frac{1}{2}}(\veta-\vnu'))+w(S^{-\frac{1}{2}}(\veta+\vnu'))&\geq 2w(S^{-\frac{1}{2}}\veta). \label{eq:eq_conv_f}
\end{align}

Finally, to show \cref{eq:eq_conv_f}, we use the assumed convexity of $w$:
$\forall \vz_1, \vz_2 \in \mathbb{R}^d$, $\forall \lambda \in [0, 1]:$
$$w(\lambda \vz_1 + (1-\lambda) \vz_2) \leq \lambda w(\vz_1) + (1-\lambda)w(\vz_2).$$

Setting $\lambda = 0.5$, $\vz_1=S^{-\frac{1}{2}}(\veta-\vnu')$, $\vz_2=S^{-\frac{1}{2}}(\veta+\vnu')$, we obtain
\begin{align*}
w{\Big(}\frac{S^{-\frac{1}{2}}(\veta-\vnu')+S^{-\frac{1}{2}}(\veta+\vnu')}{2}{\Big)} = w(S^{-\frac{1}{2}}\veta)
\leq \frac{1}{2}w(S^{-\frac{1}{2}}(\veta-\vnu'))+\frac{1}{2}w(S^{-\frac{1}{2}}(\veta+\vnu')),
\end{align*}
which directly proves \cref{eq:eq_conv_f}. We have hence shown that $\forall~ \vnu': D(p, q_{\vmu+\vnu'}) - D(p, q_{\vmu}) \geq 0.$

It remains to show that this inequality is strict under the additional condition that \csuppI. 
In particular, this condition ensures that $\int_{\mathcal{H}_4}[w(S^{-\frac{1}{2}}(\vtau-\vnu'))-w(S^{-\frac{1}{2}}\vtau)]I(\vtau + \vmu)d\vtau > 0$. 
First note that $I(\vtau + \vmu)$ is centered at the origin. 
Hence, if $I$ is supported around $\vmu$, $I(\vtau + \vmu)$ is supported around $\vzero$. Moreover, for any assignment of $\vnu'$, $\vzero \in \mathcal{H}_4$ by definition. 
Hence, $\int_{\mathcal{H}_4}[w(S^{-\frac{1}{2}}(\vtau-\vnu'))-w(S^{-\frac{1}{2}}\vtau)]I(\vtau + \vmu)d\vtau > 0$. We have shown that $\forall~ \vnu': D(p, q_{\vmu+\vnu'}) - D(p, q_{\vmu}) > 0$, which completes the proof.
\qed

\subsubsection{Proof of Theorem 2}
We now use \cref{theorem:theorem_universal_convex} to prove \cref{theorem:theorem_fkl_2}. 
\FKLC*
\paragraph{Proof.} Let $w(\vx):\wkl(\vx)$ and $I(\vx):=p(\vx)$. Since \cref{theorem:theorem_fkl_2} assumes that $w$ is convex and strictly increasing in $||\vx||$, we can apply \cref{theorem:theorem_universal_convex}, which proves \cref{theorem:theorem_fkl_2}.
\qed

\subsubsection{Proof of Theorem 4}
We now use \cref{theorem:theorem_universal_convex} to prove \cref{theorem:theorem_alpha_2}. 
\AC*
\paragraph{Proof.} Let $w(\vx):=\wa(\vx)$ and $I(\vx):=p^{\alpha}(\vx)$. Note that $\forall \alpha \in \mathbb{R}$, $p^{\alpha}(\vx) \geq 0$ for any $\vx \in \text{supp}(p)$ and $\supp(p) = \supp(p^{\alpha})$. Since \cref{theorem:theorem_fkl_2} assumes that $w$ is convex and strictly increasing in $||\vx||$, we can apply \cref{theorem:theorem_universal_convex}, which proves \cref{theorem:theorem_alpha_2}.
\qed

\textbf{Notation.} The proof of \cref{theorem:theorem_alpha_2} completes the proofs for our theorems for mean recovery when $p$ is symmetric along all coordinates. For simplicity of the following analysis, we define the following two settings, which correspond to \cref{theorem:theorem_fkl_1,theorem:theorem_alpha_1} and \cref{theorem:theorem_fkl_2,theorem:theorem_alpha_2} respectively.
\begin{setting}\label{setting_1}
$w(\vx)$ is strictly convex in $\vx$.
\end{setting}

\begin{setting}\label{setting_2}
$w(\vx)$ is convex in $\vx$, strictly increasing in $||\vx||$, \csupp.
\end{setting}

\subsection{Mean recovery under partial symmetries}\label{app:partial}
We now show that our results for mean recovery extend to the case where $p$ is only even symmetric in a subset of variables $\vx_{\parset}$. We formalize the concept of \textbf{\emph{partial even symmetry}} in \cref{def:partial_symmetry}.
\begin{definition}[Partial Even Symmetry]\label{def:partial_symmetry}
We say a density $p(\vx_{\parset}, \vx_{\notparset})$ is partially even symmetric in $\vx_{\parset}$ with a constant point of even symmetry $\vmu_{\parset}$ if 
$p(\vmu_{\parset} + \vx_{\parset}|\vx_{\notparset}) = p(\vmu_{\parset}-\vx_{\parset}| \vx_{\notparset})$ for any assignment of $\vx_{\notparset}$ and $\vmu_{\parset}$ does not depend on $\vx_{\notparset}$.
\end{definition}

The proof uses the concept of \textbf{\emph{odd symmetry}}, which is introduced below.
\begin{definition}[Odd Symmetry]\label{def:odd}
We say a function $h: \mathbb{R}^d \rightarrow \mathbb{R}$ is \textbf{odd symmetric} around $\vnu \in \mathbb{R}^d$ if for all $\vx \in \mathbb{R}^d$,
$h(\vnu + \vx)= -h(\vnu - \vx).$
\end{definition}

\cref{theorem:theorem_theorem_mean_partial} states our result for mean recovery under partial symmetries. We again assume that the divergence itself is well-defined for any considered value of $\vnu$.

\begin{theorem}\label{theorem:theorem_theorem_mean_partial}
Let $q_{\vnu}$ be a member of a location family with fixed scale matrix $S$. Assume that $q_{\vzero}$ is even symmetric around $\vzero$ and $S$ has a block-diagonal structure, such that $S =\begin{pmatrix}
S_{\parset,\parset}  & \vzero \\
\vzero & S_{\notparset,\notparset}
 \end{pmatrix}$. Let $I(\vx_{\parset},\vx_{\notparset})$ be a density that is even symmetric in $\vx_{\parset}$ with a constant point of partial even symmetry $\vmu_{\parset}$. Further, let $w = h\circ q_{\vzero}$ with $h:\mathbb{R}^+_0 \rightarrow \mathbb{R}$. Then, $D(p, q_{\vnu}) = \int w(S^{-\frac{1}{2}}(\vx-\vnu))I(\vx)d\vx$ is minimized at $\vnu_{\parset} = \vmu_{\parset}$ for any fixed $\vnu_{\notparset}$ if either
 \begin{itemize}
    \item \textbf{(SP1)}: $w$ is strictly convex
    \item \textbf{(SP2)}: $w$ is convex, strictly increasing in $||\vx||$ and $I(\vx_{\parset}|\vx_{\notparset})$ is supported around $\vmu_{\parset}$ on a set of positive measure for every $\vx_{\notparset}$
 \end{itemize}
\end{theorem}

Note that \cref{theorem:theorem_theorem_mean_partial} immediately gives results for the FKL by setting $w := \wkl$ and $I:= p$ and for the $\alpha$-DIV via $w:=\wa$ and $I:=\frac{p^\alpha}{Z}$, where $Z > 0$ is the normalizing constant of $p^{\alpha}$. Note that $Z$ in does not impact optimization and is merely introduced to ensure $I=p^{\alpha}$ is a valid density that we can condition. Below, we prove \cref{theorem:theorem_theorem_mean_partial}.

\paragraph{Proof.} 
We start by investigating \textbf{(SP1)}. First of all, note that when $w$ is strictly convex, we can conclude that $D(p, q_{\vnu}) 
= \int w(S^{-\frac{1}{2}}(\vx-\vnu))I(\vx)d\vx$ is strictly convex in $\vnu$, irregardless of the symmetry assumptions on $I$, via \cref{lemma:strict_composition} and \cref{lemma:strict_integral} (see the proofs for \cref{theorem:theorem_fkl_1} and \cref{theorem:theorem_alpha_1} in \cref{proof:theorem_1_3} for more detail). 
To show that this implies that $D(p, q_{\vnu})$ is minimized by setting $\vnu_{\parset} = \vmu_{\parset}$, we follow \citet{margossian2025generalized} and show that $\nabla{\vnu_{\parset}} D(p, q_{\vnu})=0$ when $\vnu_{\parset} = \vmu_{\parset}$, for any $\vnu_{\notparset}$. First, recall that we can rewrite the FKL and $\alpha$-DIV as below, since they are $f$-divergences:
\begin{align}
D(p,q_{\vnu}) &= \int\varphi \left(\log\left(\frac{p(\vx)}{q_{\vnu}(\vx)}\right)\right)q_{\vnu}(\vx)d\vx \\
& = \int\varphi \left(\log\left(\frac{p(\vzeta+\vnu)}{q_{\vzero}(S^{-\frac{1}{2}}\vzeta)|S^{-\frac{1}{2}}|}\right)\right)q_{\vzero}(S^{-\frac{1}{2}}\vzeta)|S^{-\frac{1}{2}}|d\vzeta 
\end{align}
where $\vzeta = \vx - \vnu$ and $\varphi(\vu) := \vu\exp(\vu)$ for the FKL and $\varphi(\vu) = \frac{\exp(\vu^{\alpha})-1}{\alpha(\alpha-1)}$ for the $\alpha$-DIV \citep{margossian2025generalized}.
Taking the derivative of $D(p,q_{\vnu})$ w.r.t. $\vnu_{\parset}$, we have \citep{margossian2025generalized}
\begin{align*}
\nabla_{\vnu_{\parset}}D(p,q_{\vnu}) 
& = \int\int \frac{[\nabla_{\vzeta_{\parset}}p(\vzeta + \vnu)]}{p(\vzeta +\vnu)} \varphi' \left(\log\left(\frac{p(\vzeta+\vnu)}{q_{\vzero}(S^{-\frac{1}{2}}\vzeta)|S^{-\frac{1}{2}}|}\right)\right)q_{\vzero}(S^{-\frac{1}{2}}\vzeta)|S^{-\frac{1}{2}}|d\vzeta_{\parset}d\vzeta_{\notparset}.
\end{align*}
Now set $\vnu_{\parset}=\vmu_\parset$, while keeping $\vnu_{\notparset}$ fixed to some arbitrary value. Then, $p(\vzeta + \vnu)$ is even symmetric in $\vzeta_{\parset}$ around the origin for any $\vzeta_{\notparset}$ and hence $\nabla_{\vzeta_{\parset}}p(\vzeta + \vnu)$ is odd symmetric around the origin. Further, note that when $S$ is block-diagonal, as stated in our assumptions, it holds that $q_{\vzero}(S^{-\frac{1}{2}}\vzeta)$ is even-symmetric around the origin in $\vzeta_{\parset}$ for any fixed  $\vzeta_{\notparset}$. Hence, all terms in the integrand except $\nabla_{\vzeta_{\parset}}p(\vzeta + \vnu)$ are even symmetric around $0$ in $\vzeta_{\parset}$, resulting in an odd symmetric integrand and vanishing derivative at $\vnu_{\parset}=\vmu_\parset$. Hence, $\vnu_{\parset}=\vmu_\parset$ is a stationary point for all assignments of $\vnu_{\notparset}$. Combined with the strict convexity of $D(p,q_{\vnu})$, we conclude that $D(p,q_{\vnu})$ is minimized when $\vnu_{\parset}=\vmu_{\parset}$, which completes the proof for \textbf{(SP1)}.

For \textbf{(SP2)}, we closely follow our proof of \cref{theorem:theorem_universal_convex} and consider the difference in divergence values for alternative assignments of the location parameter. Let $\vu$ be a vector such that $\vu_{\parset}=\vmu_{\parset}$ and $\vu_{\notparset} = \vnu_{\notparset}$ for some fixed $\vnu_{\notparset}$, i.e. $\vu =(
\vmu_{\parset},
\vnu_{\notparset})^T$. We now consider the alternative assignment $\vu+\vnu'$, where $\vnu'=
(\vnu'_{\parset},
\vzero)^T$, 
leading to the difference in divergence values $D(p, q_{\vu+\vnu'}) - D(p, q_{\vu})$, which is given as
\begin{align}
        &\int\int \left[w\left(S^{-\frac{1}{2}}(\vx-\vu-\vnu')\right) - w\left(S^{-\frac{1}{2}}(\vx-\vu) \right)\right]I(\vx_{\parset}|\vx_{\notparset})d\vx_{\parset}~I(\vx_{\notparset})d\vx_{\notparset} \\
        &=\int\int \left[w\left(S^{-\frac{1}{2}}(\vtau-\vnu')\right) - w\left(S^{-\frac{1}{2}}\vtau \right)\right]I(\vtau_{\parset} + \vu_{\parset}|\vtau_{\notparset} + \vu_{\notparset})d\vtau_{\parset}~I(\vtau_{\notparset} + \vu_{\notparset})d\vtau_{\notparset}
\end{align}
where we set $\vtau := \vx - \vu$ and use that $I(\vx_{\parset},\vx_{\notparset})=I(\vx_{\parset}|\vx_{\notparset})I(\vx_{\notparset})$.
We want to show that the above is strictly positive for any assignment of $\vnu'_{\parset}$ and $\vnu_{\notparset}$ (where the latter is equivalent to $\vu_{\notparset}$ by definition of $\vu$). We proceed in two steps: First, we show that the inner integral is strictly positive for any fixed value of $\vtau_{\notparset}$. Then, we argue that the outer integral preserves strict positivity which completes the proof.

\paragraph{Step 1.} First, we show that for the inner integral, it holds that $\mathcal{I}_1(\vtau_{\notparset}) := \int \left[w\left(S^{-\frac{1}{2}}(\vtau-\vnu')\right) - w\left(S^{-\frac{1}{2}}\vtau \right)\right]I(\vtau_{\parset} + \vu_{\parset}|\vtau_{\notparset} + \vu_{\notparset})d\vtau_{\parset} > 0$ for any fixed $\vtau_{\notparset}$. We define $\mathcal{A}(\vtau_{\notparset}) := \{(\vz, \vtau_{\notparset})^T: \vz\in \mathbb{R}^{|\parset|}\}.$ Akin to the proof of \cref{theorem:theorem_universal_convex}, we define $\mathcal{H}_2:=\{\vtau \in \mathcal{A}(\vtau_{\notparset}): \vtau \cdot S^{-1}\vnu' > \frac{||S^{-\frac{1}{2}}\vnu'||^2}{2}\}$, $\mathcal{H}_3:=\{\vtau \in \mathcal{A}(\vtau_{\notparset}): \vtau \cdot S^{-1}\vnu'<-\frac{||S^{-\frac{1}{2}}\vnu'||^2}{2}\}$ and $\mathcal{H}_4:=\{\vtau \in \mathcal{A}(\vtau_{\notparset}):~ -\frac{||S^{-\frac{1}{2}}\vnu'||^2}{2} ~\leq~ \vtau \cdot S^{-1}\vnu'~\leq~ \frac{||S^{-\frac{1}{2}}\vnu'||^2}{2}\}$ to split the domain of integration. 
By construction, for any $\vtau \in \mathcal{H}_2$, we have $w\left(S^{-\frac{1}{2}}(\vtau-\vnu')\right) - w\left(S^{-\frac{1}{2}}\vtau \right) < 0$, while for any $\vtau \in \mathcal{H}_3 \cup\mathcal{H}_4$, it holds that $w\left(S^{-\frac{1}{2}}(\vtau-\vnu')\right) - w\left(S^{-\frac{1}{2}}\vtau \right) \geq 0$. Further, note that due to the block-diagonal structure of $S$ and the fact that $\vnu'_{\notparset} = \vzero$, whether a point $(\vtau_{\parset}, \vtau_{\notparset})^T$ is assigned to $\mathcal{H}_2$, $\mathcal{H}_3$ or $\mathcal{H}_4$ does not depend on $\vtau_{\notparset}$.

We now show that, under our assumptions, it holds that 
\begin{align}
&\int_{\mathcal{H}_3}[w(S^{-\frac{1}{2}}(\vtau-\vnu'))-w(S^{-\frac{1}{2}}\vtau)]I(\vtau_{\parset} + \vu_{\parset}|\vtau_{\notparset} + \vu_{\notparset})d\vtau_{\parset}\\
&\geq \int_{\mathcal{H}_2}[-w(S^{-\frac{1}{2}}(\vtau-\vnu'))+w(S^{-\frac{1}{2}}\vtau)]I(\vtau_{\parset} + \vu_{\parset}|\vtau_{\notparset} + \vu_{\notparset})d\vtau_{\parset},
\end{align}
which implies $\mathcal{I}_1(\vtau_{\notparset}) \geq 0$. To see why the above holds, pick any $\veta:=(\veta_{\parset}, \vtau_{\notparset}) \in \mathcal{H}_3$ and set $\vomega := (-\veta_{\parset}, \vtau_{\notparset})\in \mathcal{H}_2$. By even symmetry of $I$ around $ \vu_{\parset}$, we have $I(\veta_{\parset} + \vu_{\parset}|\vtau_{\notparset} + \vu_{\notparset}) = I(\vomega_{\parset} + \vu_{\parset}|\vtau_{\notparset} + \vu_{\notparset})$. Hence, to show that the above inequality holds, it is sufficient to show
 \begin{align}
w(S^{-\frac{1}{2}}(\veta-\vnu'))-w(S^{-\frac{1}{2}}\veta) \geq -w(S^{-\frac{1}{2}}(\vomega-\vnu'))+w(S^{-\frac{1}{2}}\vomega)\label{eq:partial_sym_to_show}.
 \end{align}
Note that our assumptions on $w$ imply that $w$ is spherically symmetric, i.e., $w(\vx)=w(\vx')$ whenever $||\vx||=||\vx'||$. Due to the structure of $S$ and $\vnu'$, it holds that $||S^{-\frac{1}{2}}(\vomega-\vnu')||=||S^{-\frac{1}{2}}(-\veta-\vnu')||$ and $||S^{-\frac{1}{2}}\vomega||=||-S^{-\frac{1}{2}}\veta||$. Hence, we can replace $\vomega$ in \cref{eq:partial_sym_to_show} with $-\veta$, resulting in
\begin{align}
w(S^{-\frac{1}{2}}(\veta-\vnu'))-w(S^{-\frac{1}{2}}\veta) \geq -w(S^{-\frac{1}{2}}(-\veta-\vnu'))+w(-S^{-\frac{1}{2}}\veta),
\end{align}
which we have already shown to hold under our assumptions on $w$ in our proof of \cref{theorem:theorem_universal_convex} (see \cref{eq:symm_w} onward). Taken altogether, we conclude that $\mathcal{I}_{1}(\vtau_{\notparset}) \geq 0$ for any $\vtau_{\notparset}$, any $\vnu'_{\parset}$ and any $\vu_{\notparset}$. To establish $\mathcal{I}_{1}(\vtau_{\notparset}) > 0$, we argue that the contribution of $\mathcal{H}_4$ to the difference of divergence values is strictly positive since $I(\vtau_{\parset} + \vu_{\parset}|\vtau_{\notparset} + \vu_{\notparset})$ is supported around $\vtau_{\parset}=\vzero$ (along $\parset$) on a set of non-zero measure by assumption, and we have $(\vzero, \vtau_{\notparset})^T \in \mathcal{H}_4$ by definition of $\mathcal{H}_4$. By the same argument as in \cref{theorem:theorem_universal_convex}, we hence conclude $\int_{\mathcal{H}_4}[w(S^{-\frac{1}{2}}(\vtau-\vnu'))-w(S^{-\frac{1}{2}}\vtau)]I(\vtau_{\parset} + \vu_{\parset}|\vtau_{\notparset} + \vu_{\notparset})d\vtau_{\parset} > 0$. Taken altogether, we conclude $\mathcal{I}_{1}(\vtau_{\notparset}) > 0$ for any $\vtau_{\notparset}$, any $\vu_{\notparset}$, and any $\vnu'_{\parset}$.

 \paragraph{Step 2.} Since $I(\vtau_{\notparset} + \vu_{\notparset}) \geq 0$ and $I$ is a density (and hence is not $0$ everywhere), we have $D(p, q_{\vnu'+\vu}) - D(p, q_{\vu}) =\int \mathcal{I}_1(\vtau_{\parset})~I(\vtau_{\notparset} + \vu_{\notparset})d\vtau_{\notparset} > 0$.
for any $\vnu'_{\parset}, \vu_{\notparset}$, which completes the proof for \textbf{(SP2)}. \qed

\subsection{Recovery of correlations}\label{app:corr}
We begin by restating the assumptions used throughout our analysis of exact recovery for the correlation matrix.
\ASSUMPTIONSC*

\FKLCORR*
\ACORR*

Our proofs for exact recovery of correlations use the concept of \emph{odd symmetry}, as introduced in \cref{def:odd}.

We now  provide proofs for \emph{exact recovery of the correlation matrix} for FKL and $\alpha$-DIVs when the sufficient conditions of our theorems are fulfilled. Our proofs closely follow the proof provided by \citet{margossian2024variational, margossian2025generalized} for the RKL. We structure each proof in two parts: In \textbf{Part 1}, we show that the objective has a unique minimizer in $S$. In \textbf{Part 2}, we prove that this unique minimizer is proportional to the correct correlation matrix.

\subsubsection{FKL}\label{app:fkl_corr}
We now provide the proof for \emph{exact recovery of the correlation matrix} via the FKL (\cref{theorem:theorem_fkl_corr}).

\paragraph{Part 1.} First, we rewrite $D_{\text{FKL}}$ as follows
\begin{align}
    D_{\text{FKL}}(p, q_{\vnu, S}) &:= \int \log\left(\frac{p(\vx)}{q_{\vnu, S}(\vx)}\right)p(\vx)d\vx\\
    &= - H(p) - \int \log \left(q_{\vnu, S}(\vx)\right) p(\vx)d\vx\\
    &= -H(p) - \int \log \left( q_0{\Big(}S^{-\frac{1}{2}}\left(\vx-\vnu\right){\Big)}|S|^{-\frac{1}{2}}\right) p(\vx)d\vx\\
     &= C - \log\left(|S|^{-\frac{1}{2}}\right) + \int -\log \Big( q_0{\Big(}S^{-\frac{1}{2}}(\vx-\vnu){\Big)}\Big) p(\vx)d\vx, \label{eq:fkl_corr}
\end{align}
where $C$ summarizes constants in $\vnu$ and $S$.

Note that for both \cref{setting_1} and \cref{setting_2} \wkl is convex, and $D_{\text{FKL}}(p, q_{\vnu, S})$ has a unique minimizer w.r.t. $\vnu$ for any value of $S$, as guaranteed by our theorems for mean recovery (\cref{theorem:theorem_fkl_1}, \cref{theorem:theorem_fkl_2}). Moreover, $S$ is positive definite, and therefore so is $S^{-\frac{1}{2}}$ \citep{bhatia2009positive}. We will now show that $ D_{\text{FKL}}(p, q_{\vnu, S})$ is strictly convex in $S^{-\frac{1}{2}}$, which implies a unique minimizer in $S$, as $S$ is invertible.

It is well-known that $f(A) = - \log(|A|)$ is strictly convex on $A \in \mathbb{S}_{++}^d$ \citep[Ch. 7]{horn2012matrix}. 
Hence, $-\log(|S|^{-\frac{1}{2}})$ is strictly convex in $S^{-\frac{1}{2}}$. Now define $g: \mathbb{S}_{++}^d \rightarrow \mathbb{R}^d$ as the affine map $g(S^{-\frac{1}{2}}) := S^{-\frac{1}{2}}(\vx-\vnu)$. Since convexity is preserved under composition with affine maps \citep[Ch. 3.2.2]{boyd2004convex}, we conclude that $\wkl(g(S^{-\frac{1}{2}}))$ is convex in $S^{-\frac{1}{2}}$. Since \cref{eq:fkl_corr} is a sum of (1) a strictly convex function in $S^{-\frac{1}{2}}$ and a (2) convex function in $S^{-\frac{1}{2}}$, the overall objective is strictly convex in $S^{-\frac{1}{2}}$. Since $S^{-\frac{1}{2}}$ is invertible this corresponds to a unique minimizer in $S$ at its stationary point. Note that this partial result assumes neither log-concavity of $p$ nor elliptical symmetry of $p$. 

\paragraph{Part 2.} It remains to show that the unique minimizer of $D_{\text{FKL}}(p, q_{\vnu, S})$ w.r.t. $S$ recovers the correct correlation matrix, i.e., $S = \gamma^2M$ for some $\gamma > 0$. Below, we exploit \emph{elliptical symmetry} of $p$, using that under this assumption, $p$ can be written as in \cref{eq:p_ell}.
\begin{align}
D_{\text{FKL}}(p, q) 
&= \int\log \left(\frac{p(\vx)}{q_{\vnu, S}(\vx)}\right) p(\vx)d\vx\\
&=\int \log\left(p_\vzero\left(M^{-\frac{1}{2}}(\vx-\vmu)\right)|M^{-\frac{1}{2}}|\right) p_\vzero\left(M^{-\frac{1}{2}}(\vx-\vmu)\right)|M^{-\frac{1}{2}}| d\vx \\
&-\int \log \left( q_{\vzero} \left(S^{-\frac{1}{2}}(\vx-\vnu)\right)|S|^{-\frac{1}{2}}\right) p_\vzero\left(M^{-\frac{1}{2}}(\vx-\vmu)\right)|M^{-\frac{1}{2}}|d\vx\\
\label{eq:entropy}
&=\log\left(|M^{-\frac{1}{2}}|\right) + \int \log\left(p_\vzero\left(M^{-\frac{1}{2}}(\vx-\vmu)\right)\right) p_\vzero\left(M^{-\frac{1}{2}}(\vx-\vmu)\right)|M^{-\frac{1}{2}}| d\vx \\\label{eq:entropy_2}
&-\log\left(|S|^{-\frac{1}{2}}\right) -\int \log \left( q_\vzero\left(S^{-\frac{1}{2}}(\vx-\vnu)\right)\right) p_\vzero\left(M^{-\frac{1}{2}}(\vx-\vmu)\right)|M^{-\frac{1}{2}}|d\vx
\end{align}
We now set $\vzeta := M^{-\frac{1}{2}}(\vx-\vmu)$ (and hence $\vx = M^{\frac{1}{2}}\vzeta + \vmu$). Moreover, since we are guaranteed to recover the mean for any $S$ by either \cref{theorem:theorem_fkl_1} or \cref{theorem:theorem_fkl_2},  
we will assume $\vnu = \vmu$ from now on. Therefore, \cref{eq:entropy}, via change of variables to $\vzeta$, simplifies to 
\begin{align*}
(\ref{eq:entropy}) &=\log\left(|M^{-\frac{1}{2}}|\right) + \int \log\left(p_\vzero\left(M^{-\frac{1}{2}}(M^{\frac{1}{2}}\vzeta + \vmu -\vmu)\right)\right) p_\vzero(\vzeta)|M^{-\frac{1}{2}}||M^{\frac{1}{2}}| d\vzeta \\
&=\log\left(|M^{-\frac{1}{2}}|\right) + \int \log\left(p_\vzero\left(\vzeta\right)\right) p_\vzero\left(\vzeta\right)d\vx = \log\left(|M^{-\frac{1}{2}}|\right) - H(p_{\vzero})
\end{align*}
Similarly, for \cref{eq:entropy_2}, we obtain
\begin{align*}
(\ref{eq:entropy_2}) &= -\log\left(|S|^{-\frac{1}{2}}\right) -\int \log \left( q_\vzero\left(S^{-\frac{1}{2}}M^{\frac{1}{2}}\vzeta\right)\right) p_\vzero(\vzeta)d\vzeta
\end{align*}
We now define $J:= S^{-\frac{1}{2}}M^{\frac{1}{2}}$. Note that $\log\left(|M^{-\frac{1}{2}}|\right)-\log\left(|S|^{-\frac{1}{2}}\right)=\log\left(|M|^{-\frac{1}{2}}|S|^{\frac{1}{2}}\right) = -\log\left(|M^{\frac{1}{2}}S^{-\frac{1}{2}}| \right) = -\log\left(|S^{-\frac{1}{2}}M^{\frac{1}{2}}| \right)= -\log\left(|J|\right)$. Plugging this in we obtain
\begin{align}
D_{\text{FKL}}(p, q_{\vnu, S}) 
&= - H(p_{\vzero}) -\log(|J|) -\int \log  q_{\vzero}\left(J\vzeta\right) p_\vzero\left(\vzeta\right)d\vzeta 
\label{eq:time_for_spheres}
\end{align}

Remember that $M^{\frac{1}{2}}$ is fixed and invertible. 
We want to prove that we have a stationary point at $J= \gamma I$, for some $\gamma > 0$, since this implies $S^{\frac{1}{2}} = \frac{1}{\gamma} M^{\frac{1}{2}}$ and we have successfully recovered the correlation matrix. Since $p_\vzero$ and $q_\vzero$ are spherically symmetric, we can define them via functions 
depending on the norm of their inputs, $g: \mathbb{R}^+_0\rightarrow \mathbb{R}$ and $f:\mathbb{R}^+_0\rightarrow \mathbb{R}^+_0$, given as
\begin{align*}
g(||J\vzeta||) &= \log q_{\vzero}(J\vzeta)\\
f(||\vzeta||) &= p_{\vzero}(\vzeta).
\end{align*}
Plugging this into \cref{eq:time_for_spheres}, we arrive at
\begin{align}
D_{\text{FKL}}(p, q_{\vnu, S}) 
&= - H(p_{\vzero})-\log(|J|) -\int g\left(||J\vzeta||\right) f\left(||\vzeta\right||)d\vzeta.\label{eq:diff_J}
\end{align}
Differentiating w.r.t. $J$ and using $\nabla_J |J|= |J|J^{-T}$ as well as $\nabla_J||J\vzeta|| = \frac{J\vzeta\vzeta^T}{||J\vzeta||}$ \citep{petersen2008matrix,margossian2024variational}, we obtain 
\begin{align}
\nabla_{J}D_{\text{FKL}}(p, q) &=
-J^{-T} -\int  g'(||J\vzeta||)\frac{J\vzeta\vzeta^T}{||J\vzeta||} f\left(||\vzeta||\right)d\vzeta.
\end{align}
We want this gradient to be $0$ when setting $J= \gamma I$ for some $\gamma > 0$, or equivalently,
\begin{align}
\gamma^{-1} I &=-\int f\left(||\vzeta||\right) g'(||\gamma \vzeta||)\frac{\gamma \vzeta\vzeta^T}{||\gamma\vzeta||}d\vzeta \tag{$J = \gamma I$}\\
\gamma^{-1} I &=-\int f\left(||\vzeta||\right) g'(||\gamma\vzeta||)\frac{ \vzeta\vzeta^T}{||\vzeta||}d\vzeta \label{eq:time_for_traces}
\end{align}
We need to show that the r.h.s. exhibits the desired diagonal structure. Looking at the $i$-th component of integration, we have
\begin{align}
-\int d\vzeta_{\setminus\{i\}}\int d\vzeta_i~\left[f\left(||\vzeta||\right) g'(||\gamma\vzeta||)\right]\frac{ \vzeta_i\vzeta_j}{||\vzeta||}.
\end{align}
The above closely matches Eq. 26 in \citep{margossian2024variational} (with the main difference being $f$ and $g$ being ``swapped'' as $p$ and $q$ take different places in the FKL and the RKL) 
and we can apply the same reasoning: For $i\neq j$, the inner integrand is odd-symmetric in $\vzeta_i$ and hence the integral vanishes. If $i=j$, the spherical symmetry of the bracketed term ensures that we obtain the same result for every $i$. Hence, the overall result will be a matrix with (1) all zeros off-diagonal and (2) all equal values on the diagonal.

It remains to show that there is a solution for $\gamma$ that fulfills \cref{eq:time_for_traces}. Following \citep{margossian2024variational}, we show this by equating the traces the l.h.s. and r.h.s. of
\cref{eq:time_for_traces}:
\begin{align}
d\gamma^{-1} 
&=-\int \left[f\left(||\vzeta||\right) g'(\gamma||\vzeta||)\right]||\vzeta||d\vzeta. \label{eq:time_for_r}
\end{align}
Now, let $r:=||\vzeta||$ and $\vzeta=r\vw$ for $\vw$ being a \emph{direction in the unit sphere}, denoted as $\vw \in \mathcal{S}^{d-1}$.
We now change the variable of integration to $r$ on the r.h.s. of \cref{eq:time_for_r} \citep[Ch. 2]{folland1999real}. Below, $\sigma(w)$ denotes the surface measure on the unit sphere.
\begin{align}
-\int \left[f\left(||\vzeta||\right) g'(\gamma||\vzeta||)\right]||\vzeta||d\vzeta&=-\int_{r \in (0, \infty)}\int_{w \in \mathcal{S}^{d-1}} [f\left(r\right) g'(\gamma r)]rr^{d-1}d\sigma(w)dr\\
&=-\int_{r \in (0, \infty)} \int_{w \in \mathcal{S}^{d-1}} [f\left(r\right) g'(\gamma r)]r^dd\sigma(w)dr\\\label{eq:vol_sphere}
&=-\frac{2\pi^{\frac{d}{2}}}{\Gamma(\frac{d}{2})}\int_{r \in (0, \infty)} [f\left(r\right) g'(\gamma r)]r^ddr,
\end{align}
where \cref{eq:vol_sphere} uses that $\int_{w \in \mathcal{S}^{d-1}}d\sigma(w)=\frac{2\pi^{\frac{d}{2}}}{\Gamma(\frac{d}{2})}$.

We hence need to show that there exists a solution for $\gamma$ in \cref{eq:last_eq_gamma}:
\begin{equation}
d\gamma^{-1}  = \frac{2\pi^{\frac{d}{2}}}{\Gamma(\frac{d}{2})}\int_{r \in (0, \infty)} [-g'(\gamma r)f\left(r\right)]r^ddr. \label{eq:last_eq_gamma}
\end{equation}
Note that the l.h.s. is a non-negative, strictly decreasing function in $\gamma$, since $\gamma > 0$. 
For a unique solution of $\gamma$ to exist, we need the r.h.s. to be positive and increasing in $\gamma$.

\cref{setting_1}. Remember that $-g(\gamma r) = -\log q_\vzero(\gamma \vzeta)$ for $\vzeta = r\vw$. By assumption of \cref{setting_1}, $\wkl$ is strictly convex in its argument. This implies that $-g'$ is strictly increasing. For $\gamma > 0$ and fixed $r >0$, $-g'(\gamma r)$ is hence increasing in $\gamma$. Note that $g'$ cannot be $0$ everywhere since \cref{setting_1} requires $\wkl$ strictly convex. Moreover, all involved terms are positive: We have $-g'(0) \geq 0$ by spherical symmetry around the origin and convexity (or at least \limc when the derivative at the origin is ill-defined), and we just established that $-g'$ is increasing on $(0, \infty)$. Moreover, $f\left(r\right) \geq 0$ and $r^d>0$ since $r > 0$. Hence, the r.h.s. is indeed a positive, increasing function in $\gamma$. We hence conclude that there is a unique solution for $\gamma$ w.r.t. \cref{eq:last_eq_gamma}.

\cref{setting_2}. We can apply the same reasoning for \cref{setting_2}. The main difference in this setting is that $-g'$ is only increasing, not necessarily strictly. Note that we also assume in \cref{setting_2} that \wkl is strictly increasing in $||\vx||$, hence $g'$ is not $0$ everywhere. Hence, the same line of argumentation still holds.

Since we can find a solution for $\gamma$, we conclude that under both \cref{setting_1} and \cref{setting_2}, the unique minimizer of $D_{\text{FKL}}(p, q_{\vnu, S})$ w.r.t. $S$ established in \textbf{Part 1} accurately captures the correlations of the target.

\subsubsection{$\alpha$-DIV}\label{app:alpha_corr}
We now proceed to prove exact recovery of the target correlations under our conditions when optimizing the $\alpha$-DIV (\cref{theorem:theorem_alpha_corr}).
\begin{align}
\label{eq:first_alpha}
D_{\alpha}(p,q_{\vnu, S})
&= \frac{1}{\alpha(\alpha-1)}\int {\Big[}\Big{(}\frac{p(\vx)}{q_{\vnu, S}(\vx)}\Big{)}^{\alpha}-1{\Big]}q_{\vnu, S}(\vx) d\vx\\
&=\int \frac{1}{\alpha(\alpha-1)}q_{\vnu, S}^{1-\alpha}(\vx) p^{\alpha}(\vx)d\vx - \frac{1}{\alpha(\alpha-1)}\\
&=\int \frac{1}{\alpha(\alpha-1)}\left(q_\vzero(S^{-\frac{1}{2}}(\vx-\vnu))|S|^{-\frac{1}{2}}\right)^{1-\alpha} p^{\alpha}(\vx)d\vx - \frac{1}{\alpha(\alpha-1)}\\
&=\int \frac{1}{\alpha(\alpha-1)}\left(q_\vzero(S^{-\frac{1}{2}}(\vx-\vnu))\right)^{1-\alpha}|S^{-\frac{1}{2}}|^{1-\alpha} p^{\alpha}(\vx)d\vx + C,
\end{align}
where $C$ summarizes constants in $\vnu$ and $S$.

\paragraph{Part 1.} We first show that $D_{\alpha}(p,q_{\vnu, S})$ has a unique minimizer w.r.t. $S$. For simplicity, we define
\begin{align*}
w(S^{-\frac{1}{2}}) &:= \frac{1}{\alpha(\alpha-1)}\left(q_\vzero(S^{-\frac{1}{2}}(\vx-\vnu))\right)^{1-\alpha}\\
h(S^{-\frac{1}{2}}) &:= |S^{-\frac{1}{2}}|^{1-\alpha}\\
A &:= S^{-\frac{1}{2}}.
\end{align*}
We aim to show that $w(A)h(A)$ is strictly convex in $A$, which implies that $D_{\alpha}(p,q_{\vnu, S})$ is strictly convex in $S^{-\frac{1}{2}}$ via \cref{lemma:strict_integral}.

First, note that $\log|A|$ is strictly concave on $A \in \mathbb{S}_{++}^d$ \citep[Ch. 7]{horn2012matrix}. 
As a result, $(1-\alpha)\log|A|$ is strictly convex for $\alpha > 1$. Since strict convexity is preserved under composition with $\exp(\cdot)$, we conclude that $h(A) = |A|^{1-\alpha}= \exp((1-\alpha)\log|A|)$ is strictly convex in $A$ for $\alpha > 1$. Turning to $w(A)$, we know that it is convex in $A$ for both \cref{setting_1} and \cref{setting_2}, since $A(\vx-\vnu)$ is affine in $A$ (same reasoning as in \cref{app:fkl_corr}). 

We still need to show that the product $w(A)h(A)$ is strictly convex in $A$ for $\alpha > 1$. 
Since we assume $\alpha > 1$ (and therefore $1-\alpha < 0$) and further that $-\log q_{\vzero}$ is convex, $w(A)$ is log-convex.
Since $h(A)$ is strictly log-convex, $w(A)h(A)$ is also strictly log-convex, as (strict) log-convexity is preserved under multiplication \citep[Ch. 3.5.2.]{boyd2004convex}. Moreover, every strictly log-convex function is strictly convex (since composition with $\exp(\cdot)$ preserves convexity). Hence $w(A)h(A)$ is strictly convex and strict convexity of the integral is preserved via \cref{lemma:strict_integral}. Note that this result is valid for both \cref{setting_1} and \cref{setting_2} as we do not require strict convexity of $w(A)$. 
We conclude that under our conditions, $D_{\alpha}(p, q_{\vnu, S})$ is strictly convex in $S^{-\frac{1}{2}}$ and hence has a unique minimizer in $S$.

\paragraph{Part 2.} Akin to the FKL, we now show that $D_{\alpha}(p, q_{\vnu, S})$ has a stationary point at $S = \gamma M$ for some $\gamma >0$, such that we correctly recover correlations. 
Note that optimizing $D_{\alpha}$ as in \cref{eq:first_alpha} is equivalent to optimizing the \emph{R\'enyi} $\alpha$-divergence \citep{renyi1961measures, li2016renyi}, given as
\begin{align}
 D_{\alpha}^R(p, q_{\vnu, S}) = \frac{1}{\alpha-1}\log\int p^{\alpha}(\vx) q^{1-\alpha}(\vx)d\vx,
\end{align}
since $D_{\alpha}(p, q) = \frac{1}{\alpha(\alpha-1)}\exp((\alpha-1)D_{\alpha}^R-1)$. Therefore, we will below consider optimization w.r.t. to the R\'enyi-$\alpha$ divergence instead, as it facilitates exposition. The proof closely follows the steps performed for the FKL (\cref{app:fkl_corr}).
\begin{align}
D_{\alpha}^R(p, q_{\vnu, S}) 
&= \frac{1}{(\alpha-1)}\log\int q_{\vnu, S}^{1-\alpha}(\vx) p^{\alpha}(\vx)d\vx\\
&= \frac{1}{(\alpha-1)}\log\int q_{\vnu, S}^{1-\alpha}(\vx) (p_{\vzero}(M^{-\frac{1}{2}}(\vx-\vmu))|M^{-\frac{1}{2}}|))^{\alpha}d\vx\\
&=\frac{1}{(\alpha-1)}\log\int \left(q_{\vzero}(S^{-\frac{1}{2}}(\vx-\vnu))|S^{-\frac{1}{2}}|)\right)^{1-\alpha} \left(p_{\vzero}(M^{-\frac{1}{2}}(\vx-\vmu))|M^{-\frac{1}{2}}|)\right)^{\alpha}d\vx\\
&=\frac{1}{(\alpha-1)}\log \int \left(q_{\vzero}(S^{-\frac{1}{2}}(\vx-\vnu))\right)^{1-\alpha}\left(p_{\vzero}(M^{-\frac{1}{2}}(\vx-\vmu))\right)^{\alpha} |S^{-\frac{1-\alpha}{2}}M^{-\frac{\alpha}{2}}|d\vx
\end{align}

We now set $\vzeta := M^{-\frac{1}{2}}(\vx-\vmu)$ (and hence $\vx = M^{\frac{1}{2}}\vzeta + \vmu$). Moreover, since we are guaranteed to recover the mean for any $S$, we will assume $\vnu = \vmu$ from now on. 
\begin{align}
D_{\alpha}^R(p, q) 
&= \frac{1}{(\alpha-1)} \log \int \left(q_{\vzero}(S^{-\frac{1}{2}}M^{\frac{1}{2}}\vzeta)\right)^{1-\alpha}\left(p_{\vzero}(\vzeta)\right)^{\alpha} |S^{-\frac{1-\alpha}{2}}M^{-\frac{\alpha}{2}}||M^{1/2}|d\vzeta\\
&= \frac{1}{(\alpha-1)} \log \int \left(q_{\vzero}(S^{-\frac{1}{2}}M^{\frac{1}{2}}\vzeta)\right)^{1-\alpha}\left(p_{\vzero}(\vzeta)\right)^{\alpha} |S^{-\frac{1-\alpha}{2}}M^{\frac{1-\alpha}{2}}|d\vzeta\\
&= \frac{1}{(\alpha-1)} \log \int \left(q_{\vzero}(J\vzeta)\right)^{1-\alpha}\left(p_{\vzero}(\vzeta)\right)^{\alpha} |J|^{1-\alpha}d\vzeta  \tag{$J:=S^{-\frac{1}{2}}M^{\frac{1}{2}}$}\\
&= \frac{1}{(\alpha-1)} \log \left( |J|^{1-\alpha}\int \left(q_{\vzero}(J\vzeta)\right)^{1-\alpha}\left(p_{\vzero}(\vzeta)\right)^{\alpha}d\vzeta\right)\\
&= \frac{1}{(\alpha-1)} \log \left( |J|^{1-\alpha}\right) + \frac{1}{(\alpha-1)}\log \int \left(q_{\vzero}(J\vzeta)\right)^{1-\alpha}\left(p_{\vzero}(\vzeta)\right)^{\alpha}d\vzeta \label{eq:alpha_before_sphere}
\end{align}

Since $p_\vzero$ and $q_\vzero$ are spherically symmetric, we can define them via functions depending on the norm of their inputs, i.e., 
\begin{align*}
g(||J\vzeta||) &= (q_{\vzero}(J\vzeta))^{1-\alpha}\\
f(||\vzeta||) &= (p_{\vzero}(\vzeta))^{\alpha}.
\end{align*}
Plugging this into \cref{eq:alpha_before_sphere}, and taking the derivative w.r.t $J$
\citep{petersen2008matrix}, we obtain
\begin{align}
\nabla_{J}D_{\alpha}(p, q) 
&= \frac{1}{(\alpha-1)}\nabla_J \log \left( |J|^{1-\alpha}\right) + \frac{1}{(\alpha-1)}\nabla_J\log \int \left(q_{\vzero}(J\vzeta)\right)^{1-\alpha}\left(p_{\vzero}(\vzeta)\right)^{\alpha}d\vzeta\\
&=\frac{1-\alpha}{\alpha-1}J^{-T} + \frac{1}{(\alpha-1)}\frac{\int g'(||J\vzeta||)\frac{J\vzeta\vzeta^T}{||J\vzeta||} f\left(||\vzeta||\right)d\vzeta}{\int g(||J\vzeta||)f(||\vzeta||)d\vzeta}
\end{align}
Now setting $J = \gamma I$ for some $\gamma > 0$ and using that the gradient should be $0$ for this assignment of $J$, we obtain
\begin{align}
-\frac{1-\alpha}{\alpha-1}\gamma^{-1}I &= \frac{1}{(\alpha-1)}\frac{\int g'(||\gamma\vzeta||)\frac{\vzeta\vzeta^T}{||\vzeta||} f\left(||\vzeta||\right)d\vzeta}{\int g(||\gamma\vzeta||)f(||\vzeta||)d\vzeta}.\label{eq:gamma_alpha}
\end{align}

We again need to show that the r.h.s. exhibits proper diagonal structure. 
First of all, note that the numerator on the r.h.s. coincides with what we obtained for the FKL (\cref{app:fkl_corr}). 
Hence, it gives the desired isotropic diagonal structure. 
Moreover, the denominator $\int g(\gamma||\vzeta||)f(||\vzeta||)d\vzeta$ shares the same spherical property, and hence still ensures that all diagonal entries will obtain the same value. Hence, we maintain the necessary structure.

It remains to show that there exists a solution with $\gamma > 0$ for equation \cref{eq:last_eq_gamma}. Taking again the traces on both sides, we obtain
\begin{align}
-(1-\alpha)d\gamma^{-1} &= \frac{\int g'(\gamma||\vzeta||)||\vzeta|| f\left(||\vzeta||\right)d\vzeta}{\int g(\gamma||\vzeta||)f(||\vzeta||)d\vzeta}.\label{eq:alpha_correlations_final_boss}
\end{align}

First of all, note that under the assumption that $\alpha > 1$, the l.h.s. is positive and strictly decreasing in $\gamma >0$. 
Hence, we need to ensure that the r.h.s. is increasing in $\gamma$ and positive to guarantee a solution.

Note that the denominator on the r.h.s. is non-negative by the definitions of $g$ and $f$. Moreover, $g$ corresponds to $q_{\vzero}^{1-\alpha}$ which for $\alpha > 1$ is convex under our assumptions. Hence, $g'$ is increasing, and we know $g'(0) \geq 0$ (or at least \limca when the derivative at the origin is ill-defined), hence we have $g'(\gamma||\vzeta||) \geq 0$. Note that $g'$ cannot be $0$ everywhere since \cref{setting_1} requires $\wa$ strictly convex and \cref{setting_2} requires it to be strictly increasing in $||\vx||$. For $\alpha > 1$, also the coefficient $\frac{1}{\alpha-1}$ is positive. We hence conclude that the r.h.s. of \cref{eq:alpha_correlations_final_boss} is positive.

To see why the r.h.s. is increasing in $\gamma$, remember that we have
\begin{align}
\frac{\int g'(\gamma||\vzeta||)||\vzeta|| f\left(||\vzeta||\right)d\vzeta}{\int g(\gamma||\vzeta||)f(||\vzeta||)d\vzeta} =
 \frac{\partial}{\partial \gamma } \left[\log \int (q_{\vzero}(J \vzeta))^{1-\alpha}(p(\vzeta)^{\alpha})d\vzeta\right]_{J=\gamma I}
\label{eq:log_convex_integrand_gamma}
\end{align}
We have established in \textbf{Part 1} that the integral on the r.h.s. is 
log-convex in $S^{-\frac{1}{2}}$, which implies it is log-convex in $J=S^{-\frac{1}{2}}M^{\frac{1}{2}}$. As a result, the derivative of the r.h.s. is increasing in $\gamma \in (0, \infty)$. Therefore, the r.h.s. of \cref{eq:alpha_correlations_final_boss} is increasing in $\gamma$.

Since the r.h.s. of \cref{eq:alpha_correlations_final_boss} is positive and increasing in $\gamma$, we conclude that there exists a solution for $\gamma$, and hence $D_{\alpha}^R$ has a stationary point for $S$, such that $S^{\frac{1}{2}}= \frac{1}{\gamma}M^{\frac{1}{2}}$. We note that this solution corresponds to the unique minimizer of $D_{\alpha}(p, q_{\vnu, S})$ w.r.t. $S$, and hence our sufficient conditions allow for exact recovery of correlations with the $\alpha$-DIV.

\section{Experimental setup}\label{app:simulations_setup}
\begin{table}
\caption{A summary of characteristics of the targets used in \cref{sec:simulations}. The last column highlights the corresponding applicable theorems. Note that for the condition $\vmu \in \text{supp}(p)$, we mean \csupp, as stated in our theorems.} \label{tab:target_stats}
\resizebox{\textwidth}{!}{
\begin{tabular}{llcccccc}\toprule
& $d$ & Even symm. & Elliptically symm. & $\vmu \in \text{supp}(p)$& Log-concave & Theorems\\ \midrule
\emph{Symbolic Integration} \\ \midrule
$p_1$ (MoU) & $1$  & \yes & \yes & \no & \no & \cref{theorem:theorem_fkl_1,theorem:theorem_alpha_1}\\
$p_2$ (MoU) & $1$ &  \yes & \yes & \yes & \no &\cref{theorem:theorem_fkl_1,theorem:theorem_fkl_2,theorem:theorem_alpha_1,theorem:theorem_alpha_2}\\ \midrule 
\emph{Stochastic Optimization} \\  \midrule
MoU & $2$ & \yes & \no & \no & \no &\cref{theorem:theorem_fkl_1,theorem:theorem_alpha_1}\\
Ellipse & $2$ & \yes & \yes & \yes & \no &\cref{theorem:theorem_fkl_1,theorem:theorem_fkl_2,theorem:theorem_alpha_1,theorem:theorem_alpha_2,theorem:theorem_fkl_corr,theorem:theorem_alpha_corr}\\
GMM & $2$ & \yes & \no & \yes & \no &\cref{theorem:theorem_fkl_1,theorem:theorem_fkl_2,theorem:theorem_alpha_1,theorem:theorem_alpha_2}\\
A-GMM & $2$ & \no & \no & \yes & \no & /\\
GMM ($4D$) & $4$ & \yes & \no & \yes & \no &\cref{theorem:theorem_fkl_1,theorem:theorem_fkl_2,theorem:theorem_alpha_1,theorem:theorem_alpha_2}\\
GMM ($8D$) & $8$ & \yes & \no & \yes & \no &\cref{theorem:theorem_fkl_1,theorem:theorem_fkl_2,theorem:theorem_alpha_1,theorem:theorem_alpha_2}\\
GMM ($16D$) & $16$ & \yes & \no & \yes & \no &\cref{theorem:theorem_fkl_1,theorem:theorem_fkl_2,theorem:theorem_alpha_1,theorem:theorem_alpha_2}\\
Ellipse ($4D$) & $4$ & \yes & \yes & \yes & \no &\cref{theorem:theorem_fkl_1,theorem:theorem_fkl_2,theorem:theorem_alpha_1,theorem:theorem_alpha_2,theorem:theorem_fkl_corr,theorem:theorem_alpha_corr}\\
Ellipse ($8D$) & $8$ & \yes & \yes & \yes & \no &\cref{theorem:theorem_fkl_1,theorem:theorem_fkl_2,theorem:theorem_alpha_1,theorem:theorem_alpha_2,theorem:theorem_fkl_corr,theorem:theorem_alpha_corr}\\
Ellipse ($16D$) & $16$ & \yes & \yes & \yes & \no &\cref{theorem:theorem_fkl_1,theorem:theorem_fkl_2,theorem:theorem_alpha_1,theorem:theorem_alpha_2,theorem:theorem_fkl_corr,theorem:theorem_alpha_corr}\\
\bottomrule
\end{tabular}
}
\end{table}
\subsection{Targets}\label{app:targets}
\subsubsection{Two-dimensional targets}\label{app:targets_2D}
\paragraph{MoU.} We define an even symmetric two-dimensional mixture of uniforms with \emph{disjoint support}, akin to the targets used in our symbolic experiments. 
\begin{align*}
    p^{\text{MoU}} = 0.5 \cdot \text{Unif}(x_1;[5.75,9.75])\cdot \text{Unif}(x_2; [-1.72, 2.28]) + 0.5 \cdot \text{Unif}(x_1;[-5.75,-1.75])\cdot \text{Unif}(x_2; [1.72, 5.72]),
\end{align*}

For the full mixture, we have $\vmu=(2.0, 2.0)^T$, $\text{Corr}_p=\begin{pmatrix}
1 & -0.81 \\
-0.81 & 1
\end{pmatrix}$\\ and $\text{Cov}_p=\begin{pmatrix}
34.36 & -9.9 \\
-9.9 & 4.3058
\end{pmatrix}$. 

The target is not log-concave and not elliptically symmetric. We sample this target via standard ancestral sampling \citep{bishop2006pattern}. Note that for this target, the RKL is not well-defined. Moreover, for the $\alpha$-divergence, we use the REINFORCE gradient estimator instead of reparameterization, since  $p'$ is not informative for this target.

\paragraph{GMM.} We define a GMM target that is \emph{even symmetric} around its mean.
\begin{align*}
    p^{\text{GMM}}(\vx) = 0.5 \cdot \mathcal{N}\left(\vx;(-3, -3)^T, \Sigma\right) + 0.5 \cdot \mathcal{N}\left(\vx; (7, 7)^T, \Sigma\right),
\end{align*}
with
$\Sigma = \begin{pmatrix}
1 & -0.57 \\
-0.57 & 2
\end{pmatrix}$.

For the full mixture, we have $\vmu=(2, 2)^T$, $\text{Corr}_p=\begin{pmatrix}
1 & 0.92 \\
0.92 & 1
\end{pmatrix}$ \\and $\text{Cov}_p=\begin{pmatrix}
26 & 24.43 \\
24.43 & 27
\end{pmatrix}$. The target is not log-concave and not elliptically symmetric. We sample this target via standard ancestral sampling \citep{bishop2006pattern}.

\paragraph{Ellipse.} 
We use a \emph{squared subtractive mixture model (SMM)} \citep{loconte2024subtractive} to represent an elliptic target with a valley in its middle. For a two component GMM, where each component has mean $0$, \emph{squaring} an SMM results in 
\begin{align}
p^{\text{Ellipse}}(\vx) &= \frac{1}{Z} \left(w_1\mathcal{N}(\vx; (0,0)^T, \Sigma_{1}) - w_2\mathcal{N}(\vx, (0,0)^T, \Sigma_{2})\right)^2,
\label{eq:squared}
\end{align} 
where $Z$ is the tractable normalizing constant of the mixture.

The base distribution of our \emph{Ellipse} target is given as \cref{eq:squared}, parameterized with
where $\Sigma_{1} = \begin{pmatrix}
1 & 0 \\
0 & 1
\end{pmatrix}$ and $\Sigma_{2} = \begin{pmatrix}
0.3 & 0 \\
0 & 0.3
\end{pmatrix}$, $w_{1}=0.75$, $w_{2}=0.25$. The resulting density is visualized as the \emph{spherically symmetric} density in \cref{fig:symmetry_illustration}. We transform this base target into an ellipse via \cref{eq:p_ell} with $M=\begin{pmatrix}
4 & 2 \\
2 & 4
\end{pmatrix}$ and $\vmu=(1, 1)^T$.

For the full mixture, we have $\vmu=(1, 1)^T$, $\text{Corr}_p=\begin{pmatrix}
1 & 0.5 \\
0.5 & 1
\end{pmatrix}$ and $\text{Cov}_p=\begin{pmatrix}
3.70 & 1.85 \\
1.85 & 3.70
\end{pmatrix}$.

The resulting target is \emph{elliptically symmetric} (and hence even symmetric) but not log-concave. To sample $p^{\text{Ellipse}}$, we use rejection sampling based on the positively weighted components of the base distribution \citep{bignami1971note}, and transform the resulting samples:
\begin{align*}
\vx &\sim p_{\vzero}^{\text{Ellipse}} \text{ (via rejection)}\\
\vx &= M^{\frac{1}{2}}\vx + \vmu.
\end{align*}
We use the cirkit library \citep{The_APRIL_Lab_cirkit_2024} (released under the GPL-3.0 license) for implementing this target, as well as the \emph{Ellipse} targets in higher dimensions.

\paragraph{A-GMM.} We also construct an \emph{asymmetric GMM}, which is neither elliptically nor even symmetric.
\begin{align*}
    p^{\text{AGMM}}(\vx) = 0.5 \cdot \mathcal{N}\left(\vx;(5, 4)^T, \Sigma_1\right) + 0.5 \cdot \mathcal{N}\left(\vx;(-2, -6)^T, \Sigma_2\right),
\end{align*}
with
$\Sigma_1 = \begin{pmatrix}
1 & -0.57 \\
-0.57 & 2
\end{pmatrix}$ and $\Sigma_2 = \begin{pmatrix}
0.5 & 0.14\\
0.14 & 1
\end{pmatrix}$

For the full mixture, we have $\vmu=(1.5, -1)^T$, $\text{Corr}_p=\begin{pmatrix}
1 & 0.93 \\
0.93 & 1
\end{pmatrix}$ and $\text{Cov}_p=\begin{pmatrix}
13.0 & 17.29 \\
17.29 & 26.5
\end{pmatrix}$. The target is not log-concave and not elliptically symmetric. We sample this target via standard ancestral sampling \citep{bishop2006pattern}.

\subsubsection{Higher-dimensional 
targets}\label{app:targets_high_d}
\paragraph{GMM (d-dimensional).} We define a $d$-dimensional \emph{even symmetric} GMM target. Let $\mathbf{1}^d$ denote a vector with all entries equal to $1$ in $d$ dimensions.

\begin{align*}
    p^{\text{GMM}}(\vx) &= 0.5 \cdot \mathcal{N}\left(\vx;0\cdot\mathbf{1}^d, \Sigma\right) + 0.5 \cdot \mathcal{N}\left(\vx;2\cdot\mathbf{1}^d, \Sigma\right),
\end{align*}
with an isotropic $\Sigma$ with marginal variance $1$. The overall target has a mean of $\vmu= \mathbf{1}^d$, correlation of $0.5$ on all off-diagonals and the covariance matrix has marginal variance $2$ and covariance $1$. We sample this target via standard ancestral sampling \citep{bishop2006pattern}.

\paragraph{Ellipse (d-dimensional).} We create a higher-dimensional version of the two-dimensional squared SMM. The general setup is the same as for $d=2$, with the main exception that we vary the mixture weight of the negatively weighted component \emph{before squaring} (see \citep{loconte2024subtractive}) with $d$, as $w_1= 1 - 1 /d^2$ and $w_2=1 / d^2$. The mean of the overall mixture is always a $d$-dimensional vector filled with $1$ and the correlation between all pairs of distinct variables is $0.5$. The covariance vary slightly for each $d$, but the marginal variance is around $2$ for all $d$, while off-diagonal entries of the covariance matrix are roughly $1$. We again use rejection sampling to sample this target.

\subsection{Optimization details}\label{app:optimization_details}
Below, we describe our setup for both symbolic integration and stochastic optimization. 

\paragraph{Hardware.} We run experiments with symbolic integration on CPU (11th Gen Intel(R) Core(TM) i5-1145G7 @ 2.60GHz) and experiments with stochastic optimization on a cluster with $8$ NVIDIA RTX A6000 (48GiB VRAM) GPUs and $2$ AMD EPYC 7452 32-Core Processors.

\paragraph{Symbolic integration.} For the experiments shown in \cref{fig:exps_mean}, we use symbolic integration with SymPy \citep{sympy} to compute the divergence. We evaluate $\nu$ on a grid with step size $0.01$, in the range $\vnu \in [-15, 15]$.

\paragraph{Stochastic optimization.} For experiments involving stochastic optimization, we use the Adam optimizer \citep{kingma2014adam} with $lr = \lr$. We learn on two-dimensional targets for a maximum of $10^4$ steps with $10^5$ samples per update. On higher-dimensional targets, we train for $4\cdot10^4$ steps and $10^4$ samples for $d\in\{4, 8\}$ and $2\cdot10^5$ samples for $d=16$. For the $4$-dimensional Ellipse, we instead use $10^4$ samples per update, training for $2\cdot10^4$ update steps with a learning rate of $0.01$.
Our implementation is based on Pytorch \citep{ansel2024pytorch} (version 2.9.1). Below, we provide implementation details for our variational families. \cref{tab:inits} describes the initializations used for each target density.

\textbf{Gaussian.} We base our implementation on torch.distributions \citep{ansel2024pytorch}.

\paragraph{Laplace (2D).} For the two-dimensional Laplace distribution \citep{eltoft2006multivariate}, we use its symmetric formulation:
\begin{align*}
q_{\vnu, S}^{\text{Laplace}}(\vx) &= |S^{-\frac{1}{2}}|q_{\vzero}^{\text{Laplace}}\left(S^{-\frac{1}{2}}(\vx-\vnu)\right), \text{with}\\
q_{\vzero}^{\text{Laplace}}(\vz) &= \frac{1}{\pi}K_0\left(\sqrt{2\vz^T\vz}\right)
\end{align*}
where $K_0$ is the modified Bessel function of the second kind \citep[Ch. 9]{abramowitz1948handbook},
\begin{align*}
K_0(z) &= \int_{0}^{\infty} \cos(z~\text{sinh}(t))dt.
\end{align*}
To sample from the two-dimensional Laplace distribution, we exploit its connection to the exponential distribution, as well as the fact that it is a Gaussian scale mixture. We generate samples from $q_{\vnu, S}^{\text{Laplace}}$ as detailed below \citep{kotz2001asymmetric}, where $\text{Exp}(1)$ denotes a one-dimensional exponential distribution with $\lambda=1$:
\begin{align*}
z &\sim \text{Exp}(1)\\
\boldsymbol{w} &\sim \mathcal{N}(0, I^{2 \times 2})\\
\vx &= \vnu + \sqrt{z}S^{\frac{1}{2}}\boldsymbol{w},
\end{align*}
where now $\vx \sim \text{Laplace}(\vnu, S)$. Since the multivariate Laplace is not readily available in any major deep learning library, we manually implement the two-dimensional Laplace using built-in functions from PyTorch \citep{ansel2024pytorch}. Note that, while the one-dimensional Laplace distribution is log-concave, its higher-dimensional generalization as described above is not.

\paragraph{Student-t.} We use the implementation provided in Pyro \citep{bingham2019pyro}. In our experiments, we use $v=5$ for two-dimensional targets and for the higher-dimensional \emph{GMM} targets. For the higher-dimensional \emph{Ellipse} targets, we use $v=8$, as we found this configuration to be more stable during optimization.

\paragraph{Parameterization.} We parameterize the scale of the variational approximation via an unconstrained lower-triangular matrix $L$. To ensure that the resulting covariance matrix is positive definite, we define a matrix $L'$ as follows: 
$$L'_{ij}=
\begin{cases}
L_{ij} \quad \text{if }  i \neq j\\
\exp(L_{ij}) \quad \text{if } i = j.\\
\end{cases}
$$
The covariance of $q$ is then defined as $Cov_q=L'L'^T$. See \cref{tab:inits} for the initializations used for each target.

\paragraph{Initialization.} \cref{tab:inits} summarizes our initialization scheme for each target. We found the reported initializations to be stable for all tested divergences. For repeated runs we \emph{noise} the values reported in \cref{tab:inits} as described below, where $\boldsymbol{\epsilon}_{\vnu}$ is drawn from a $d$-dimensional standard Gaussian and $\boldsymbol{\epsilon}_{L}$ denotes a $d\times d$ lower-triangular matrix, whose non-zero entries are similarly drawn from a standard Gaussian. For $d < 16$, we then set
\begin{align*}
\vnu_{\text{init}} &= \vnu_{\text{init}} + \boldsymbol{\epsilon}_{\vnu}\\
L_{\text{init}} &= L_{\text{init}} + 0.2\cdot\boldsymbol{\epsilon}_{L}.
\end{align*}

For $d=16$, we found the $\alpha$-divergence with $\alpha=2.0$ to be instable at the start of learning unless carefully initialized. Hence, we use softer noising for $d=16$, as described below.
\begin{align*}
\vnu_{\text{init}} &= \vnu_{\text{init}} + 0.1\cdot\boldsymbol{\epsilon}_{\vnu}\\
L_{\text{init}} &= L_{\text{init}} + 0.2\cdot\boldsymbol{\epsilon}_{L}.
\end{align*}

\begin{table*}
\caption{Initializations for experiments with stochastic optimization. We initialize means as $0$-vectors in $d$ dimensions. For the Ellipse in $16$ dimensions, we defined separate initializations for the Student-t and Gaussian to improve learning stability.}\label{tab:inits}
\begin{center}
\begin{tabular}{lllll}\toprule
Target & d & Location ($\vnu$) & Marginal Variance & Off-diagonal Covariance\\\midrule
MoU & 2 & 0 & 25 & 0\\
GMM & 2 & 0 & 25 & 0\\
Ellipse & 2 & 0 & 30 & 25\\
A-GMM & 2 & 0 & 25 & 0\\\midrule
GMM & 4 & 0 & 4 & 3\\
GMM & 8 & 0 & 4 & 3\\
GMM & 16 & 0 & 4 & 3\\\midrule
Ellipse & 4 & 0 & 30 & 25\\
Ellipse & 8 & 0 & 30 & 25\\
Ellipse (Gaussian $q$) & 16 & 0 & 3 & 1.8\\ 
Ellipse (Student-t $q$) & 16 & 0 & 3 & 1\\ 
\bottomrule
\end{tabular}
\end{center}
\end{table*}

\section{Additional results and analysis}\label{app:additional_results}
\subsection{Failure of Laplace variational family on $p_1$}\label{app:case_1_2_details}
Below, we formalize why the Laplace distribution in  \cref{fig:exps_mean} fails to distinguish between different assignments of the location parameter $\nu' \in [-3, 3]$ when approximating $p_1$. The target is given as $X \sim0.5 \cdot \text{Unif}(X; [-9, -3])+0.5\cdot\text{Unif}(X; [3, 9])$, and the variational family is  $\mathcal{Q} = \text{Laplace}(\nu, 4)$. We have $\supp(p) = [-9, -3] \cup [3, 9]$ by definition of $p$.
As shown in \cref{fig:exps_mean}, optimizing $\nu$ with the FKL results in a plateau of the divergence around the true mean $(\mu = 0)$.

By the definition of the Laplace distribution (with scale $4$), members of our variational family take the form
\begin{align}
q_{\nu}(x) = \frac{1}{2\cdot 4}e^{-\frac{|x-\nu|}{4}} = \frac{1}{4}q_{0}{\Big(}\frac{1}{4}(x-\nu){\Big)}, ~~\text{with}~~q_{0}(x) = \frac{1}{2}e^{-|x|}.
\end{align}
Hence, we have $w(x) := -\log(q_{0}(x)) = -\log \frac{1}{2}+|x|$. Now, consider the difference in $w$ at alternative function inputs $\frac{1}{4}(x-\nu')$ and $\frac{1}{4}x$. Note hat this corresponds to $\frac{1}{4}(x-\mu-\nu')$ and $\frac{1}{4}(x-\mu)$ since $\mu=0$.
\begin{align}
\Delta w_{\nu'}(x):=w{\Big(}\frac{1}{4}(x-\nu'){\Big)}-w{\Big(}\frac{1}{4}x{\Big)} &= \left(\frac{|x- \nu'|}{4} - \frac{|x|}{4}\right).\label{eq:deltaw}
\end{align}
We have $\Delta w_{\nu'}(x) \geq 0$ whenever $|x - \nu'| > |x|$. We now consider the case $\nu' > 0$. We have
\begin{align*}
\Delta w_{\nu'}(x)
=
\begin{cases}
\frac{\nu'}{4} ~\text{for } x \leq 0\\
\frac{-2x+\nu'}{4}  ~\text{for }  0 < x < \nu'\\
-\frac{\nu'}{4} ~\text{for } x \geq \nu'\\
\end{cases}
\end{align*}
As a result, in line with the proof of \cref{theorem:theorem_universal_convex}, we can define $\mathcal{H}_1:=(-\infty, \frac{\nu'}{2}]$ (indicating $\Delta w_{\nu'} \geq 0$) and $\mathcal{H}_2 := (\frac{\nu'}{2}, \infty)$ (where $\Delta w_{\nu'} < 0$). We further split $\mathcal{H}_1$ into $\mathcal{H}_3:=(-\infty, -\frac{\nu'}{2})$, and $\mathcal{H}_4 := [-\frac{\nu'}{2}, \frac{\nu'}{2}]$. 

Now pick any $\nu' \in (0, 3]$: For this range of $\nu'$, $p$ is not supported on $\mathcal{H}_4$. Hence, $\mathcal{H}_4$ does not contribute to the difference in FKL values. The support of $p$ further restricts the maximum ``effective'' regions of $\mathcal{H}_3$ and $\mathcal{H}_2$ (contributing to $\Delta\text{FKL}$) to $\mathcal{H}_3'=[-9, -3]$ and $\mathcal{H}_2'=[3, 9]$ respectively. On these intervals, for $\nu' \in (0, 3)$, we have $\Delta w_{\nu'}(x)= \frac{\nu'}{4}$ for $x \in \mathcal{H}'_3$ and $\Delta w_{\nu'}(x)= -\frac{\nu'}{4}$ for $x \in \mathcal{H}'_2$. Combined with the even symmetry of $p$, this confirms that we cannot distinguish any $\nu' \in (0, 3]$ from the true mean as the contributions of $\mathcal{H}_2$ and $\mathcal{H}_3$ to $\Delta\text{FKL}$ cancel.
Analogous reasoning can be applied for $\nu' \in [-3, 0)$.

\subsection{Additional results}\label{app:additional_results_sim}
In this section, we provide additional empirical results supporting our main theorems.

\subsubsection{Additional results for symbolic integration with the $\alpha$-DIV}\label{app:extra_alpha_case_3}
\begin{figure*}
\begin{center}
\begin{tabular}{cccc}
\multicolumn{4}{c}{$\alpha > 1$}\\\midrule
$\alpha=1.1$ & $\alpha=1.2$ & $\alpha=1.3$ & $\alpha=1.4$\\ 
\includegraphics[scale=0.2]{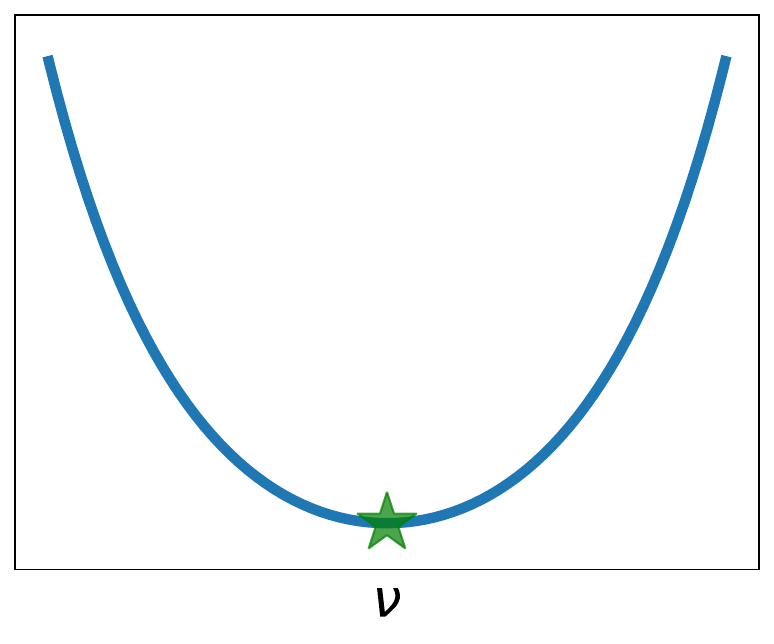}&
\includegraphics[scale=0.2]{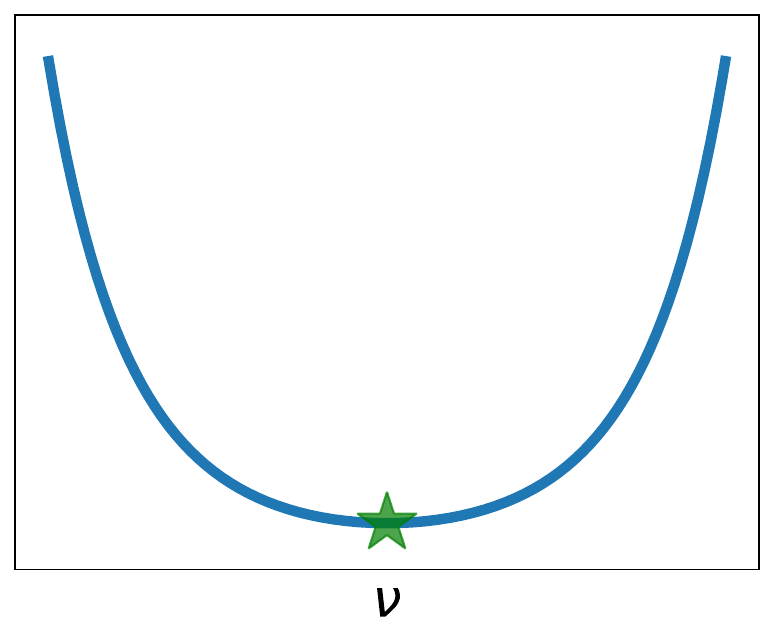}&
\includegraphics[scale=0.2]{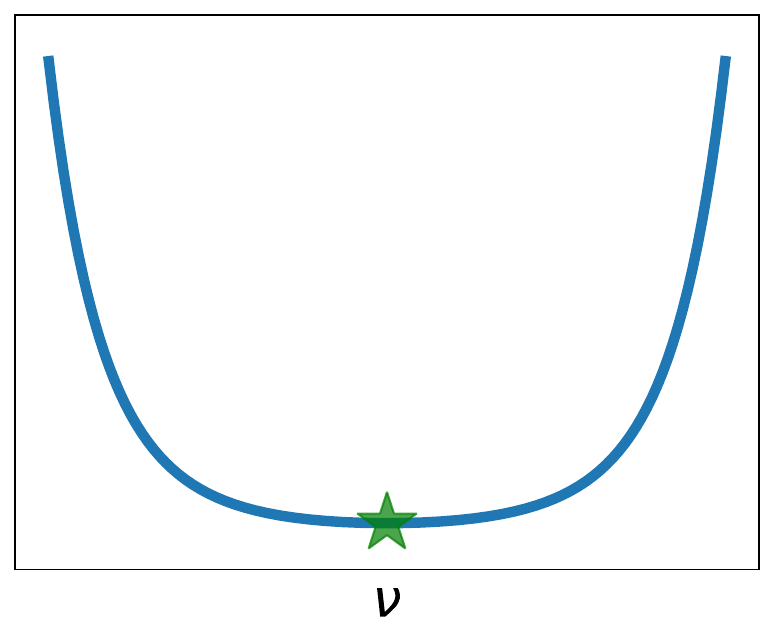}&
\includegraphics[scale=0.2]{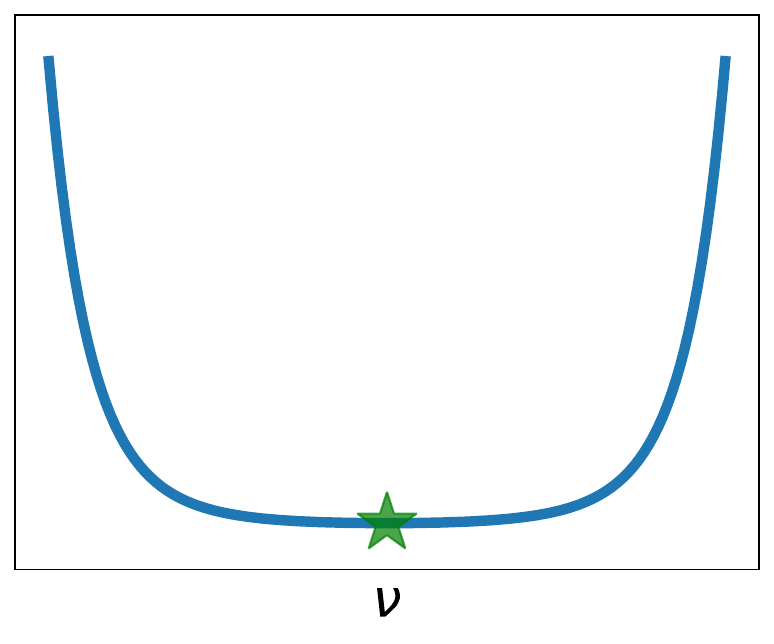}\\
\includegraphics[scale=0.2]{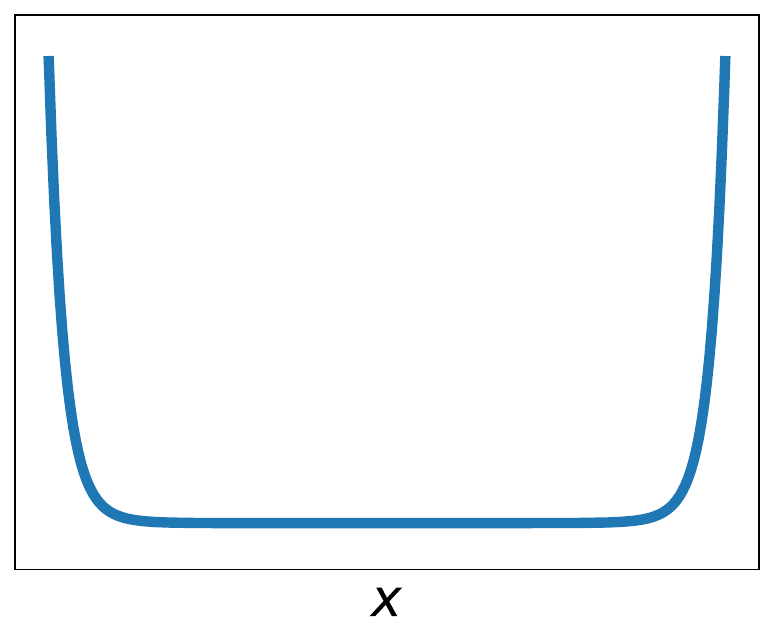}&
\includegraphics[scale=0.2]{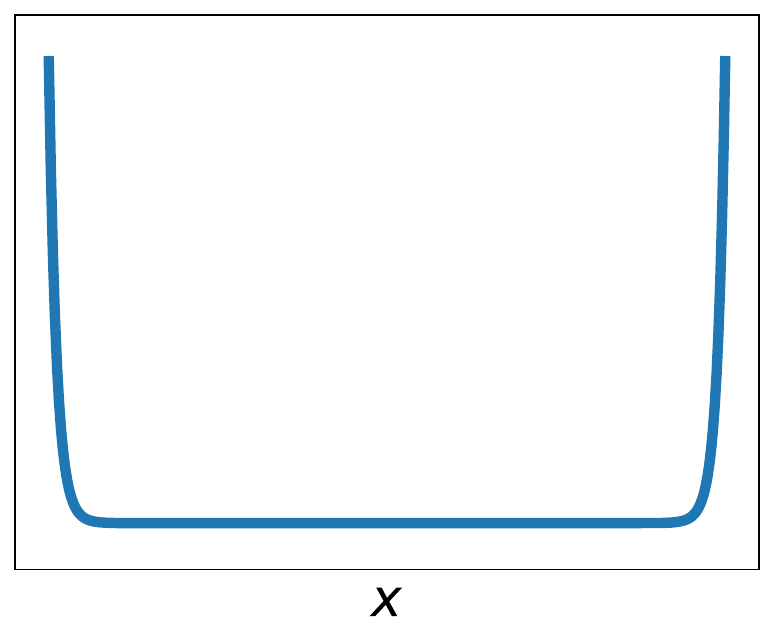}&
\includegraphics[scale=0.2]{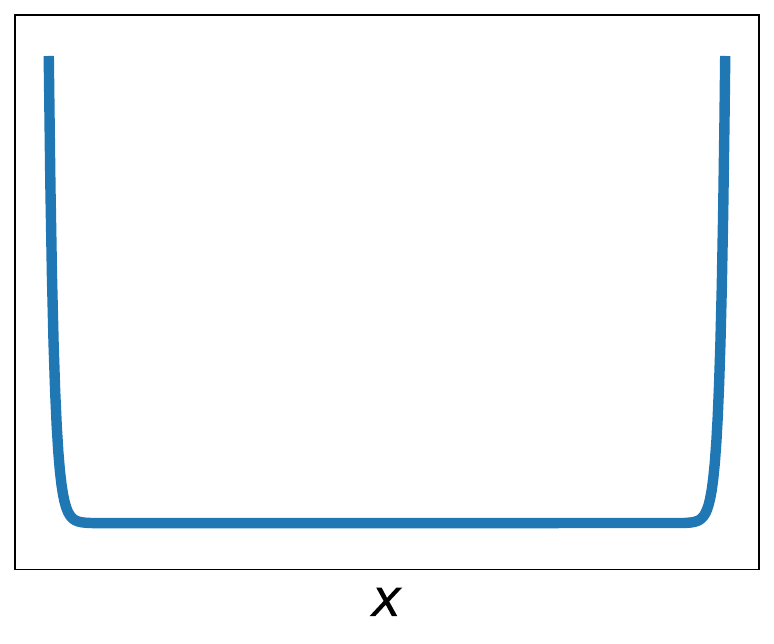}&
\includegraphics[scale=0.2]{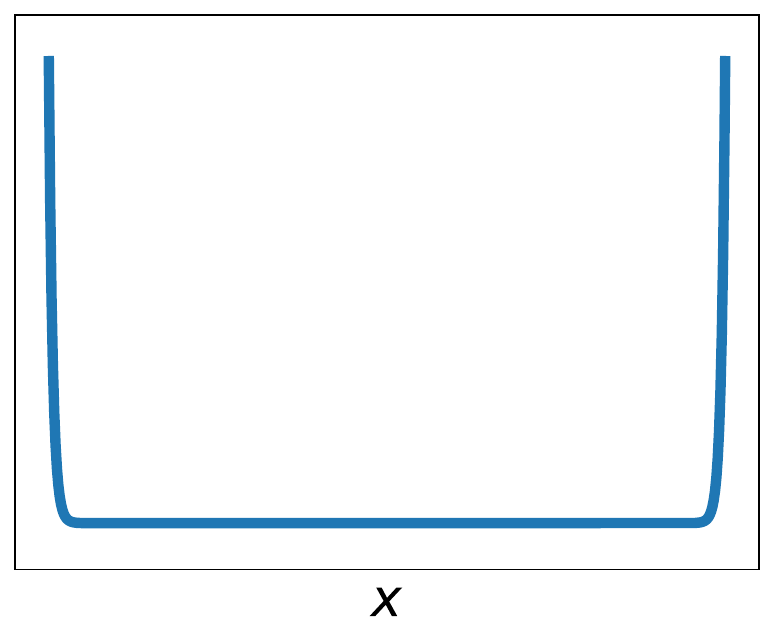}\\
$\alpha=1.5$ & $\alpha=1.6$ & $\alpha=1.7$ & $\alpha=1.8$\\
\includegraphics[scale=0.2]{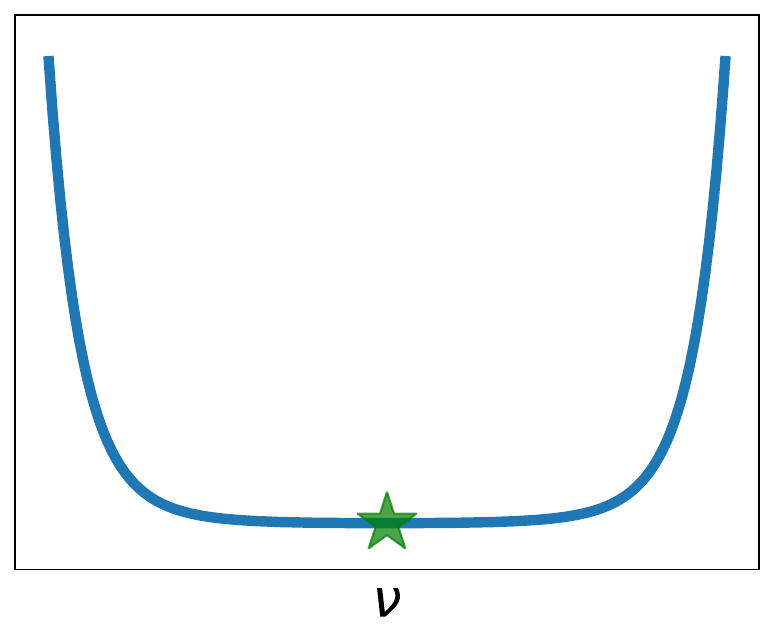}&
\includegraphics[scale=0.2]{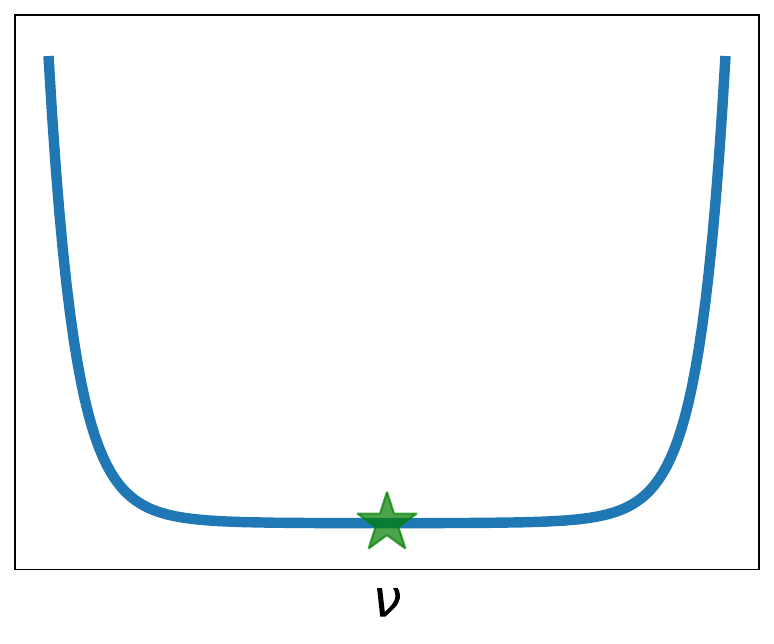}&
\includegraphics[scale=0.2]{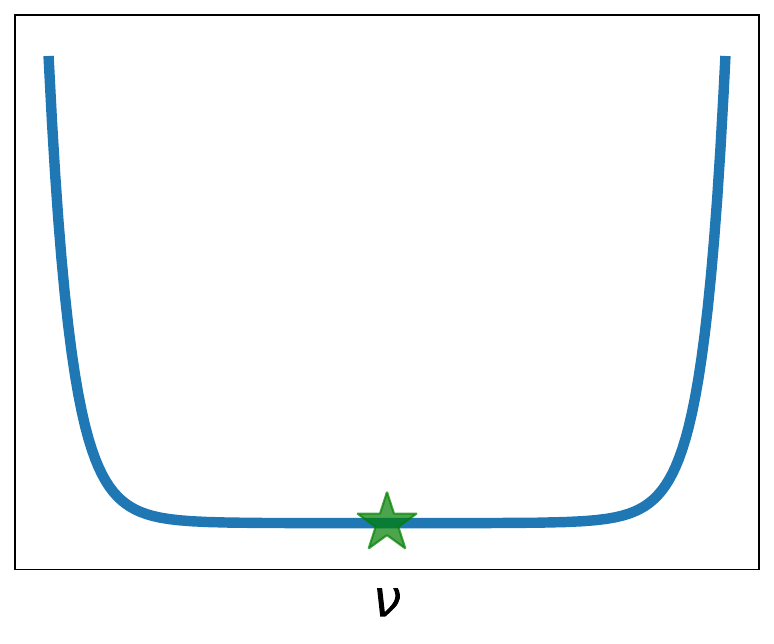}&
\includegraphics[scale=0.2]{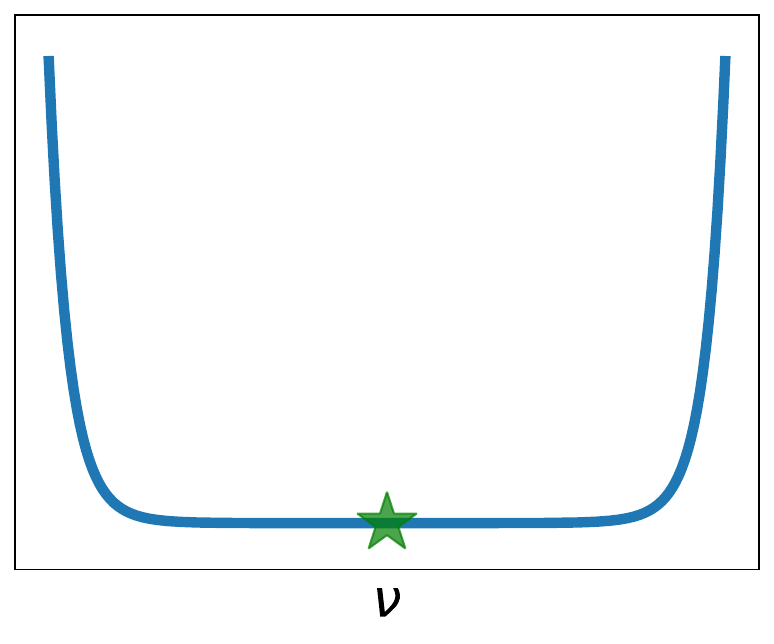}\\
\includegraphics[scale=0.2]{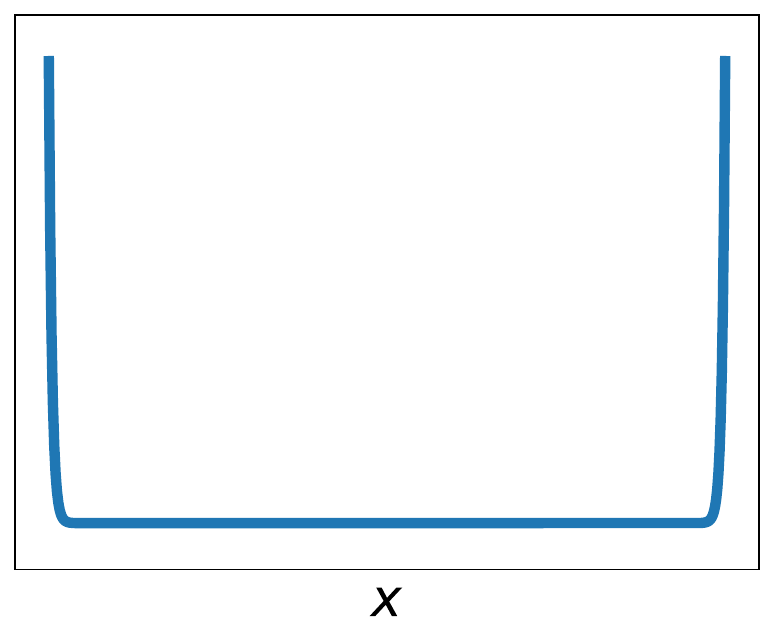}&
\includegraphics[scale=0.2]{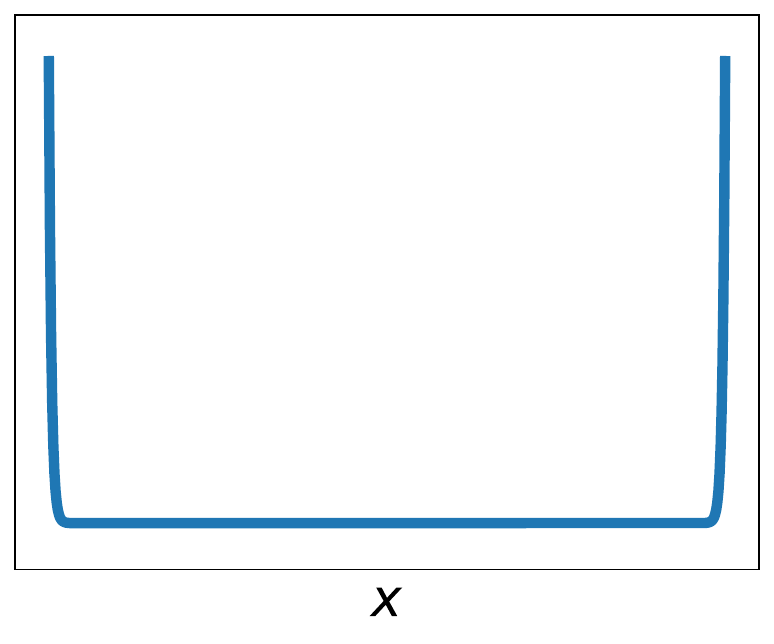}&
\includegraphics[scale=0.2]{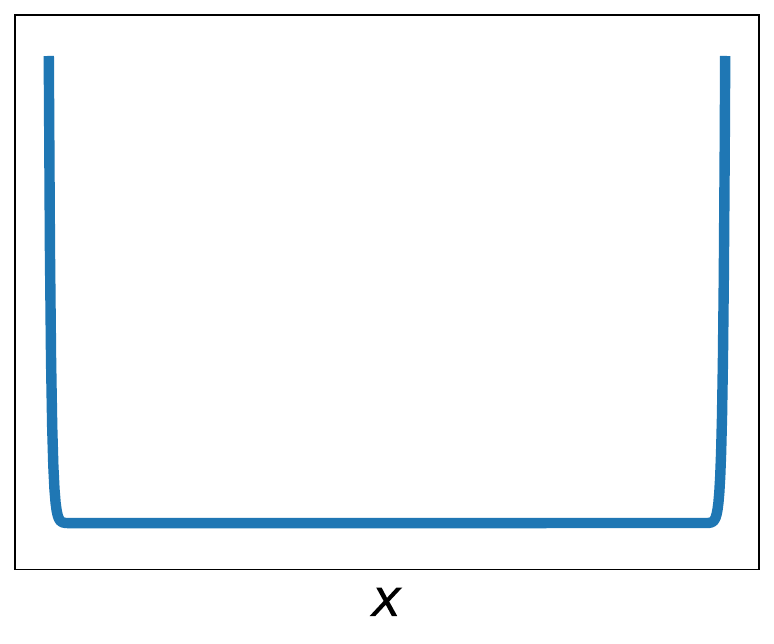}&
\includegraphics[scale=0.2]{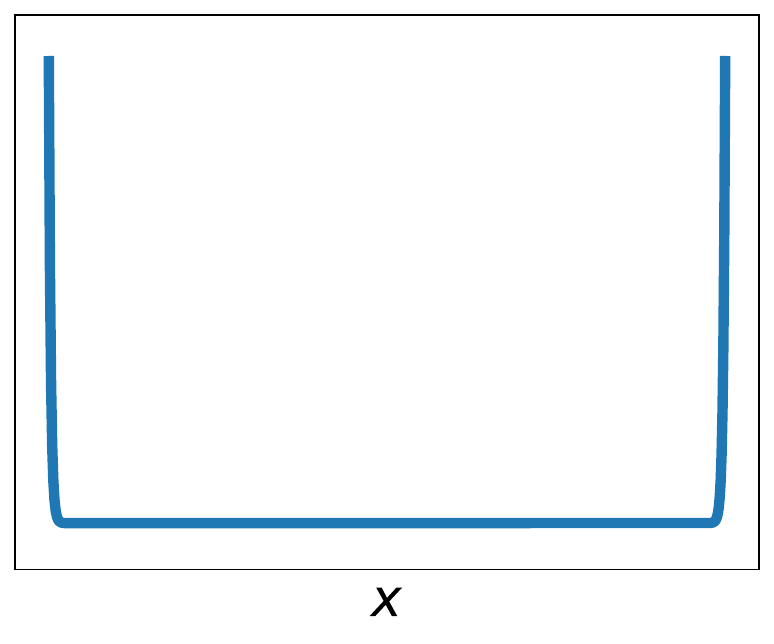}\\
\multicolumn{4}{c}{$\alpha<1$}\\\midrule
$\alpha=0.1$ & $\alpha=0.2$ & $\alpha=0.3$ & $\alpha=0.4$\\
\includegraphics[scale=0.2]{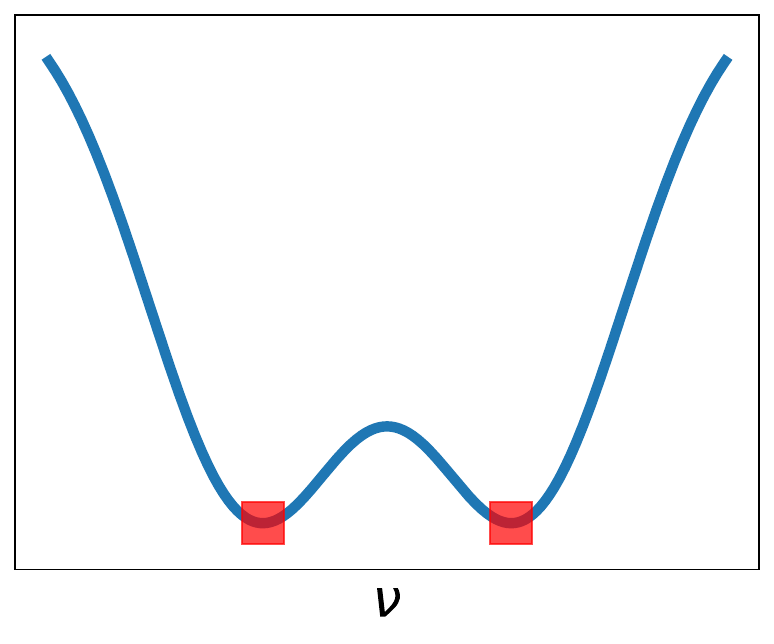}&
\includegraphics[scale=0.2]{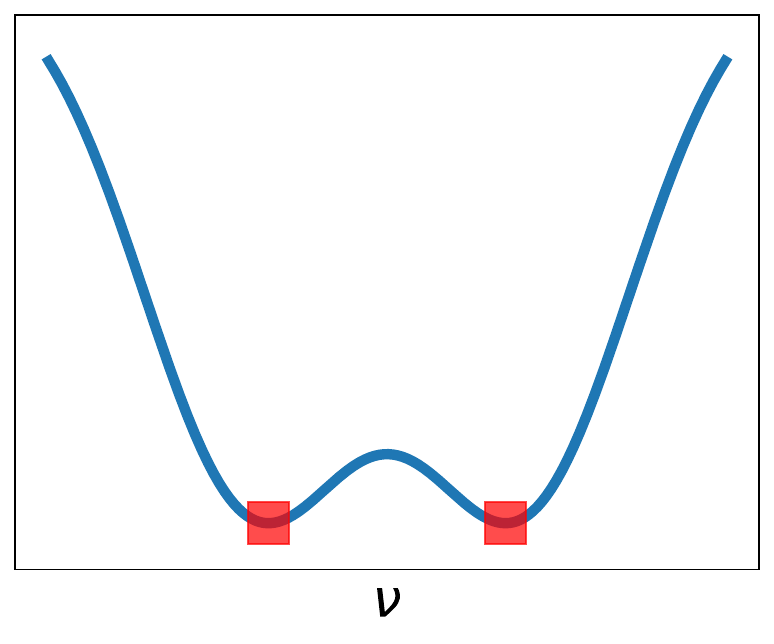}&
\includegraphics[scale=0.2]{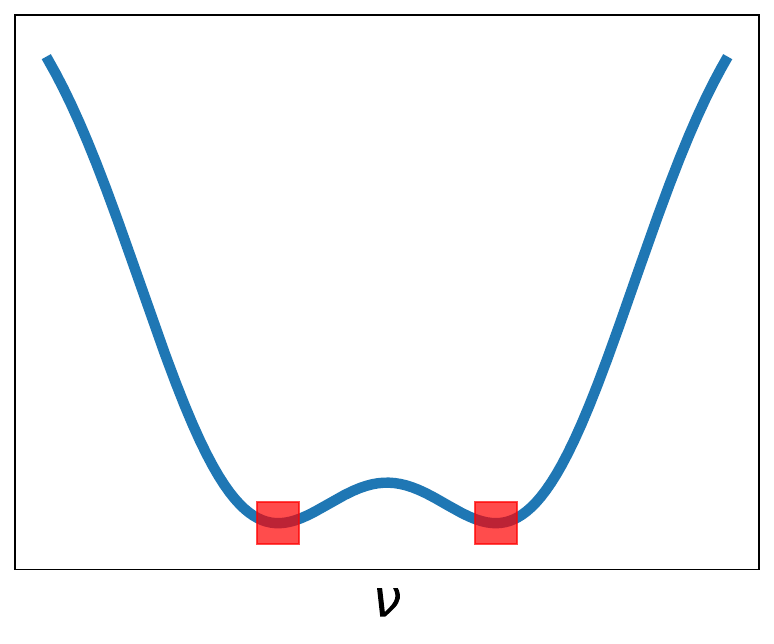}&
\includegraphics[scale=0.2]{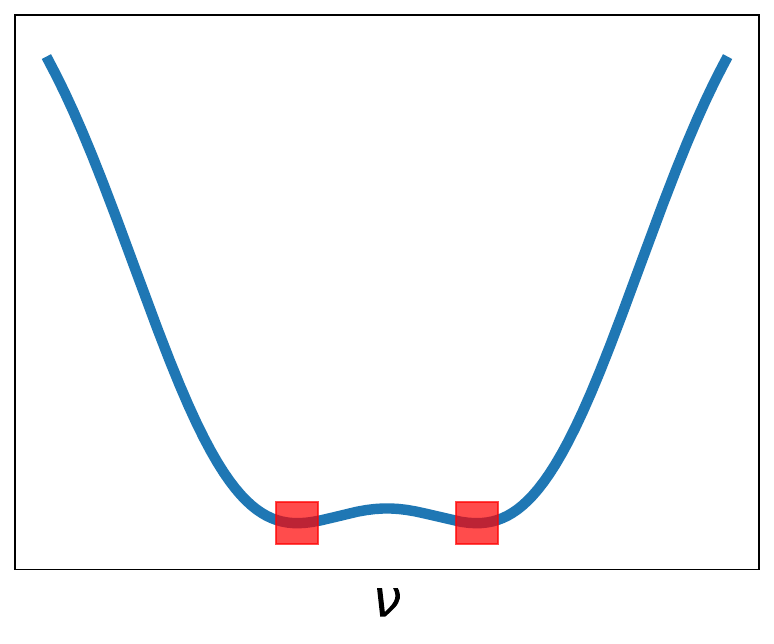}\\
\includegraphics[scale=0.2]{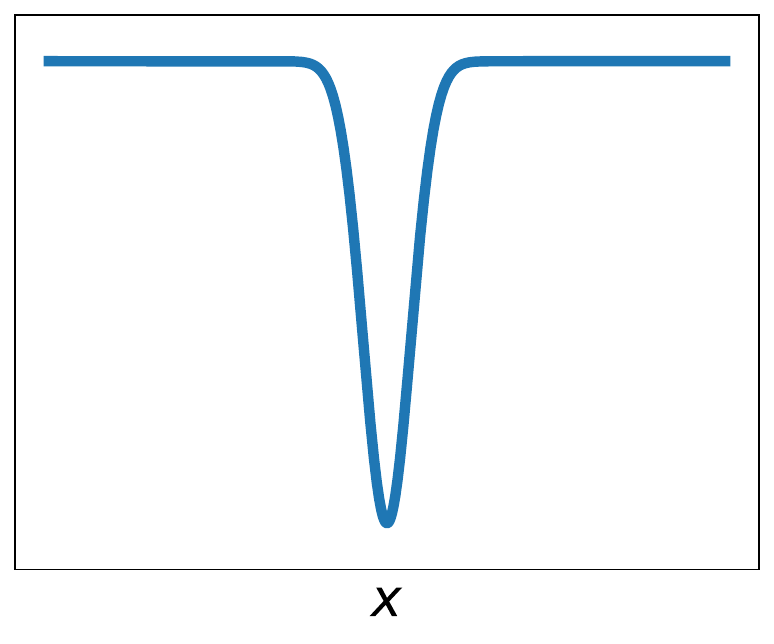}&
\includegraphics[scale=0.2]{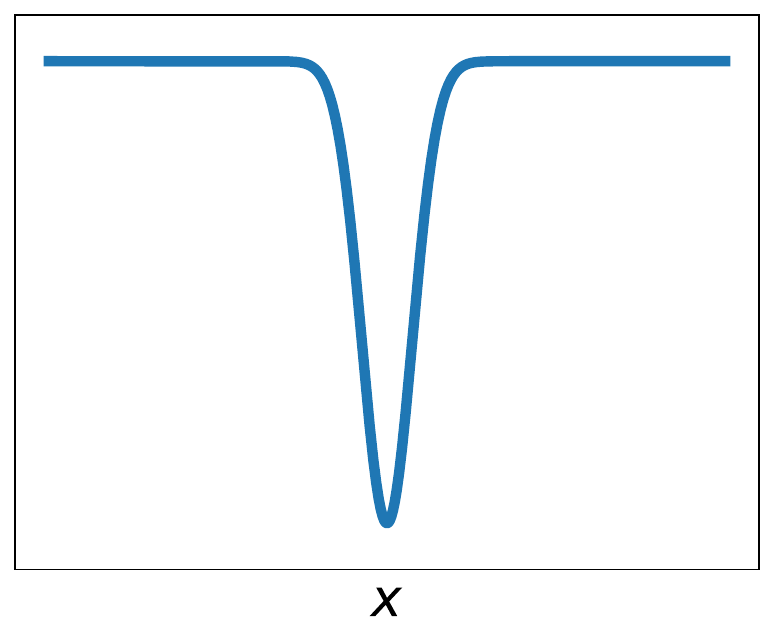}&
\includegraphics[scale=0.2]{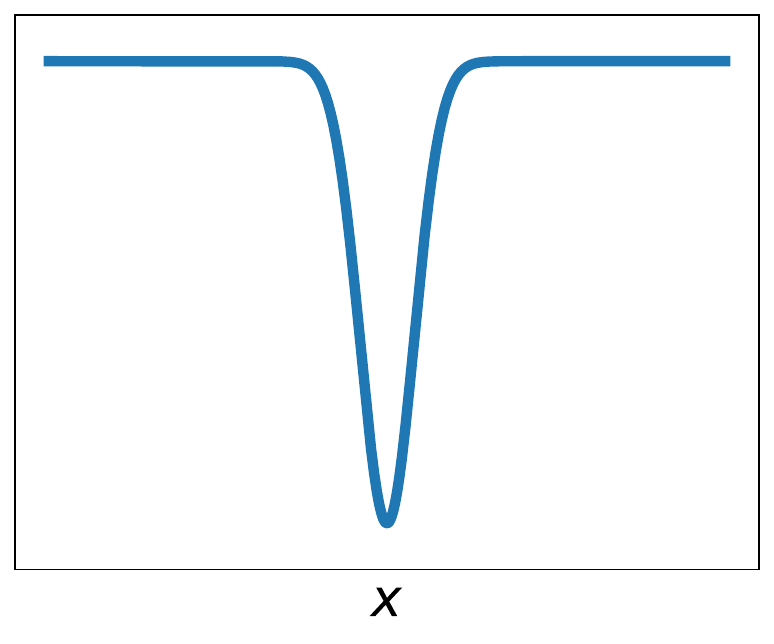}&
\includegraphics[scale=0.2]{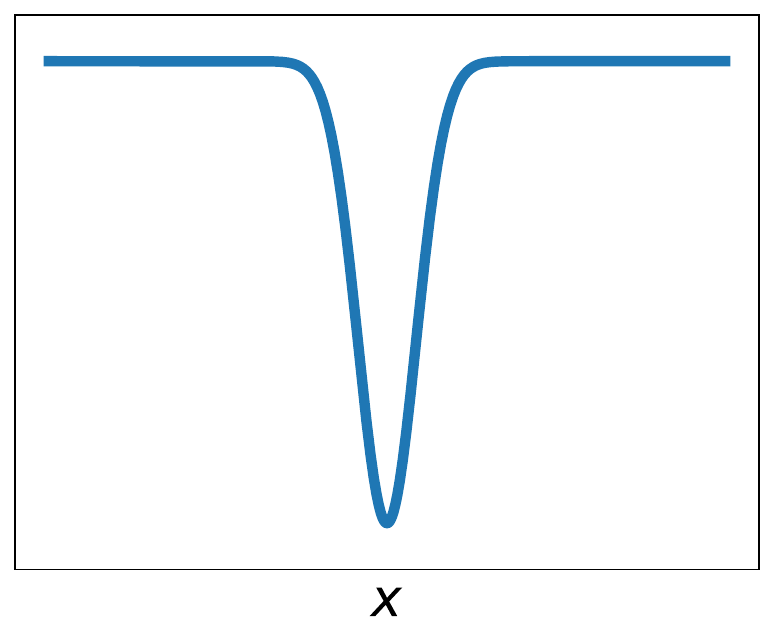}\\
$\alpha=0.5$ & $\alpha=0.6$ & $\alpha=0.7$ & $\alpha=0.8$\\
\includegraphics[scale=0.2]{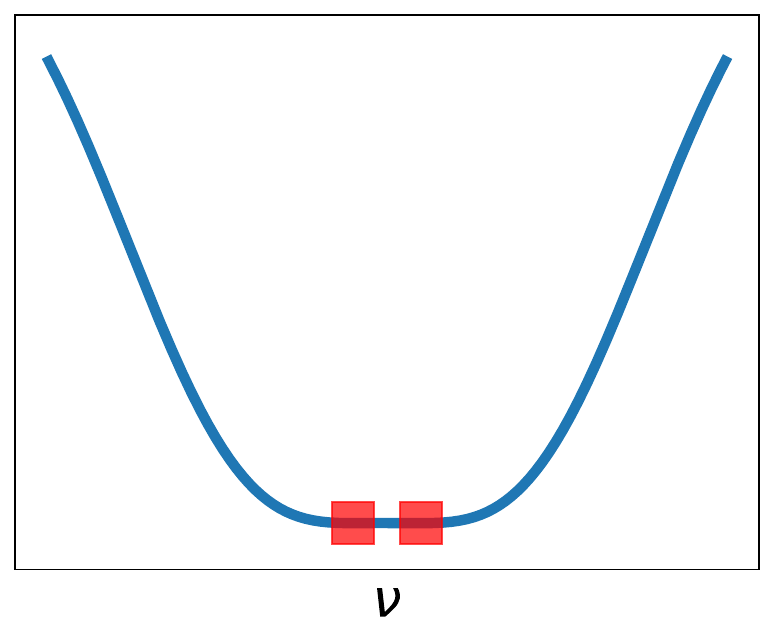}&
\includegraphics[scale=0.2]{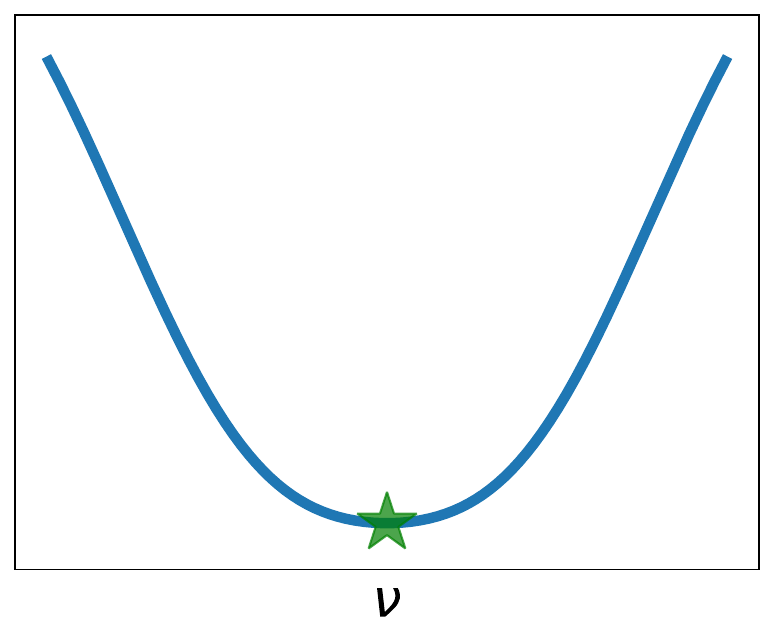}&
\includegraphics[scale=0.2]{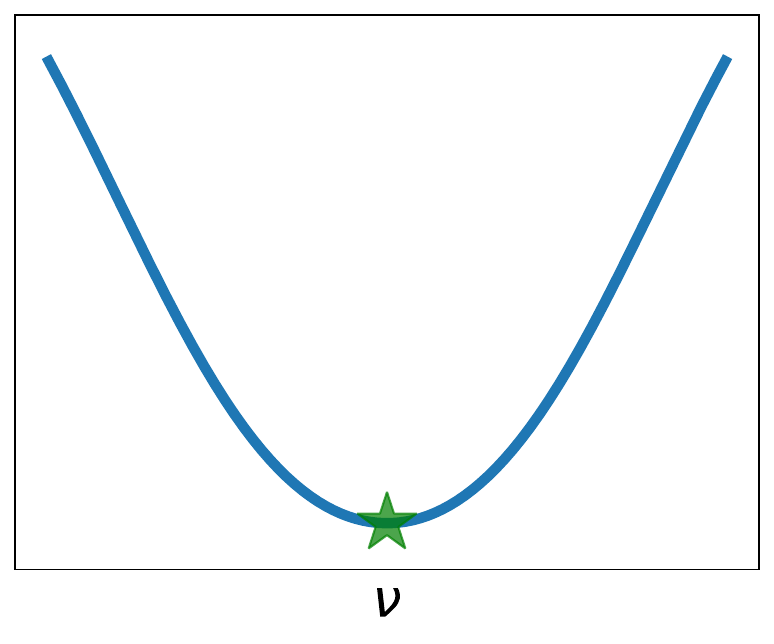}&
\includegraphics[scale=0.2]{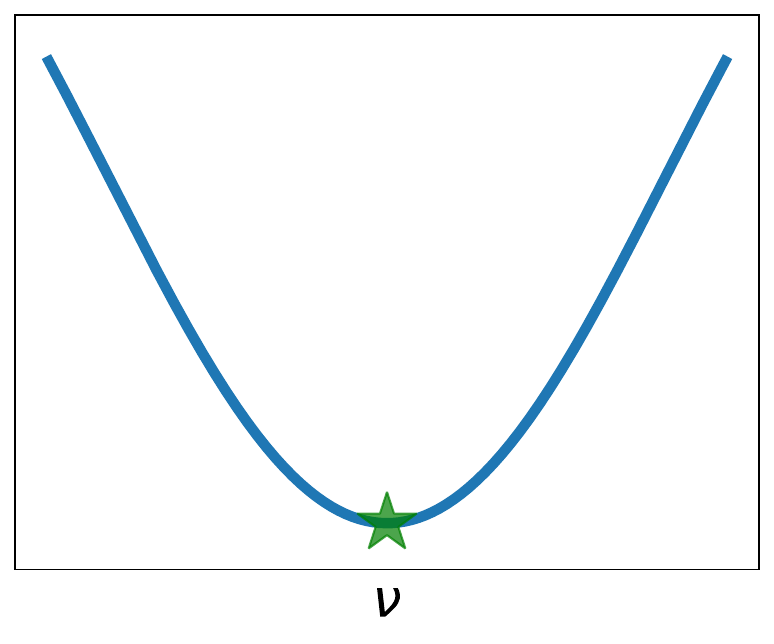}\\
\includegraphics[scale=0.2]{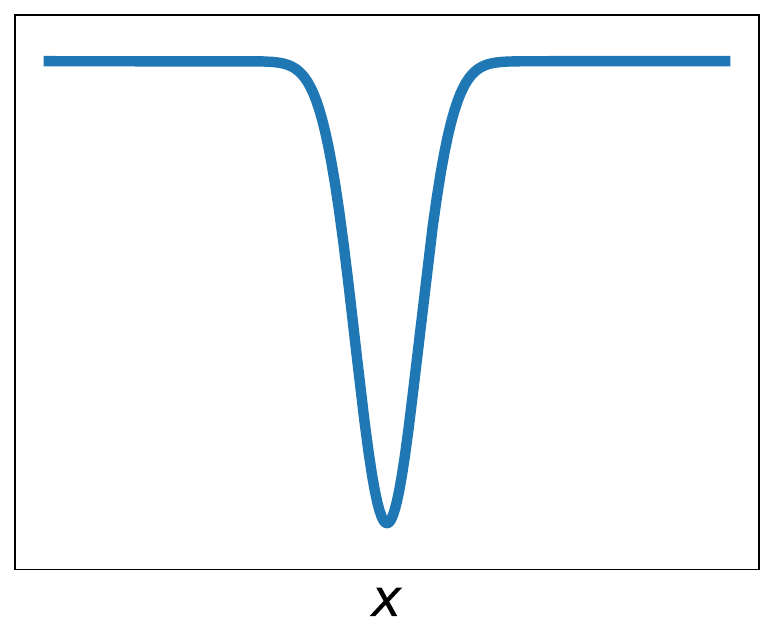}&
\includegraphics[scale=0.2]{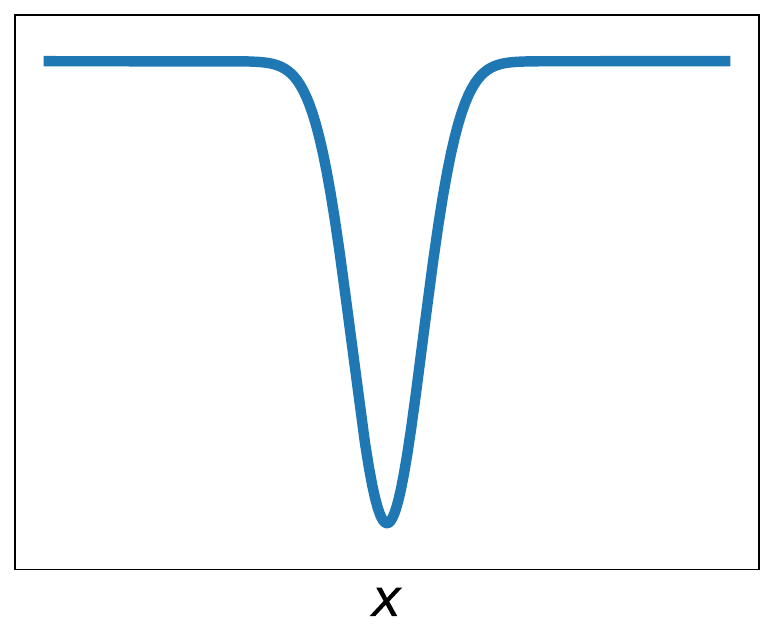}&
\includegraphics[scale=0.2]{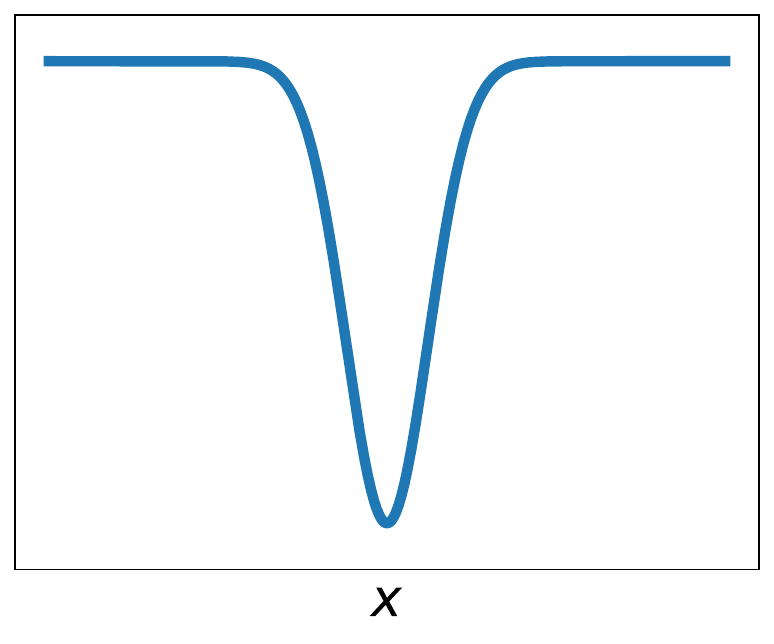}&
\includegraphics[scale=0.2]{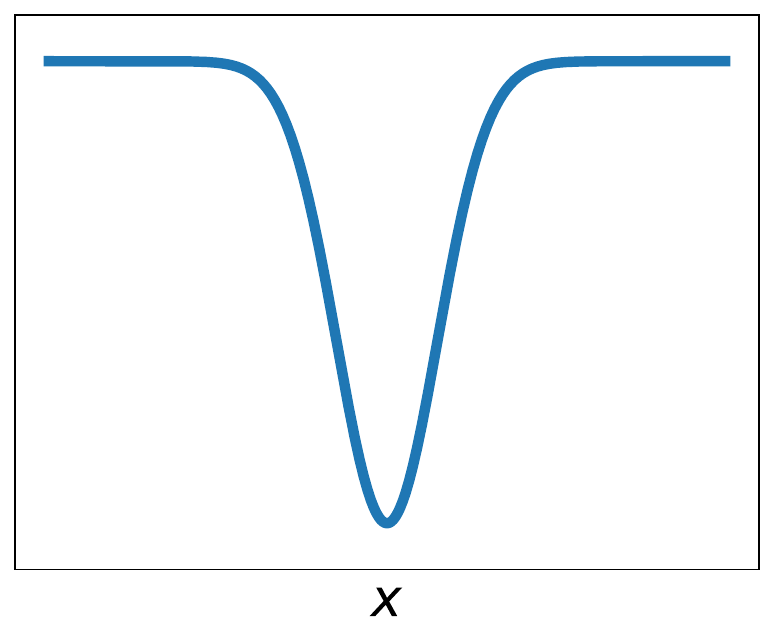}\\
\end{tabular}
\end{center}
\caption{Results for additional $\alpha$-values on $p_{1}$ (cf. \cref{sec:simulations_analytic}). First row for each $\alpha$ shows the divergence, the second row shows the associated \wa.
}\label{fig:alpha_exps}
\end{figure*}

In \cref{fig:alpha_exps}, we present additional results for the $\alpha$-DIV for $p_1$ in \cref{sec:simulations_analytic}. We find that choosing an $\alpha$ results in a strictly convex $\wa$, which guarantees a unique minimizer at the true mean as given in our \cref{theorem:theorem_alpha_1}. On the other hand, we also observe that lower values of $\alpha$, which do not result in a strictly convex \wa, can be sufficient. To get an intuition what $\wa$ looks like for some cases, please refer to the \emph{Gaussian} row in \cref{fig:wa_curves}.

\subsubsection{Additional results with stochastic optimization}\label{app:full_sgd_results}
In \cref{tab:results_simulations}, we provide additional results with stochastic optimization, complementing \cref{fig:results_simulations}. The metrics are defined in \cref{sec:learning}. For the experimental setup, please refer to \cref{app:simulations_setup}. 

We find that, whenever our theorems guarantee a unique minimizer corresponding to the correct mean or correlation we observe low error in practice, given that optimization is stable. For the $\alpha$-divergence with $\alpha=2.0$, for numbers denoted in black and bold in the table, learning was instable leading to non-negligible error in practice despite a theoretical guarantee of exact recovery.

\begin{sidewaystable}
\caption{\textbf{Whenever our sufficient conditions are satisfied, we recover the target characteristics with low error in practice.} We distinguish $3$ cases (1) \textbf{\textcolor{ForestGreen}{low error in practice with guarantees (by us or previous results)}}, (2) \textcolor{orange}{no guarantee but low error ($< 10^{-2}$) in practice}, (3) \textcolor{red}{no guarantee and high error ($> 10^{-2}$) in practice}. We note that for higher-dimensional targets, the $\alpha$-divergence with $\alpha=2.0$ occasionally \textbf{failed to optimize properly due to instability}, despite our guarantees. We denote these entries in black and bold. We highlight rows of approximations learned with the FKL and a Gaussian proposal via an asterisk (*) as mean and correlation recovery is guaranteed (regardless of target symmetry) via previous results by \citet{wainwright2008graphical}.
The metrics are defined in \cref{sec:learning}. We average results across \nruns runs.}\label{tab:results_simulations}
\resizebox{.99\textwidth}{!}
{
{
\begin{tabular}{llllllllll}
\toprule
& & \multicolumn{2}{c}{RKL} & \multicolumn{2}{c}{FKL} &
\multicolumn{2}{c}{$\alpha$-DIV ($\alpha=0.5$)} & \multicolumn{2}{c}{$\alpha$-DIV ($\alpha=2.0$)}\\
$p$ & $q$ &
$\Delta\vmu$ & $\Delta\text{Corr}$ &
$\Delta\vmu$ & $\Delta\text{Corr}$ &
$\Delta\vmu$ & $\Delta\text{Corr}$ &
$\Delta\vmu$ & $\Delta\text{Corr}$ \\
\midrule
\multirow{3}{*}{MoU} & Gaussian & / & / & $\textcolor{ForestGreen}{\boldsymbol{2.9 \cdot 10^{-4} {\scriptsize \pm 1.7 \cdot 10^{-4}}}}^{\boldsymbol{*}}$ & $\textcolor{ForestGreen}{\boldsymbol{6.3 \cdot 10^{-5} {\scriptsize \pm 3.7 \cdot 10^{-5}}}}^{\boldsymbol{*}}$ & $\textcolor{orange}{\boldsymbol{6.4 \cdot 10^{-4} {\scriptsize \pm 2.5 \cdot 10^{-4}}}}$ & $\textcolor{red}{\boldsymbol{2.1 \cdot 10^{-2} {\scriptsize \pm 1.3 \cdot 10^{-4}}}}$ & $\textcolor{ForestGreen}{\boldsymbol{2.0 \cdot 10^{-4} {\scriptsize \pm 1.1 \cdot 10^{-4}}}}$ & $\textcolor{red}{\boldsymbol{2.5 \cdot 10^{-2} {\scriptsize \pm 3.1 \cdot 10^{-4}}}}$ \\
  & Laplace & / & / & $\textcolor{orange}{\boldsymbol{2.1 \cdot 10^{-3} {\scriptsize \pm 2.8 \cdot 10^{-3}}}}$ & $\textcolor{red}{\boldsymbol{1.9 \cdot 10^{-2} {\scriptsize \pm 4.5 \cdot 10^{-5}}}}$ & $\textcolor{red}{\boldsymbol{9.1 \cdot 10^{-1} {\scriptsize \pm 1.9 \cdot 10^{-4}}}}$ & $\textcolor{red}{\boldsymbol{4.0 \cdot 10^{-1} {\scriptsize \pm 2.7 \cdot 10^{-3}}}}$ & $\textcolor{orange}{\boldsymbol{5.1 \cdot 10^{-4} {\scriptsize \pm 2.8 \cdot 10^{-4}}}}$ & $\textcolor{red}{\boldsymbol{1.0 \cdot 10^{-2} {\scriptsize \pm 2.4 \cdot 10^{-4}}}}$ \\
  & Student-t & / & / & $\textcolor{orange}{\boldsymbol{3.4 \cdot 10^{-4} {\scriptsize \pm 1.1 \cdot 10^{-4}}}}$ & $\textcolor{orange}{\boldsymbol{9.4 \cdot 10^{-3} {\scriptsize \pm 9.4 \cdot 10^{-5}}}}$ & $\textcolor{orange}{\boldsymbol{1.3 \cdot 10^{-3} {\scriptsize \pm 7.4 \cdot 10^{-4}}}}$ & $\textcolor{red}{\boldsymbol{3.1 \cdot 10^{-2} {\scriptsize \pm 2.2 \cdot 10^{-4}}}}$ & $\textcolor{ForestGreen}{\boldsymbol{4.8 \cdot 10^{-4} {\scriptsize \pm 2.7 \cdot 10^{-4}}}}$ & $\textcolor{red}{\boldsymbol{1.8 \cdot 10^{-2} {\scriptsize \pm 1.6 \cdot 10^{-4}}}}$ \\
\midrule
\multirow{3}{*}{GMM} & Gaussian & $\textcolor{red}{\boldsymbol{9.7 \cdot 10^{-1} {\scriptsize \pm 3.9 \cdot 10^{-4}}}}$ & $\textcolor{red}{\boldsymbol{6.6 \cdot 10^{-1} {\scriptsize \pm 3.9 \cdot 10^{-4}}}}$ & $\textcolor{ForestGreen}{\boldsymbol{1.9 \cdot 10^{-4} {\scriptsize \pm 7.5 \cdot 10^{-5}}}}^{\boldsymbol{*}}$ & $\textcolor{ForestGreen}{\boldsymbol{7.4 \cdot 10^{-4} {\scriptsize \pm 8.1 \cdot 10^{-4}}}}^{\boldsymbol{*}}$ & $\textcolor{red}{\boldsymbol{5.9 \cdot 10^{-1} {\scriptsize \pm 5.3 \cdot 10^{-1}}}}$ & $\textcolor{red}{\boldsymbol{3.9 \cdot 10^{-1} {\scriptsize \pm 3.6 \cdot 10^{-1}}}}$ & $\textcolor{ForestGreen}{\boldsymbol{9.1 \cdot 10^{-4} {\scriptsize \pm 5.7 \cdot 10^{-4}}}}$ & $\textcolor{orange}{\boldsymbol{1.6 \cdot 10^{-3} {\scriptsize \pm 9.3 \cdot 10^{-4}}}}$ \\
  & Laplace & $\textcolor{red}{\boldsymbol{9.7 \cdot 10^{-1} {\scriptsize \pm 5.7 \cdot 10^{-5}}}}$ & $\textcolor{red}{\boldsymbol{6.6 \cdot 10^{-1} {\scriptsize \pm 3.5 \cdot 10^{-4}}}}$ & $\textcolor{orange}{\boldsymbol{2.1 \cdot 10^{-3} {\scriptsize \pm 1.6 \cdot 10^{-3}}}}$ & $\textcolor{red}{\boldsymbol{1.3 \cdot 10^{-2} {\scriptsize \pm 1.8 \cdot 10^{-4}}}}$ & $\textcolor{red}{\boldsymbol{9.7 \cdot 10^{-1} {\scriptsize \pm 1.1 \cdot 10^{-4}}}}$ & $\textcolor{red}{\boldsymbol{6.6 \cdot 10^{-1} {\scriptsize \pm 4.4 \cdot 10^{-3}}}}$ & $\textcolor{orange}{\boldsymbol{1.0 \cdot 10^{-3} {\scriptsize \pm 9.1 \cdot 10^{-4}}}}$ & $\textcolor{orange}{\boldsymbol{8.0 \cdot 10^{-3} {\scriptsize \pm 7.5 \cdot 10^{-4}}}}$ \\
  & Student-t & $\textcolor{red}{\boldsymbol{9.7 \cdot 10^{-1} {\scriptsize \pm 3.0 \cdot 10^{-5}}}}$ & $\textcolor{red}{\boldsymbol{6.6 \cdot 10^{-1} {\scriptsize \pm 4.1 \cdot 10^{-4}}}}$ & $\textcolor{orange}{\boldsymbol{2.7 \cdot 10^{-4} {\scriptsize \pm 1.3 \cdot 10^{-4}}}}$ & $\textcolor{orange}{\boldsymbol{7.8 \cdot 10^{-3} {\scriptsize \pm 7.7 \cdot 10^{-4}}}}$ & $\textcolor{red}{\boldsymbol{9.7 \cdot 10^{-1} {\scriptsize \pm 2.7 \cdot 10^{-4}}}}$ & $\textcolor{red}{\boldsymbol{6.5 \cdot 10^{-1} {\scriptsize \pm 3.6 \cdot 10^{-3}}}}$ & $\textcolor{ForestGreen}{\boldsymbol{1.2 \cdot 10^{-3} {\scriptsize \pm 1.1 \cdot 10^{-3}}}}$ & $\textcolor{orange}{\boldsymbol{7.0 \cdot 10^{-3} {\scriptsize \pm 1.1 \cdot 10^{-4}}}}$ \\
\midrule
\multirow{3}{*}{Ellipse} & Gaussian & $\textcolor{red}{\boldsymbol{4.8 \cdot 10^{-1} {\scriptsize \pm 1.4 \cdot 10^{-1}}}}$ & $\textcolor{red}{\boldsymbol{6.1 \cdot 10^{-2} {\scriptsize \pm 2.6 \cdot 10^{-2}}}}$ & $\textcolor{ForestGreen}{\boldsymbol{5.5 \cdot 10^{-4} {\scriptsize \pm 2.9 \cdot 10^{-4}}}}^{\boldsymbol{*}}$ & $\textcolor{ForestGreen}{\boldsymbol{1.3 \cdot 10^{-4} {\scriptsize \pm 6.7 \cdot 10^{-5}}}}^{\boldsymbol{*}}$ & $\textcolor{orange}{\boldsymbol{1.2 \cdot 10^{-3} {\scriptsize \pm 2.4 \cdot 10^{-4}}}}$ & $\textcolor{orange}{\boldsymbol{1.3 \cdot 10^{-4} {\scriptsize \pm 9.5 \cdot 10^{-5}}}}$ & $\textcolor{ForestGreen}{\boldsymbol{4.5 \cdot 10^{-4} {\scriptsize \pm 2.5 \cdot 10^{-4}}}}$ & $\textcolor{ForestGreen}{\boldsymbol{1.3 \cdot 10^{-4} {\scriptsize \pm 1.4 \cdot 10^{-4}}}}$ \\
  & Laplace & $\textcolor{red}{\boldsymbol{5.5 \cdot 10^{-1} {\scriptsize \pm 1.6 \cdot 10^{-1}}}}$ & $\textcolor{red}{\boldsymbol{5.5 \cdot 10^{-2} {\scriptsize \pm 3.8 \cdot 10^{-2}}}}$ & $\textcolor{orange}{\boldsymbol{1.1 \cdot 10^{-3} {\scriptsize \pm 6.1 \cdot 10^{-4}}}}$ & $\textcolor{orange}{\boldsymbol{1.1 \cdot 10^{-4} {\scriptsize \pm 1.3 \cdot 10^{-4}}}}$ & $\textcolor{red}{\boldsymbol{1.3 \cdot 10^{-1} {\scriptsize \pm 1.2 \cdot 10^{-1}}}}$ & $\textcolor{orange}{\boldsymbol{2.3 \cdot 10^{-3} {\scriptsize \pm 2.4 \cdot 10^{-3}}}}$ & $\textcolor{orange}{\boldsymbol{1.7 \cdot 10^{-3} {\scriptsize \pm 2.1 \cdot 10^{-3}}}}$ & $\textcolor{orange}{\boldsymbol{9.9 \cdot 10^{-4} {\scriptsize \pm 1.3 \cdot 10^{-3}}}}$ \\
  & Student-t & $\textcolor{red}{\boldsymbol{5.2 \cdot 10^{-1} {\scriptsize \pm 1.7 \cdot 10^{-1}}}}$ & $\textcolor{red}{\boldsymbol{4.5 \cdot 10^{-2} {\scriptsize \pm 2.4 \cdot 10^{-2}}}}$ & $\textcolor{orange}{\boldsymbol{6.7 \cdot 10^{-4} {\scriptsize \pm 4.0 \cdot 10^{-4}}}}$ & $\textcolor{orange}{\boldsymbol{1.0 \cdot 10^{-4} {\scriptsize \pm 6.1 \cdot 10^{-5}}}}$ & $\textcolor{orange}{\boldsymbol{1.4 \cdot 10^{-3} {\scriptsize \pm 8.8 \cdot 10^{-4}}}}$ & $\textcolor{orange}{\boldsymbol{3.4 \cdot 10^{-4} {\scriptsize \pm 2.9 \cdot 10^{-4}}}}$ & $\textcolor{ForestGreen}{\boldsymbol{5.8 \cdot 10^{-4} {\scriptsize \pm 2.6 \cdot 10^{-4}}}}$ & $\textcolor{orange}{\boldsymbol{1.5 \cdot 10^{-3} {\scriptsize \pm 2.4 \cdot 10^{-3}}}}$ \\
\midrule
\multirow{3}{*}{A-GMM} & Gaussian & $\textcolor{red}{\boldsymbol{3.9 \cdot 10^{-1} {\scriptsize \pm 4.9 \cdot 10^{-1}}}}$ & $\textcolor{red}{\boldsymbol{8.7 \cdot 10^{-2} {\scriptsize \pm 1.6 \cdot 10^{-1}}}}$ & $\textcolor{ForestGreen}{\boldsymbol{2.6 \cdot 10^{-4} {\scriptsize \pm 1.6 \cdot 10^{-4}}}}^{\boldsymbol{*}}$ & $\textcolor{ForestGreen}{\boldsymbol{1.3 \cdot 10^{-4} {\scriptsize \pm 1.4 \cdot 10^{-4}}}}^{\boldsymbol{*}}$ & $\textcolor{orange}{\boldsymbol{1.6 \cdot 10^{-3} {\scriptsize \pm 9.3 \cdot 10^{-4}}}}$ & $\textcolor{orange}{\boldsymbol{4.9 \cdot 10^{-3} {\scriptsize \pm 1.8 \cdot 10^{-4}}}}$ & $\textcolor{red}{\boldsymbol{4.9 \cdot 10^{-2} {\scriptsize \pm 2.0 \cdot 10^{-3}}}}$ & $\textcolor{orange}{\boldsymbol{7.4 \cdot 10^{-3} {\scriptsize \pm 8.6 \cdot 10^{-4}}}}$ \\
  & Laplace & $\textcolor{red}{\boldsymbol{9.7 \cdot 10^{-1} {\scriptsize \pm 3.1 \cdot 10^{-4}}}}$ & $\textcolor{red}{\boldsymbol{4.2 \cdot 10^{-1} {\scriptsize \pm 1.3 \cdot 10^{-1}}}}$ & $\textcolor{red}{\boldsymbol{5.8 \cdot 10^{-1} {\scriptsize \pm 8.0 \cdot 10^{-3}}}}$ & $\textcolor{red}{\boldsymbol{1.2 \cdot 10^{-2} {\scriptsize \pm 2.9 \cdot 10^{-4}}}}$ & $\textcolor{red}{\boldsymbol{9.7 \cdot 10^{-1} {\scriptsize \pm 2.7 \cdot 10^{-4}}}}$ & $\textcolor{red}{\boldsymbol{5.0 \cdot 10^{-1} {\scriptsize \pm 1.4 \cdot 10^{-1}}}}$ & $\textcolor{red}{\boldsymbol{1.6 \cdot 10^{-1} {\scriptsize \pm 3.2 \cdot 10^{-3}}}}$ & $\textcolor{orange}{\boldsymbol{8.0 \cdot 10^{-3} {\scriptsize \pm 7.5 \cdot 10^{-4}}}}$ \\
  & Student-t & $\textcolor{red}{\boldsymbol{9.7 \cdot 10^{-1} {\scriptsize \pm 5.7 \cdot 10^{-3}}}}$ & $\textcolor{red}{\boldsymbol{3.6 \cdot 10^{-1} {\scriptsize \pm 3.9 \cdot 10^{-4}}}}$ & $\textcolor{red}{\boldsymbol{9.6 \cdot 10^{-2} {\scriptsize \pm 8.0 \cdot 10^{-4}}}}$ & $\textcolor{orange}{\boldsymbol{9.3 \cdot 10^{-3} {\scriptsize \pm 5.3 \cdot 10^{-4}}}}$ & $\textcolor{red}{\boldsymbol{3.4 \cdot 10^{-1} {\scriptsize \pm 3.5 \cdot 10^{-1}}}}$ & $\textcolor{red}{\boldsymbol{8.0 \cdot 10^{-2} {\scriptsize \pm 1.5 \cdot 10^{-1}}}}$ & $\textcolor{red}{\boldsymbol{8.5 \cdot 10^{-2} {\scriptsize \pm 2.1 \cdot 10^{-3}}}}$ & $\textcolor{orange}{\boldsymbol{5.0 \cdot 10^{-3} {\scriptsize \pm 1.0 \cdot 10^{-3}}}}$ \\
\midrule
\multirow{2}{*}{GMM (4D)} & Gaussian & $\textcolor{orange}{\boldsymbol{1.6 \cdot 10^{-3} {\scriptsize \pm 3.3 \cdot 10^{-4}}}}$ & $\textcolor{red}{\boldsymbol{4.2 \cdot 10^{-2} {\scriptsize \pm 1.7 \cdot 10^{-3}}}}$ & $\textcolor{ForestGreen}{\boldsymbol{6.0 \cdot 10^{-3} {\scriptsize \pm 7.5 \cdot 10^{-3}}}}^{\boldsymbol{*}}$ & $\textcolor{ForestGreen}{\boldsymbol{5.6 \cdot 10^{-4} {\scriptsize \pm 2.1 \cdot 10^{-4}}}}^{\boldsymbol{*}}$ & $\textcolor{orange}{\boldsymbol{1.6 \cdot 10^{-3} {\scriptsize \pm 6.5 \cdot 10^{-4}}}}$ & $\textcolor{red}{\boldsymbol{1.6 \cdot 10^{-2} {\scriptsize \pm 1.7 \cdot 10^{-3}}}}$ & $\textcolor{ForestGreen}{\boldsymbol{1.7 \cdot 10^{-3} {\scriptsize \pm 7.4 \cdot 10^{-4}}}}$ & $\textcolor{red}{\boldsymbol{1.6 \cdot 10^{-2} {\scriptsize \pm 1.0 \cdot 10^{-3}}}}$ \\
  & Student-t & $\textcolor{orange}{\boldsymbol{4.5 \cdot 10^{-3} {\scriptsize \pm 2.5 \cdot 10^{-3}}}}$ & $\textcolor{red}{\boldsymbol{6.8 \cdot 10^{-2} {\scriptsize \pm 1.1 \cdot 10^{-3}}}}$ & $\textcolor{orange}{\boldsymbol{1.2 \cdot 10^{-3} {\scriptsize \pm 6.5 \cdot 10^{-4}}}}$ & $\textcolor{red}{\boldsymbol{3.1 \cdot 10^{-2} {\scriptsize \pm 3.2 \cdot 10^{-4}}}}$ & $\textcolor{orange}{\boldsymbol{2.7 \cdot 10^{-3} {\scriptsize \pm 8.8 \cdot 10^{-4}}}}$ & $\textcolor{red}{\boldsymbol{1.1 \cdot 10^{-2} {\scriptsize \pm 6.2 \cdot 10^{-4}}}}$ & $\textcolor{ForestGreen}{\boldsymbol{1.6 \cdot 10^{-3} {\scriptsize \pm 4.8 \cdot 10^{-4}}}}$ & $\textcolor{red}{\boldsymbol{4.4 \cdot 10^{-2} {\scriptsize \pm 8.0 \cdot 10^{-4}}}}$ \\
\midrule
\multirow{2}{*}{GMM (8D)} & Gaussian & $\textcolor{red}{\boldsymbol{1.4 \cdot 10^{-1} {\scriptsize \pm 3.1 \cdot 10^{-1}}}}$ & $\textcolor{red}{\boldsymbol{1.4 \cdot 10^{-1} {\scriptsize \pm 1.6 \cdot 10^{-1}}}}$ & $\textcolor{ForestGreen}{\boldsymbol{9.3 \cdot 10^{-4} {\scriptsize \pm 2.3 \cdot 10^{-4}}}}^{\boldsymbol{*}}$ & $\textcolor{ForestGreen}{\boldsymbol{1.0 \cdot 10^{-3} {\scriptsize \pm 1.9 \cdot 10^{-4}}}}^{\boldsymbol{*}}$ & $\textcolor{orange}{\boldsymbol{7.6 \cdot 10^{-3} {\scriptsize \pm 1.2 \cdot 10^{-2}}}}$ & $\textcolor{red}{\boldsymbol{1.9 \cdot 10^{-2} {\scriptsize \pm 1.7 \cdot 10^{-3}}}}$ & ${\boldsymbol{2.3 \cdot 10^{-1} {\scriptsize \pm 5.1 \cdot 10^{-1}}}}$ & $\textcolor{red}{\boldsymbol{5.6 \cdot 10^{-2} {\scriptsize \pm 9.7 \cdot 10^{-2}}}}$ \\
  & Student-t & $\textcolor{red}{\boldsymbol{6.9 \cdot 10^{-1} {\scriptsize \pm 5.0 \cdot 10^{-4}}}}$ & $\textcolor{red}{\boldsymbol{4.1 \cdot 10^{-1} {\scriptsize \pm 3.7 \cdot 10^{-4}}}}$ & $\textcolor{orange}{\boldsymbol{1.3 \cdot 10^{-3} {\scriptsize \pm 5.0 \cdot 10^{-4}}}}$ & $\textcolor{red}{\boldsymbol{2.9 \cdot 10^{-2} {\scriptsize \pm 2.6 \cdot 10^{-4}}}}$ & $\textcolor{orange}{\boldsymbol{3.2 \cdot 10^{-3} {\scriptsize \pm 8.8 \cdot 10^{-4}}}}$ & $\textcolor{red}{\boldsymbol{1.2 \cdot 10^{-2} {\scriptsize \pm 8.4 \cdot 10^{-4}}}}$ & $\textcolor{ForestGreen}{\boldsymbol{2.3 \cdot 10^{-3} {\scriptsize \pm 6.0 \cdot 10^{-4}}}}$ & $\textcolor{red}{\boldsymbol{3.4 \cdot 10^{-2} {\scriptsize \pm 6.7 \cdot 10^{-4}}}}$ \\
\midrule
\multirow{2}{*}{GMM (16D)} & Gaussian & $\textcolor{orange}{\boldsymbol{1.5 \cdot 10^{-3} {\scriptsize \pm 9.1 \cdot 10^{-4}}}}$ & $\textcolor{red}{\boldsymbol{8.5 \cdot 10^{-2} {\scriptsize \pm 3.0 \cdot 10^{-4}}}}$ & $\textcolor{ForestGreen}{\boldsymbol{1.2 \cdot 10^{-3} {\scriptsize \pm 2.8 \cdot 10^{-4}}}}^{\boldsymbol{*}}$ & $\textcolor{ForestGreen}{\boldsymbol{9.1 \cdot 10^{-4} {\scriptsize \pm 1.2 \cdot 10^{-4}}}}^{\boldsymbol{*}}$ & $\textcolor{orange}{\boldsymbol{5.0 \cdot 10^{-3} {\scriptsize \pm 5.8 \cdot 10^{-3}}}}$ & $\textcolor{red}{\boldsymbol{1.3 \cdot 10^{-2} {\scriptsize \pm 1.3 \cdot 10^{-3}}}}$ & $\textcolor{ForestGreen}{\boldsymbol{1.7 \cdot 10^{-3} {\scriptsize \pm 5.3 \cdot 10^{-4}}}}$ & $\textcolor{orange}{\boldsymbol{7.3 \cdot 10^{-3} {\scriptsize \pm 1.0 \cdot 10^{-3}}}}$ \\
  & Student-t & $\textcolor{red}{\boldsymbol{7.0 \cdot 10^{-1} {\scriptsize \pm 2.6 \cdot 10^{-4}}}}$ & $\textcolor{red}{\boldsymbol{4.6 \cdot 10^{-1} {\scriptsize \pm 1.6 \cdot 10^{-4}}}}$ & $\textcolor{orange}{\boldsymbol{1.3 \cdot 10^{-3} {\scriptsize \pm 4.6 \cdot 10^{-4}}}}$ & $\textcolor{red}{\boldsymbol{2.1 \cdot 10^{-2} {\scriptsize \pm 1.3 \cdot 10^{-4}}}}$ & $\textcolor{orange}{\boldsymbol{2.8 \cdot 10^{-3} {\scriptsize \pm 1.0 \cdot 10^{-3}}}}$ & $\textcolor{red}{\boldsymbol{1.5 \cdot 10^{-2} {\scriptsize \pm 9.0 \cdot 10^{-4}}}}$ & $\textcolor{ForestGreen}{\boldsymbol{2.0 \cdot 10^{-3} {\scriptsize \pm 3.5 \cdot 10^{-4}}}}$ & $\textcolor{red}{\boldsymbol{2.0 \cdot 10^{-2} {\scriptsize \pm 5.2 \cdot 10^{-4}}}}$ \\
\midrule
\multirow{2}{*}{Ellipse (4D)} & Gaussian & $\textcolor{orange}{\boldsymbol{4.2 \cdot 10^{-3} {\scriptsize \pm 1.1 \cdot 10^{-3}}}}$ & $\textcolor{orange}{\boldsymbol{2.0 \cdot 10^{-3} {\scriptsize \pm 3.8 \cdot 10^{-4}}}}$ & $\textcolor{ForestGreen}{\boldsymbol{2.9 \cdot 10^{-3} {\scriptsize \pm 7.4 \cdot 10^{-4}}}}^{\boldsymbol{*}}$ & $\textcolor{ForestGreen}{\boldsymbol{1.6 \cdot 10^{-3} {\scriptsize \pm 6.3 \cdot 10^{-4}}}}^{\boldsymbol{*}}$ & $\textcolor{orange}{\boldsymbol{4.4 \cdot 10^{-3} {\scriptsize \pm 1.9 \cdot 10^{-3}}}}$ & $\textcolor{orange}{\boldsymbol{3.9 \cdot 10^{-3} {\scriptsize \pm 1.1 \cdot 10^{-3}}}}$ & $\textcolor{ForestGreen}{\boldsymbol{4.3 \cdot 10^{-3} {\scriptsize \pm 1.2 \cdot 10^{-3}}}}$ & $\textcolor{ForestGreen}{\boldsymbol{3.1 \cdot 10^{-3} {\scriptsize \pm 1.4 \cdot 10^{-3}}}}$ \\
  & Student-t & $\textcolor{orange}{\boldsymbol{3.1 \cdot 10^{-3} {\scriptsize \pm 1.9 \cdot 10^{-3}}}}$ & $\textcolor{orange}{\boldsymbol{2.1 \cdot 10^{-3} {\scriptsize \pm 9.6 \cdot 10^{-4}}}}$ & $\textcolor{orange}{\boldsymbol{3.6 \cdot 10^{-3} {\scriptsize \pm 2.1 \cdot 10^{-3}}}}$ & $\textcolor{orange}{\boldsymbol{1.7 \cdot 10^{-3} {\scriptsize \pm 9.8 \cdot 10^{-4}}}}$ & $\textcolor{orange}{\boldsymbol{4.2 \cdot 10^{-3} {\scriptsize \pm 2.0 \cdot 10^{-3}}}}$ & $\textcolor{orange}{\boldsymbol{4.2 \cdot 10^{-3} {\scriptsize \pm 1.3 \cdot 10^{-3}}}}$ & $\textcolor{ForestGreen}{\boldsymbol{2.6 \cdot 10^{-3} {\scriptsize \pm 1.1 \cdot 10^{-3}}}}$ & $\textcolor{orange}{\boldsymbol{2.5 \cdot 10^{-3} {\scriptsize \pm 1.8 \cdot 10^{-3}}}}$ \\
\midrule
\multirow{2}{*}{Ellipse (8D)} & Gaussian & $\textcolor{red}{\boldsymbol{2.1 \cdot 10^{-2} {\scriptsize \pm 7.6 \cdot 10^{-3}}}}$ & $\textcolor{orange}{\boldsymbol{3.4 \cdot 10^{-3} {\scriptsize \pm 7.9 \cdot 10^{-4}}}}$ & $\textcolor{ForestGreen}{\boldsymbol{1.3 \cdot 10^{-3} {\scriptsize \pm 2.5 \cdot 10^{-4}}}}^{\boldsymbol{*}}$ & $\textcolor{ForestGreen}{\boldsymbol{1.0 \cdot 10^{-3} {\scriptsize \pm 1.7 \cdot 10^{-4}}}}^{\boldsymbol{*}}$ & $\textcolor{orange}{\boldsymbol{2.2 \cdot 10^{-3} {\scriptsize \pm 5.2 \cdot 10^{-4}}}}$ & $\textcolor{orange}{\boldsymbol{1.7 \cdot 10^{-3} {\scriptsize \pm 2.7 \cdot 10^{-4}}}}$ & $\textcolor{ForestGreen}{\boldsymbol{1.3 \cdot 10^{-3} {\scriptsize \pm 3.8 \cdot 10^{-4}}}}$ & $\textcolor{ForestGreen}{\boldsymbol{1.0 \cdot 10^{-3} {\scriptsize \pm 2.2 \cdot 10^{-4}}}}$ \\
  & Student-t & $\textcolor{red}{\boldsymbol{2.1 \cdot 10^{-1} {\scriptsize \pm 7.9 \cdot 10^{-2}}}}$ & $\textcolor{red}{\boldsymbol{1.3 \cdot 10^{-2} {\scriptsize \pm 5.1 \cdot 10^{-3}}}}$ & $\textcolor{orange}{\boldsymbol{1.2 \cdot 10^{-3} {\scriptsize \pm 3.4 \cdot 10^{-4}}}}$ & $\textcolor{orange}{\boldsymbol{1.0 \cdot 10^{-3} {\scriptsize \pm 1.7 \cdot 10^{-4}}}}$ & $\textcolor{orange}{\boldsymbol{2.1 \cdot 10^{-3} {\scriptsize \pm 5.4 \cdot 10^{-4}}}}$ & $\textcolor{orange}{\boldsymbol{1.6 \cdot 10^{-3} {\scriptsize \pm 2.7 \cdot 10^{-4}}}}$ & $\textcolor{ForestGreen}{\boldsymbol{1.3 \cdot 10^{-3} {\scriptsize \pm 3.0 \cdot 10^{-4}}}}$ & $\textcolor{orange}{\boldsymbol{1.2 \cdot 10^{-3} {\scriptsize \pm 2.1 \cdot 10^{-4}}}}$ \\
\midrule
\multirow{2}{*}{Ellipse (16D)} & Gaussian & $\textcolor{red}{\boldsymbol{4.5 \cdot 10^{-1} {\scriptsize \pm 2.8 \cdot 10^{-2}}}}$ & $\textcolor{red}{\boldsymbol{2.6 \cdot 10^{-2} {\scriptsize \pm 3.5 \cdot 10^{-3}}}}$ & $\textcolor{ForestGreen}{\boldsymbol{1.0 \cdot 10^{-3} {\scriptsize \pm 2.0 \cdot 10^{-4}}}}^{\boldsymbol{*}}$ & $\textcolor{ForestGreen}{\boldsymbol{8.8 \cdot 10^{-4} {\scriptsize \pm 1.2 \cdot 10^{-4}}}}^{\boldsymbol{*}}$ & $\textcolor{orange}{\boldsymbol{1.8 \cdot 10^{-3} {\scriptsize \pm 3.4 \cdot 10^{-4}}}}$ & $\textcolor{orange}{\boldsymbol{1.3 \cdot 10^{-3} {\scriptsize \pm 1.2 \cdot 10^{-4}}}}$ & ${\boldsymbol{1.3 \cdot 10^{-2} {\scriptsize \pm 1.6 \cdot 10^{-2}}}}$ & $\textcolor{ForestGreen}{\boldsymbol{4.4 \cdot 10^{-3} {\scriptsize \pm 2.2 \cdot 10^{-3}}}}$ \\
  & Student-t & $\textcolor{red}{\boldsymbol{6.0 \cdot 10^{-1} {\scriptsize \pm 2.0 \cdot 10^{-2}}}}$ & $\textcolor{red}{\boldsymbol{2.9 \cdot 10^{-2} {\scriptsize \pm 4.0 \cdot 10^{-3}}}}$ & $\textcolor{orange}{\boldsymbol{1.1 \cdot 10^{-3} {\scriptsize \pm 1.4 \cdot 10^{-4}}}}$ & $\textcolor{orange}{\boldsymbol{9.4 \cdot 10^{-4} {\scriptsize \pm 1.3 \cdot 10^{-4}}}}$ & $\textcolor{orange}{\boldsymbol{1.7 \cdot 10^{-3} {\scriptsize \pm 5.0 \cdot 10^{-4}}}}$ & $\textcolor{orange}{\boldsymbol{1.4 \cdot 10^{-3} {\scriptsize \pm 1.7 \cdot 10^{-4}}}}$ & $\textcolor{ForestGreen}{\boldsymbol{1.4 \cdot 10^{-3} {\scriptsize \pm 3.3 \cdot 10^{-4}}}}$ & $\textcolor{orange}{\boldsymbol{1.2 \cdot 10^{-3} {\scriptsize \pm 1.2 \cdot 10^{-4}}}}$ \\
\bottomrule
\end{tabular}
}
}

\end{sidewaystable}

\end{document}